\documentclass[runningheads]{llncs}

\usepackage{eccv}

\usepackage{eccvabbrv}

\usepackage{graphicx}
\usepackage{booktabs}

\usepackage[accsupp]{axessibility}  

\usepackage{hyperref}

\usepackage{orcidlink}

\graphicspath{{./images/}}
\usepackage{braket}

\def\equationautorefname~#1\null{Equation~(#1)\null}

\def\etal{{et al.}}
\def\ie{{i.e.}}
\def\eg{{e.g.}}

\definecolor{dblue}{rgb}{0.0,0.0,0.5}
\definecolor{dgreen}{rgb}{0.0,0.5,0.0}
\definecolor{dred}{rgb}{0.6,0.0,0.0}
\definecolor{dorange}{rgb}{0.6,0.25,0.0}
\definecolor{dyellow}{rgb}{0.5,0.5,0.0}

\newcommand{\ignorethis}[1]{}

\let\shortcite\cite  
\renewcommand\paragraph[1]{\vspace{0.2cm}\noindent\textit{#1}}

\usepackage{colortbl}
\usepackage{paralist}

\definecolor{best}{RGB}{255, 160, 160}
\definecolor{second}{RGB}{255, 224, 192}
\definecolor{third}{RGB}{255, 255, 204}

\usepackage{tikz}
\usepackage{pgfplots}
\pgfplotsset{compat=1.18}
\usepgfplotslibrary{groupplots}

\usepackage{overpic}    

\usepackage{multirow}
\usepackage{rotating}

\def\RF2{RenderFormer-V2\xspace}

\begin{document}

\title{\RF2: Neural Rendering with Heterogeneous Scene Primitives}

\titlerunning{\RF2}

\author{Chong Zeng\inst{1}\orcidlink{0009-0004-6373-6848} \and
  Yue Dong\inst{2}\orcidlink{0000-0003-0362-337X} \and
  Pieter Peers\inst{3}\orcidlink{0000-0001-7621-9808} \and
  Lvmin Zhang\inst{1}\orcidlink{0000-0003-3503-5791} \and
  Maneesh Agrawala\inst{1}\orcidlink{0000-0002-8996-7327}
  }

\authorrunning{C. Zeng et al.}

\institute{Stanford University, USA \and
  Microsoft Research, China \and
  College of William \& Mary, USA}

\maketitle


\begin{abstract}
  We present '\RF2', a unified learned transformer-based neural
  rendering model, complementary to modern physics-based rendering
  systems, that can handle diverse light-transport effects such as
  caustics, volumetric scattering, environment lighting, textured and
  displaced surfaces and out-of-distribution materials without
  per-scene training or specialized code.  \RF2 models global
  light transport as a sequence-to-sequence transformation. Following
  its predecessor, \RF2 also employs a two stage process: a
  view-independent stage that resolves intra-scene primitive to
  primitive transport, and a view-dependent stage that transforms the
  internal neural scene representation into image pixels.  Different
  from RenderFormer, our model employs a novel combined
  windowed-attention and rendering-informed attention sink in the
  view-independent stage to improve scalability while maintaining
  render accuracy.  To further improve versatility, \RF2 supports
  heterogeneous scene primitives, including environment maps and
  participating media, and it employs a material encoding independent
  of the underlying surface reflectance model that encodes material
  appearance via a novel neural embedding.  We demonstrate the
  versatility of \RF2 on a variety of scenes and perform an
  extensive ablation of the improved attention mechanism.
\end{abstract}


\section{Introduction}

Neural rendering aims to visualize virtual scenes without relying on
manually encoded rules of light transport, but instead based on
relations between geometry, materials, and light learned from data.
Many neural rendering solutions offer limited generalizability beyond
the training data~\cite{Granskog:2021:NSG,Granskog:2020:CNS} or rely
on per-scene training strategies~\cite{Tewari:2022:ANR}.  Recently,
RenderFormer~\cite{Zeng:2025:RFT} formulated light transport
simulation as a regressive sequence-to-sequence translation problem,
where an input sequence of triangle tokens is transformed into
pixel-patch tokens through a transformer-based two stage pipeline: a
view-independent stage that resolves light transport between
triangles, and a view-dependent stage that resolves transport from
triangles to the camera.  Once trained, RenderFormer can render a wide
variety of virtual scenes without fine-tuning or further training.
Although RenderFormer is more general than prior solutions, it is still far
from practical: it is limited to scenes of less than $4$k triangles,
it only supports a hard-coded GGX BRDF model~\cite{Walter:2007:MMR},
and it is limited to scenes with (max. $8$) triangular diffuse light
sources.


\begin{figure}[tb]
    \centering
    \begin{subfigure}{0.25\textwidth}
        \centering
        \includegraphics[
            width=\linewidth,
            height=\linewidth
        ]{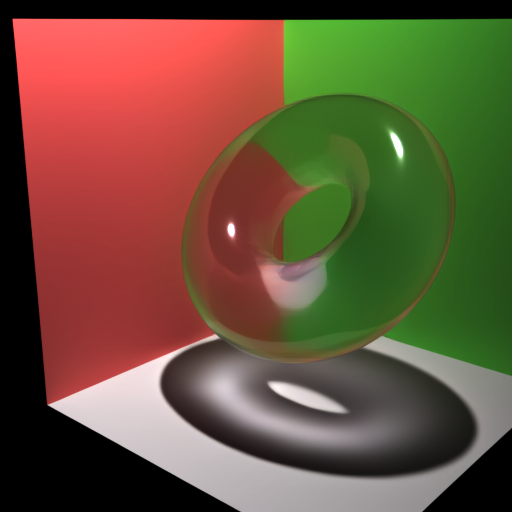}
    \end{subfigure}\hfill
    \begin{subfigure}{0.25\textwidth}
        \centering
        \includegraphics[
            width=\linewidth,
            height=\linewidth
        ]{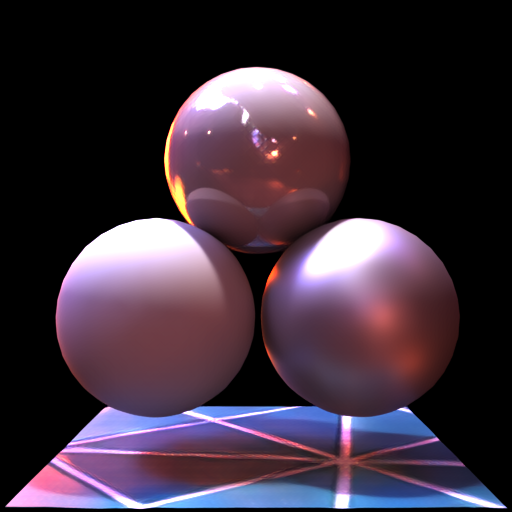}
    \end{subfigure}\hfill
    \begin{subfigure}{0.25\textwidth}
        \centering
        \includegraphics[
            width=\linewidth,
            height=\linewidth
        ]{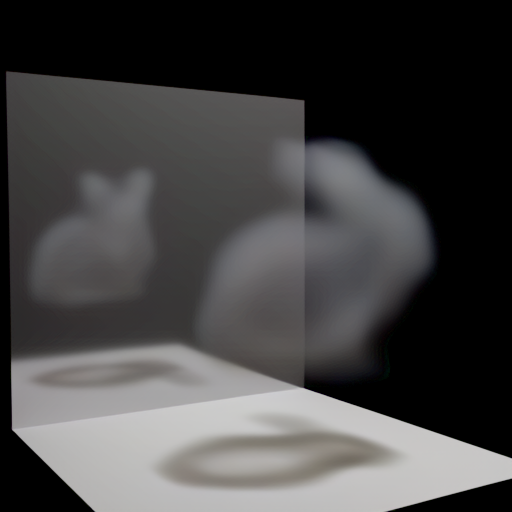}
    \end{subfigure}\hfill
    \begin{subfigure}{0.25\textwidth}
        \centering
        \includegraphics[
            width=\linewidth,
            height=\linewidth
        ]{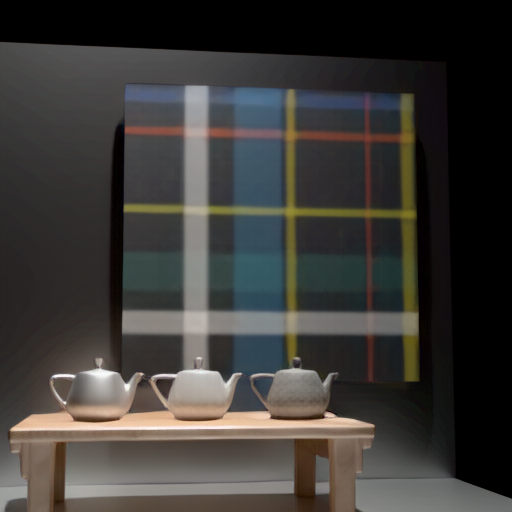}
    \end{subfigure}
    \caption{\RF2 can simulate global light transport in scenes that
      include transparent materials with caustics, environment
      lighting, volumetric scattering, textures, and scenes with over
      $100$k primitives, without the need for per-scene training.}
    \label{fig:teaser}
\end{figure}

In this paper we introduce '\RF2', a versatile transformer-based
neural rendering system that addresses RenderFormer's limitations via
a number of carefully designed architectural innovations
(\autoref{fig:teaser}).  Similar to RenderFormer, \RF2 formulates
light transport simulation as a regressive sequence-to-sequence
translation.  RenderFormer's main bottleneck in supporting larger
triangle meshes is the brute-force self-attention between triangle
tokens in the view-independent stage. Not only does this have a
quadratic complexity with respect to the number of triangles, it also
leads RenderFormer to loose focus for very large triangle meshes.
Inspired by recent advances in supporting larger context windows for
large-language models, \RF2 employs a novel sparse-attention variant,
consisting of a combination of windowed attention~\cite{Liu:2021:STH}
and render-aware attention-sinks~\cite{Xiao:2024:ESL}, tuned for
resolving view-independent light transport.  Furthermore, to improve
generalizability, we allow for other primitives than triangles (\eg,
voxels) and employ a simpler positional encoding based on the centroid
of the primitive.

\RF2 further decouples the material specification from a hard-coded
BRDF model, and instead employs a latent material embedding for
encoding different BRDF models including transparent and measured
materials.  A key observation is that the latent material encoding
does not need to be invertible to the input (BRDF) parameters, but it
only needs to encode the appearance of the material; we rely on \RF2
to learn how to map the material appearance into pixel values.
Moreover, to model spatially varying materials, we reuse a pretrained
VAE encoder~\cite{Wu:2025:QwenImage} to encode $32 \times 32$ texture
patches (of latent material properties) per primitive.  Similar to the
latent material appearance space, we only require the VAE encoder, and
let \RF2 learn how to interpret the encoded textures during rendering.

Whereas RenderFormer has a dedicated emittance parameter associated
with each triangle to model light sources, we leverage \RF2's
ability to mix different primitives to embed different lighting types,
ranging from triangular light sources to environment maps, into
specialized tokens.

We demonstrate the versatility of \RF2 by rendering more complex and
larger scenes than RenderFormer with a greater variety in lighting and materials.
We perform an in-depth ablation study to validate our design decisions.
The trained \RF2 model and code can be found at: \url{https://renderformer.github.io/v2}.

\section{Related Work}

\paragraph{Neural Rendering}
\cite{Tewari:2022:ANR} aims to predict the effects of light transport
through a virtual scene. Early work in neural rendering employs
specially learned neural representations of the
scene~\cite{Granskog:2020:CNS,Granskog:2021:NSG,Yuan:2022:NGE,Haque:2023:INE,Zheng:2024:NGI}
and thus are overfitted to a single or limited number of scenes.  To
circumvent the need to learn neural scene representations, image-space
neural rendering
systems~\cite{Liang:2025:DRN,Zeng:2024:RID,Nalbach:2017:DSC} take as
input G-buffers of intrinsic components of the scene, and output a
shaded image seen from the same viewpoint. Because the G-buffers only
capture a portion of the scene, image-space neural rendering methods
must necessarily hallucinate (or ignore) transport between visible and
non-visible parts of the scene.

Recently, a new class of neural rendering systems leverage attention
layers~\cite{Vaswani:2017:AIA} to model light transport between 3D
primitives.  Xu~\etal~\shortcite{Xu:2025:GLT} model diffuse light
transport in a point cloud representation of the scene.  Closest to
our method is RenderFormer~\cite{Zeng:2025:RFT} which employs a
two-stage transformer architecture that models the transport: (1)
between triangles and (2) from the triangles to the camera.  However,
RenderFormer employs a brute-force attention mechanism which does not
scale well to large triangle meshes.  Moreover, RenderFormer only
supports triangles as geometric primitives, diffuse (triangle-shaped)
light sources, and a per-triangle hard-coded GGX microfacet BRDF
model~\cite{Walter:2007:MMR}.  In contrast, \RF2 employs an efficient
sparse attention mechanism to support a large number ($> 100$k) of
primitives, and flexible geometry, lighting, and materials
representations and textures.

\paragraph{Long Context Modeling with Transformers} 
Classic transformers compute attention between all pair-wise token
combinations, resulting a quadratic complexity with respect to the
number of tokens in the sequence.  Moreover, when the sequence grows,
attention per-token tends to decrease and be spread over many tokens,
and as a consequence the transformer loses focus, resulting in a
decreased performance.  Addressing both issues is critical for scaling
a transformer-based rendering architecture beyond a few thousand
tokens.  Here, we focus on the most relevant classes of transformer
scaling methods, and refer to Tay~\etal~\cite{Tay:2022:ETS} for a
detailed overview.

Windowed attention
mechanisms~\cite{Liu:2021:STH,Yang:2023:SPT,Wu:2024:PT3,Beltagy:2020:LLD}
focus on addressing the compute complexity, and built on the
observation that in many cases proximity is a good indicator of
importance, and hence these mechanism hard-constrain the attention
computation to a small window around the target token. Consequently,
windowed attention mechanisms ignore long-range interactions which can
be important for light transport modeling.  More generally, windowed
attention mechanisms belong to a class of \emph{sparse} attention
methods that employ static attention
patterns~\cite{Li:2025:RAS,Parmar:2018:IT,Ho:2019:AAM} and their
effectiveness is highly dependent on whether the attention sparsity
matches the attention pattern. While light transport through a scene
can be sparse, it does not follow a pre-determined sparsity pattern.

Native-Sparse Attention (NSA)~\cite{Yuan:2025:NSA} dynamically
determines the sparseness by employing three different attention
streams: (i) a sliding window to capture local attention, (ii)
compressed attention that determines the importance of groups of input
tokens, and (iii) a fine-grained attention on the groups of tokens
identified as important. However, the computational cost of NSA is
significantly higher than static sparse attention patterns due to the
secondary retrieval stage.  Moreover, NSA requires Grouped-Query
Attention~\cite{Ainslie:2023:GQA} which lowers the model's capacity, and
thus adversely affects performance.

Hierarchical attention mechanisms
(\eg,~\cite{Yang:2021:FAL,Zhu:2021:HTF,Zhang:2021:MSV}) leverage the
observation that attention tends to be focused near the query, and
that the attention variation at distant tokens decreases.  Hence, by
creating a multi-resolution hierarchy of token and computing attention
with the token selected from the hierarchy based on distance,
attention can be better focused and more efficiently computed.
However, multi-resolution hierarchies implicitly assume that
positional distance is proportional to distance in the sequence or
image, and thus implicitly assume a (regular) uniform spatial
distribution of tokens.  This is not the case for 3D scenes, where
primitives are clustered at various points in space (\ie, objects).
Point Transformer v3~\cite{Wu:2024:PT3} addresses this limitation by
(i) serializing the point cloud along space-filling curves, and (ii)
grouping and padding to ensure the point cloud is divisible by the
target patch size. While, Point Transformer v3 improves speed and
memory overhead, its implementation is more complex and is
computationally more expensive than sparse attention methods.  We 
employ a less complex and resource intensive strategy for extending
the context window using a similar serialization strategy as Point
Transformer v3.

Xiao~\etal~\shortcite{Xiao:2024:ESL} observed that the soft-max
operation in the attention computation tends to \emph{'dump'} excess
attention in a single token (\ie, attention sink). A similar behavior
was also observed in vision transformers~\cite{Kang:2025:SWA}.  To
avoid attention being dumped in a random token,
Xiao~\etal~\shortcite{Xiao:2024:ESL} propose to keep a few dedicated
attention sink tokens to model global relations and a local sliding
windowed attention to model local relations. This local-global
dichotomy has been further refined in follow up
work~\cite{Zhang:2023:HHO,Munkhdalai:2024:LNC}.  We also build on this
idea, and introduce rendering-relevant semantics for the sinks.
First, we place all light sources in the sinks as these are likely to
interact with all surfaces. Second, inspired by the compressed tokens
in NSA~\cite{Yuan:2025:NSA}, we add to the sink summarization tokens
for groups of primitives based on Hilbert space-filling curves.

\section{Background - RenderFormer}
\label{sec:renderformer}

\RF2 builds and improves on RenderFormer~\cite{Zeng:2025:RFT}. We
therefore first review RenderFormer's architecture before detailing
\RF2.

RenderFormer is an end-to-end trained transformer-based neural
renderer that takes as input a sequence of triangles with GGX BRDF
parameters\,\cite{Walter:2007:MMR} and emittance strength, as well as
camera parameters, and it outputs a rendered image of the scene with
full global illumination.  RenderFormer consist of two stages with a
slightly different architecture.  The first (\ie, view-independent)
stage, consisting of $12$ self-attention
layers~\cite{Vaswani:2017:AIA}, transforms the input sequence of
embedded triangle tokens (expressed in \emph{world} coordinates) to a
sequence of per-triangle tokens which encode triangle-to-triangle
light transport.  The second stage (\ie, view-dependent stage),
consisting of 6 repetitions of a cross-attention
layer~\cite{Vaswani:2017:AIA} followed by a self-attention layer,
operates on view-bundle tokens. A view-bundle token is an embedding of
a $8 \times 8$ grid of camera rays expressed in the \emph{camera}
coordinate system.  The cross-attention layer computes the attention
between the view-bundle tokens and the transformed triangle tokens
from the first stage.  The view-dependent stage is followed by a dense
vision transformer to convert the transformed ray-bundle tokens into
pixel values for each ray.
The triangles are embedded as the sum of: (a) the per-vertex normal
embedding (using NeRF positional encoding with 6 frequencies that is
subsequently expanded to the $768$ token-length vector through a
linear layer), (b) the GGX BRDF parameters (expanded by a linear layer
to the $768$ token-length vector), and (c) the monochrome emittance
(expanded by a linear layer).  RenderFormer adapts
RoPE~\cite{Su:2024:RET} to apply a relative positional encoding on the
$9$D vector obtained by stacking the $3$D coordinates of the
triangle's vertices.  RoPE is applied at each layer in RenderFormer
with the vertices expressed in world coordinates in the
view-independent stage and in camera coordinates in the view-dependent
stage.
Hence, only the ray direction of the $8 \times 8$ camera rays is
embedded by stacking the $64$ view rays and subsequently expanded them
to a $768$ length ray-bundle embedding via a linear layer. %
RenderFormer is trained end-to-end, first at a $256 \times 256$
resolution and with scenes containing at most $1.5$k triangles
followed by a second training stage where the output resolution is
increased to $512 \times 512$ and the triangle count is increased to
$4$k.  RenderFormer is trained with a weighted $L_1$ and
LPIPS~\cite{Zhang:2018:TUE} loss on log-transformed reference renders.

\section{Overview}
\label{sec:overview}
\begin{figure}[tb]
    \centering
    \includegraphics[
        width=0.88\linewidth,
    ]{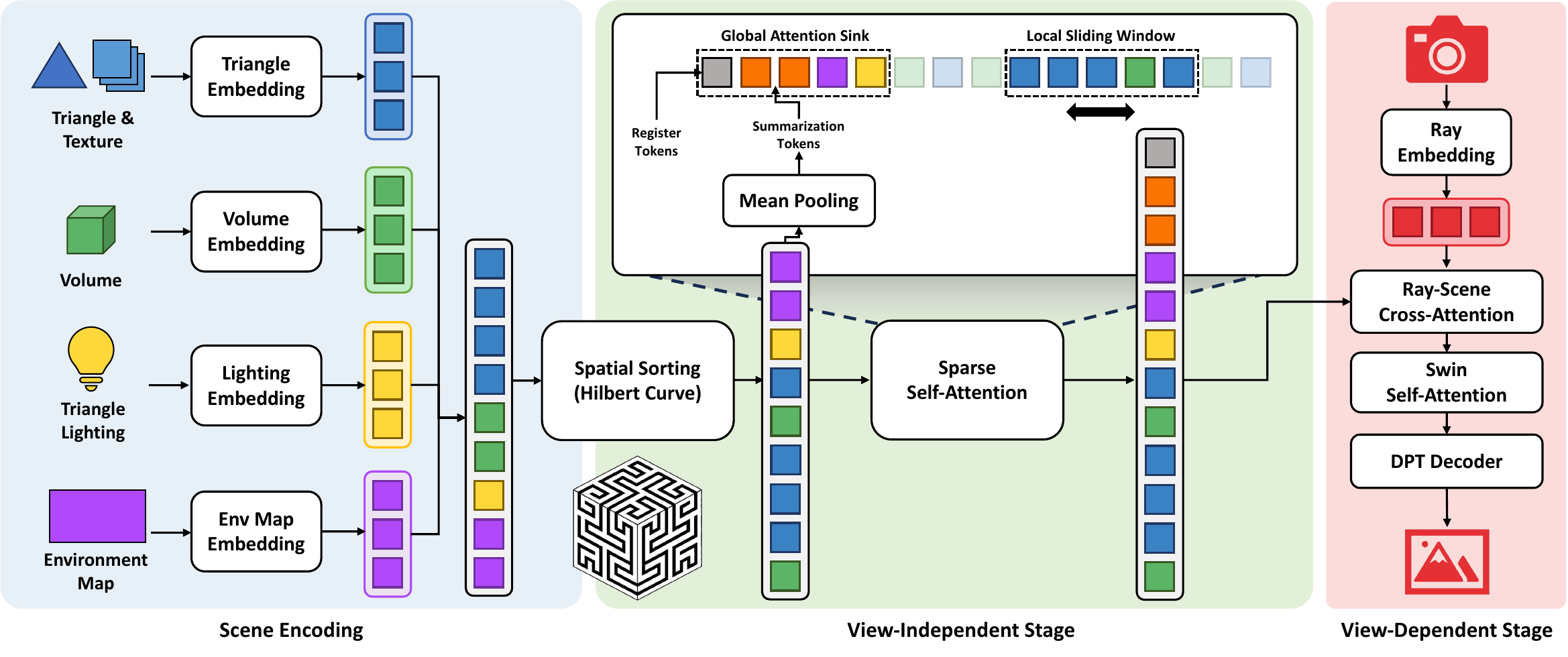}

    \caption{\RF2 Neural Rendering Pipeline.}
    \label{fig:pipeline}
\end{figure}

Similar to RenderFormer, \RF2 features a two-stage transformer-based
neural rendering pipeline where the first stage resolves
view-independent intra-primitive transport and the second stage
transforms view-dependent ray-bundles to output tokens based on the
transformed scene primitives from the first stage.  However, \RF2
deviates from RenderFormer is a number of critical steps: (i) \RF2 is
not limited to only triangle tokens and it supports a mixture of
different scene primitives, including different types of light sources
(\autoref{sec:embedding}), and (ii) \RF2 scales better in terms of
efficiency and accuracy to a larger number of scene primitives
(\autoref{sec:architecture}). \autoref{fig:pipeline} summarizes the
\RF2 pipeline.


\section{Scene Embedding}
\label{sec:embedding}
We represent a virtual scene as a sequence of heterogeneous tokens
that encode geometry, material, lighting, and camera information.  In
contrast to RenderFormer where the positional encoding is tailored to
triangles as scene primitives, and which relies on a hard-coded
camera-transformation to encode the camera position, we employ a
uniform relative positional encoding strategy for all tokens
(including ray-bundles):
\begin{compactenum}
\item For \textbf{tokens representing a concept with a 3D spatial
    location} (\eg, geometric primitive or camera) we use RoPE to
  encode the centroid of the concept with a RoPE dimension of $40$
  (\ie, 20 frequencies).  RoPE ensures that the scene embedding is
  invariant to scene translations.
\item For \textbf{tokens representing positionless concepts} (\eg,
  environment map) we employ RoPE using the centroid of the whole
  scene to ensure that translating the scene does not affect the
  relative attention computation between tokens from both categories.
\end{compactenum}
As the different concepts are defined by different parameters, we
employ a separate embedding for each token type, and rely on the
training process to enable \RF2 to differentiate between the
different primitive embeddings.

\subsection{Triangle \& Material Embedding}
To embed a triangle, we first embed the different components
(vertices, normals, and materials) and combine them via addition into
the final token.

\paragraph{Vertex Embedding}
We stack the 3 positions of vertices (minus the centroid of the
triangle) in a $9$D vector, and apply (NeRF) positional
encoding~\cite{Mildenhall:2021:NRS} with $12$ frequencies
exponentially spaced between $2^0$ and $2^{11}$.  Finally, we apply a
(trainable) linear layer to expand to a token-length (\ie, $768$)
vector followed by RMS-normalization.

\paragraph{Normal Embedding} We apply the same process as for vertex
embedding to encode the per-vertex normals. Note, the vertex and
normal embedding use separately trainable linear layers for
expansion.

\begin{figure}[t!]
\input{figures/tsne-material-latent}%
\hfill %
\input{figures/measured-brdf.tex}
\end{figure}

\paragraph{Material Embedding}
We desire an embedding of material appearance that is not tied to a
particular BRDF model. Inspired by prior work on learning a latent
embedding for
BRDFs~\cite{Sztrajman:2021:NBR,Hu:2020:DBA,Zheng:2021:CRM,Guo:2023:MAM,Gokbudak:2024:HGB,Serrano:2021:TES},
we also learn a material appearance embedding.  A key advantage of
\RF2's transformer architecture is that it is powerful enough to
directly learn how to evaluate the embedded material appearance (given
the view and lighting) without the need to rely on a pretrained
reverse mapping from latent code to material appearance.  As we are
interested in encoding the appearance rather than the exact BRDF, we
follow an encoding inspired by
Serrano~\etal's~\shortcite{Serrano:2021:TES} perceptual material
similarity metric and embed rendered images of a sphere under the
Uffizi Gallery light probe. We opt for a sphere for it simplicity and
the Uffizi Gallery light probe because it is color neural and it
contains a good mix of low and high frequency lighting
features~\cite{Bieron:2020:ABF}.  Practically, we employ a CNN-based
auto-encoder with a $9$D latent feature vector at the bottleneck.  We
pretrain this encoder with an L1 loss on images rendered with Blender
Cycles of randomly generated materials with the Principled BRDF
model~\cite{Burley:2012:PBS}. Furthermore, to encourage a coherent
manifold, we apply a smoothness regularization
term~\cite{Gao:2019:DIR}, and a $\tanh$ activation to constrain the
values in the embedding to $[-1,+1]$.
\autoref{fig:tsne_latent_material} visualizes the learned latent
material appearance space. While we currently use the Principled BRDF
model for generating training data, this can easily be extended to
include other analytical BRDF models or measured BRDFs.  For
efficiency, we also train an additional MLP for each analytical BRDF
model (after the latent space is trained) to map its parameters
directly into the latent space to bypass the need to render a sphere;
for measured BRDFs we render the material and use the pretrained
encoder.

\paragraph{Texture Embedding}
To support spatially varying materials, we embed all materials in a
texture. For each triangle, we first project the BRDF parameters into
the learned latent space and subsequently rasterize the per-triangle
texture ($9$ channels), local normal map ($3$ channels), and a
displacement map (as a $1$ channel height offset) in $32 \times 32$
image patches, which we subsequently encode with a pretrained
VAE~\cite{Wu:2025:QwenImage} into an $4 \times 4$ $80$-channel latent
feature map.  Finally, we compress the feature map via a single
learnable linear layer to a $768$-length vector and add it to the
token embedding.

\subsection{Voxel Embedding}
To demonstrate \RF2's ability to handle heterogenous geometric
primitives, we also encode voxels filled with a scattering medium into
a separate token-type.

\paragraph{Rotation and Scale Embedding}
For each voxel we encode the rotation matrix and scale vector that
describes the voxel's relative rotation and per-axis scale in world
coordinates.  Similar to the normal encoding for triangles, we employ
(NeRF) positional encoding with $12$ frequencies, which is
subsequently expanded via a linear layer to a token-length vector.

\paragraph{Scattering and Absorption}
As (RGB) scattering and (RGB) absorption coefficients are optical
parameters of scattering media, and thus model independent, we opt to
directly encode them. In addition, we also encode the anisotropy
coefficient of the scattering function, yielding a $7$D vector.  To
support spatially-varying scattering, we encode a
$4 \times 4 \times 4$ volumetric texture of the $7$D scattering
feature vector, and linearly project the feature vector into a
token-length vector that is added to the rotation and scale embedding.

\subsection{Triangular Light Source Embedding}
Similar to RenderFormer, we embed triangular light sources with
homogeneous diffuse emittance.  In contrast to RenderFormer, we store
colored RGB emittance as an explicit light source token (instead of
combining it with geometry tokens) which is expanded via a separate
linear layer to the token-length.  Similar to the triangle primitives,
we add the vertex positions and per-vertex normals embedding to the
token.

\subsection{Environment Lighting Embedding}
We follow an environment encoding similar to
DiffusionRenderer~\cite{Liang:2025:DRN}.

\paragraph{Texture Embedding}
We store both an LDR (clamped to $[0,1]$) and (log-encoded) HDR
version (normalized by the log maximum value) of the environment map
at $512 \times 256$ resolution, and encode each map using a pretrained
VAE~\cite{Wu:2025:QwenImage} yielding a $64 \times 32 \times 16$
latent feature map for each.  This feature map is too large to store
in a single token.  Hence, we opt to split the latent feature map in
$8 \times 4$ patches (of size $8 \times 8 \times 16$) that are
compressed by a linear layer into two token-length vectors; hence
yielding separate LDR and HDR environment map tokens.

\paragraph{Direction Embedding}
While a environment map token does not have a position, each pixel in
the $64 \times 64$ pixel patch does correspond to a lighting
direction.  To make \RF2 aware of the exact bundle of rays that
correspond to the pixels in the patch per token, we create a direction
map that, for each pixel, stores the corresponding (normalized) 3D
direction vector in world coordinates.  This direction map is
projected via a linear layer to a token-length vector and added to the
environment lighting embedding.

\paragraph{Lighting Strength Embedding}
During texture embedding we normalized the log-encoded HDR by the log
maximum value.  To retain this information, we expand this value to a
token-length vector and add it to the final embedding.

\subsection{Camera Embedding}
Similar to RenderFormer, we embed the (normalized) ray directions in
an $8 \times 8$ map, and project it to a token-length vector using a
single linear layer.  Whereas in RenderFormer the ray directions are
expressed in camera coordinates, we encode them directly in world
coordinates; the origin of the ray bundle is already taken care of via
the uniform relative positional encoding strategy.

\section{\RF2 Architecture}
\label{sec:architecture}
While \RF2 follows RenderFormer's two stage design, each stage
differs in how attention is computed.  We first discuss the
modification to the second stage (\ie, view-dependent), follow by the
more significant modifications to the first stage (\ie,
view-independent).

\subsection{View-dependent Stage}
Unlike RenderFormer's view-dependent stage, \RF2's
view-de\-pen\-dent stage operates in world coordinates (rather than
camera coordinates).  Moreover, we replace the full self-attention
layers by SWIN windowed attention layers~\cite{Liu:2021:STH} with a
window size of $8$ and a shift of $4$.  While the computation cost
changes modestly from $\mathcal{O}(T \times R + R^2)$ to
$\mathcal{O}(T \times R + W^2)$ ($T$ is the number of scene tokens
(the dominating factor), $R$ the number of ray-bundles, and $W$ the
window size), its main advantage is the ability to scale to larger
resolutions more easily because the context window size is now
resolution independent.

\subsection{View-independent Stage}
The view-independent stage is the main bottle neck when increasing the
number of tokens as computation cost scales quadratically with the
number of primitives. Moreover, when the sequence grows large, naive
attention tends to loose focus, resulting in a loss of render
fidelity. While often the dominating factor, primitive-to-primitive
light transport is not solely a local phenomena; a subset of global
factors such as illumination and large-scale occlusion can
significantly affect light transport through the scene.  Hence, we
combine attention from local primitives with attention from global
tokens, while mitigating focus-loss on large sequences.

\paragraph{Sorting \& Serialization}
Unlike ray-bundles, scene primitives are not regularly spaced, making
it more challenging to exploit locality while retaining efficient
GPU-computation.  Inspired by Point Transformer v3~\cite{Wu:2024:PT3},
we sort and linearize geometry tokens using Hilbert curves such that
nearby (in the 1D sequence) tokens are likely close in 3D space too.
This allows us to model local attention with sliding window
attention~\cite{Beltagy:2020:LLD}; we take $256$ tokens before and
after the target token, yielding a computational complexity
independent of the number of primitives.

\paragraph{Attention Sinks for Rendering}
To model global light transport, we adapt attention
sinks~\cite{Xiao:2024:ESL}. The tokens in the attention sink are
always included in the attention calculations.  We strategically
assign three different token types with rendering relevant semantics
to the sink:
\begin{compactenum}
\item \emph{Global Register Tokens}: we add $16$ register
  tokens~\cite{Darcet:2024:VTN} to the input sequence for storing
  storing global information.
\item \emph{Light Source Tokens}: it is likely that the light
  transport on most geometric primitives is directly affected by the
  light sources in the scene.  Hence, by placing the light sources in
  the sink we ensure that each light source is taken in account for
  each primitive regardless of distance. Conceptually, the attention
  computation with respect to the light sources is analogous to
  importance sampling the light sources in path tracing.
\item \emph{Summarization Tokens}: while long range light transport is
  important, we argue that the precise details of distant geometry
  matter less.  Therefore, we model long-range interactions with a
  coarser geometry representation.  Specifically, we perform mean
  pooling over every $64$ consecutive geometry tokens to form a
  summarization token that we add to the attention sink.
\end{compactenum}

\section{Training}
\label{sec:training}

\paragraph{Training Data}
Similarly to RenderFormer, we generate scenes by placing $1$ to $3$
randomly selected objects from the ObjaVerse dataset in one of four
randomly selected template scenes (a ground plane with one, two or
three walls).  Different from RenderFormer's scene generation process,
we assign randomly generated materials (using the Principled BRDF
model~\cite{Burley:2012:PBS} mapped into our material appearance
latent) from the following five categories: (a) homogeneous opaque
diffuse+specular material ($2/9$ chance; with the sum of the diffuse
and specular albedo restricted to $[0.9, 1]$, and roughness
log-sampled in $[0.01, 1]$), (b) a homogeneous opaque material with
metallic+rough\-ness parameters ($2/9$), (c) a diffuse+specular SVBRDF
randomly sampled from the MatSynth SVBRDF
dataset~\cite{Vecchio:2024:MMP} ($1/9$), (d) a metallic+rough\-ness
SVBRDF from MatSynth ($1/9$), or (e) a homogeneous transparent
metallic+roughness material ($3/9$; with metallic sampled in $[0,1]$,
roughness log-sampled in $[0.01, 1]$, and each RGB channel of the base
color sampled in $[0,1]$). In $2/3$ of the scenes, we add one (1/2) or
two (1/2) volumes with randomly chosen scattering and absorption.  The
camera is randomly placed and aimed at the scene with a FOV chosen
between $30^\circ$ and $60^\circ$. Unlike RenderFormer, we allow the
camera to be placed inside the scene.  Up to $8$ triangular light
sources are randomly placed outside the scene with a random (RGB)
intensity between $2,\!500$ and $5,\!000$ $W/ m^2$.  Furthermore, for
$5/6$ of the scenes, we add environment lighting randomly selected
from PolyHaven. To avoid color bias, we randomly swap the color
channels in the environment map.  Each scene is rendered offline using
Blender Cycles with $4,\!096$ samples per pixel. For efficiency, we
pre-generate a training dataset of $\sim\!10$M randomly sampled scenes
spanning resolutions $256^2$--$2048^2$ with $1$k--$64$k primitives,
totaling $\sim\!70$TB.

\paragraph{Loss Function} We employ the same loss as RenderFormer:
\begin{equation}
  L_1(\log I) + 0.05 \ L_{LPIPS}(clamp(\log I / \log 2, 0, 1)),
\end{equation}
where the log encoding of the image $I$ serves to avoid specular
reflections dominating the L1 error, and the LPIPS
loss~\cite{Zhang:2018:TUE} minimizes perceptual differences. 

\paragraph{Training Process \& Refinement} We employ a five-stage
training regime to first focus on learning the principles of light
transport, before adding scene-complexity:
\begin{compactitem}
\item In stage 1, we decimate the generated scenes to $1$k primitives
  and render the scene at $256 \times 256$ resolution.  To prioritize
  learning accurate coarse-scale transport, we employ full-attention
  at this stage.  Furthermore, we gradually increase the complexity of
  the scene by first training exclusively on homogeneous materials
  ($1$ day on $32\times$ A100 GPUs). Next we include SVBRDFs ($1$ additional
  day), followed by the inclusion of environment lighting ($1$ day; to
  focus training on the lighting effects, we mask out pixels that
  directly see the environment map), and finally adding volumetric
  objects ($1$ day).
\item In stage 2 we increase the primitive budget to $4$k as well as
  the resolution to $512 \times 512$, and continue to train for $2$
  days on the same setup.
\item In stage 3, we keep the scene parameters the same, but switch
  from full-attention to our combined attention sink and windowed
  attention ($3$ days).
\item In stage 4, we increase the primitive budget to $16$k ($1$
  week).
\item Finally, in stage 5 we increase the primitive budget to $64$k as
  well as the resolution to $2048 \times 2048$ for another $3$ days of
  training.
\end{compactitem}
Yielding a total training time of $19$ days on $32\times$ A100 GPUs.
While the training cost is significant, we emphasize that once
pretrained, no fine-tuning or training is needed for rendering a new
scene.

\begin{figure}[tb]
    \centering
    \begin{subfigure}{0.25\textwidth}
        \centering
        \includegraphics[
            width=\linewidth,
            height=\linewidth
        ]{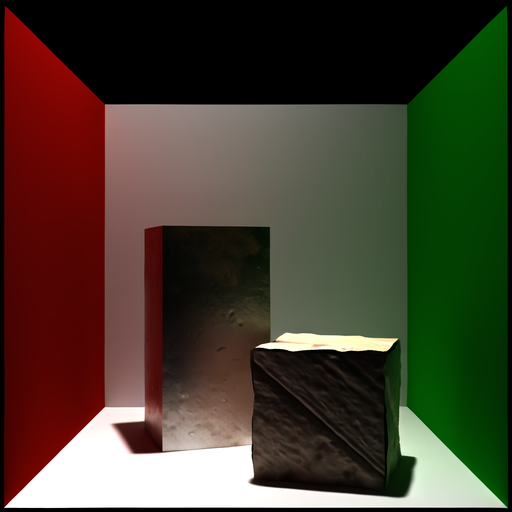}
    \end{subfigure}\hfill
    \begin{subfigure}{0.25\textwidth}
        \centering
        \includegraphics[
            width=\linewidth,
            height=\linewidth
        ]{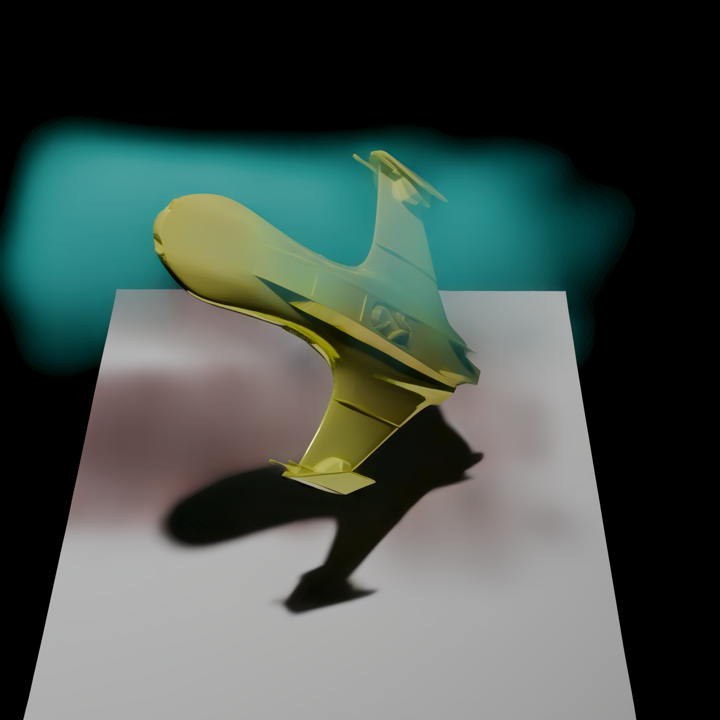}
    \end{subfigure}\hfill
    \begin{subfigure}{0.25\textwidth}
        \centering
        \includegraphics[
            width=\linewidth,
            height=\linewidth
        ]{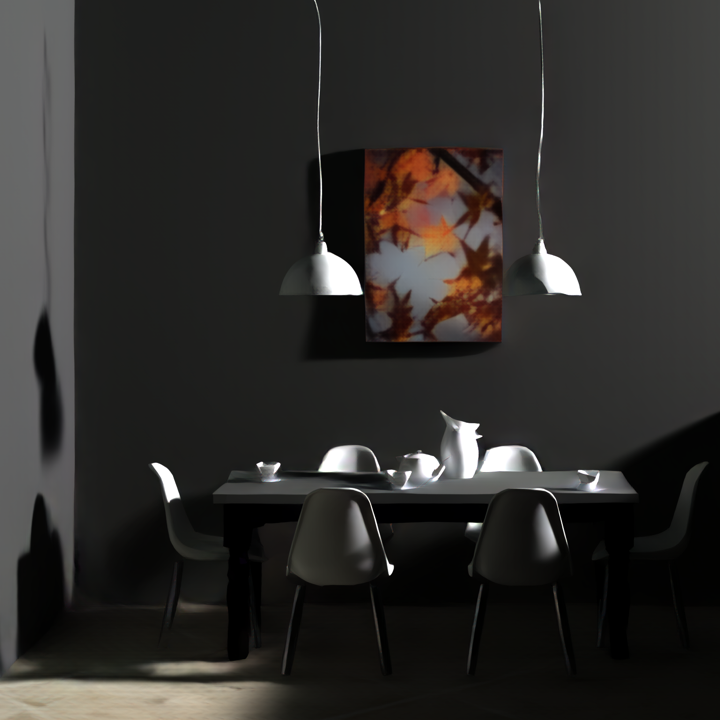}
    \end{subfigure}\hfill
    \begin{subfigure}{0.25\textwidth}
        \centering
        \includegraphics[
            width=\linewidth,
            height=\linewidth
        ]{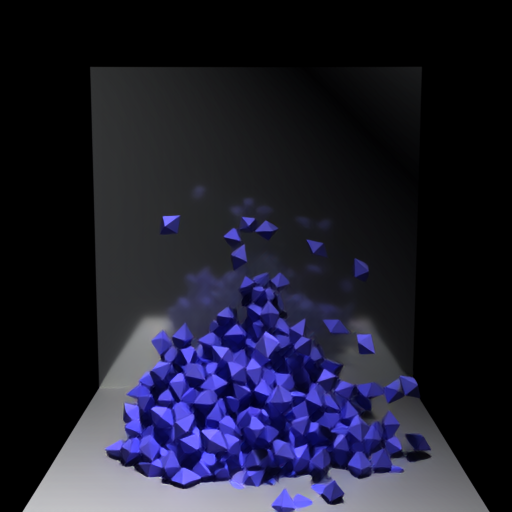}
    \end{subfigure}

    \caption{Additional results demonstrating \RF2's ability to
      handle displacement mapping and volumetric scattering without
      the need for specialized code, and two complex scenes with a
      large number of primitives.}
    \label{fig:more_results}
\end{figure}

\section{Results}
\paragraph{Capabilities}
Figures~\ref{fig:teaser} and~\ref{fig:more_results} demonstrate the
capabilities of \RF2 on a wide variety of
scenes. \autoref{fig:teaser} demonstrates that \RF2 can render,
with full global light transport, scenes containing refractive
surfaces (1st) including caustics, environment lighting (2nd),
volumetric scattering (3rd), and textures
(4th). \autoref{fig:more_results} demonstrates displacement mapping
(1st), volumetric scattering (2nd), and complex scenes with a large
number of primitives (3rd and 4th).  Prior neural rendering methods
either require per-scene fine-tuning and/or cannot support all of
these effects.  Compared to path tracing, \RF2 does not require
specialized code to handle displacement mapping or volumetric
scattering, and it can learn such effects solely by example.

\RF2 employs a flexible material appearance latent space for
assigning material properties to surface in the scene.
\autoref{fig:measured_brdf} demonstrates that, even though our latent
space is trained on materials modeled with the Disney principled BRDF
model~\cite{Burley:2012:PBS}, it can also model materials not part of
the training set (\eg, in this case selected measured materials from
the RGL BRDF dataset~\cite{Dupuy:2018:APE}).  Moreover, as our latent
space is based on encoding rendered images, it can easily be retrained
to encompass material appearances currently not covered (\eg,
anisotropic and color changing materials).

\begin{figure}[t]
\centering

\begin{minipage}[t]{0.31\linewidth}
\centering
\begin{tikzpicture}
\begin{axis}[
    width=1.15\linewidth,
    height=5cm,
    xmode=log,
    log basis x=2,
    xtick={4096,8192,16384,32768,65536,131072},
    xticklabels={,8K,,32K,,128K},
    grid=both,
    major grid style={opacity=0.35},
    minor grid style={opacity=0.15},
    xlabel={\#Triangles},
    ylabel={LPIPS $\downarrow$},
    ymin=0, ymax=0.7,
    label style={font=\scriptsize},
    tick label style={font=\scriptsize},
    legend style={
        draw=none,
        font=\scriptsize,
        at={(0,1)},
        anchor=north west,
        row sep=-0.05pt,
        nodes={scale=0.8},
    },
    legend image post style={scale=0.75},
    legend cell align=left,
]

\addplot+[mark=o, thick] coordinates {
    (4096,0.0198)
    (8192,0.0305)
    (16384,0.0502)
    (32768,0.0729)
    (65536,0.0873)
    (131072,0.0999)
};
\addlegendentry{RF-V2 (ours)}

\addplot+[mark=square, thick] coordinates {
    (4096,0.0244)
    (8192,0.0358)
    (16384,0.0604)
    (32768,0.1500)
    (65536,0.3070)
    (131072,0.4772)
};
\addlegendentry{RF-V1}

\end{axis}
\end{tikzpicture}
\vspace{-0.5cm}
\captionof{figure}{Render quality degradation for increasing primitive
  budget.}
\label{fig:lpips_tricount_small}
\end{minipage}
\hfill
\begin{minipage}[t]{0.31\linewidth}
\centering
\begin{tikzpicture}
\begin{axis}[
    width=1.15\linewidth,
    height=5cm,
    xmode=log,
    log basis x=10,
    xmin=1500, xmax=140000,
    grid=both,
    major grid style={opacity=0.35},
    minor grid style={opacity=0.15},
    xlabel={\#Triangles},
    ylabel={Time (s/frame) $\downarrow$},
    ymax=7.9,
    label style={font=\scriptsize},
    tick label style={font=\scriptsize},
    xtick={1597,6397,21757,127233},
    xticklabels={1.6K,6.4K,22K,128K},
    legend style={
        draw=none,
        font=\scriptsize,
        at={(0,1)},
        anchor=north west,
        row sep=-0.05pt,
        nodes={scale=0.8},
    },
    legend image post style={scale=0.75},
    legend cell align=left,
]

\addplot+[mark=o, thick] coordinates {
    (1597,0.0468)
    (2557,0.0502)
    (6397,0.0687)
    (12277,0.1118)
    (21757,0.18)
    (63277,0.5382)
    (83197,0.6821)
    (127233,1.1212)
};
\addlegendentry{RF-V2 (ours)}

\addplot+[mark=square, thick] coordinates {
    (1597,0.056921)
    (2557,0.060798)
    (6397,0.081398)
    (12277,0.142014)
    (21757,0.258925)
    (63277,1.198455)
    (83197,1.88105)
    (127233,3.922935)
};
\addlegendentry{RF-V1}

\addplot+[mark=triangle, thick] coordinates {
    (1597,3.634)
    (2557,3.762)
    (6397,4.002)
    (12277,4.238)
    (21757,4.374)
    (63277,4.857)
    (83197,4.954)
    (127233,5.248)
};
\addlegendentry{Cycles}

\end{axis}
\end{tikzpicture}
\vspace{-0.5cm}
\captionof{figure}{Runtime scaling versus mesh complexity.}
\label{fig:second_metric_tricount}
\end{minipage}
\hfill
\begin{minipage}[t]{0.31\linewidth}
\centering
\begin{tikzpicture}
\begin{axis}[
    width=1.15\linewidth,
    height=5cm,
    xmode=log,
    log basis x=2,
    xmin=460, xmax=4500,
    ymin=0.4, ymax=299,
    xlabel={Resolution ($N \times N$)},
    ylabel={Time (s/frame) $\downarrow$},
    label style={font=\scriptsize},
    tick label style={font=\scriptsize},
    grid=both,
    major grid style={opacity=0.35},
    minor grid style={opacity=0.15},
    xtick={512, 1024, 2048, 4096},
    xticklabels={512, 1024, 2048, 4096},
    legend style={
        draw=none,
        font=\scriptsize,
        at={(0,1)},
        anchor=north west,
        row sep=-2pt,
        nodes={scale=0.8},
    },
    legend image post style={scale=0.7},
    legend cell align=left,
]
\addplot+[mark=o, thick] coordinates {
    (512, 0.525) (768, 0.791) (1024, 1.226) (1536, 2.325)
    (2048, 3.917) (3072, 8.500) (4096, 14.915)
};
\addlegendentry{RF-V2 (ours)}

\addplot+[mark=square, thick] coordinates {
    (512, 1.307) (768, 1.618) (1024, 2.196) (1536, 4.076)
    (2048, 7.775) (3072, 24.821) (4096, 65.056)
};
\addlegendentry{RF-V1}

\addplot+[mark=triangle, thick] coordinates {
    (512, 5.13) (768, 9.41) (1024, 15.07) (1536, 31.11)
    (2048, 53.75) (3072, 127.30) (4096, 216.80)
};
\addlegendentry{Cycles}

\end{axis}
\end{tikzpicture}
\vspace{-0.5cm}
\captionof{figure}{Resolution scaling on a $64$k-triangle scene.}
\label{fig:scaling}

\end{minipage}

\end{figure}


\begin{figure}[t]
    \captionsetup{subrefformat=empty}
    \captionsetup[subfigure]{labelformat=empty}
    \centering
    \begin{subfigure}{0.16666\textwidth}
        \centering
        \includegraphics[width=\linewidth,height=\linewidth]{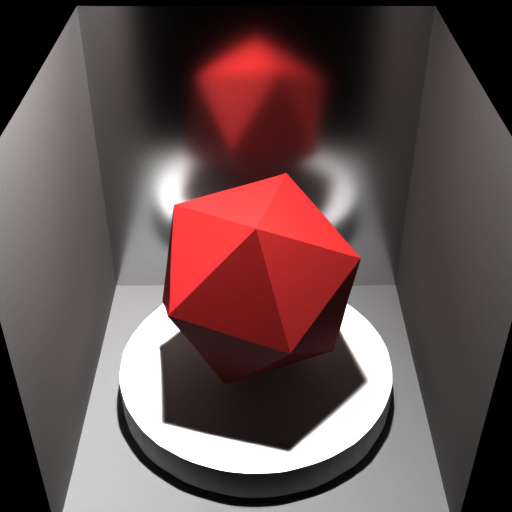}
        \caption{RF, 6.4K}
    \end{subfigure}\hfill
    \begin{subfigure}{0.16666\textwidth}
        \centering
        \includegraphics[width=\linewidth,height=\linewidth]{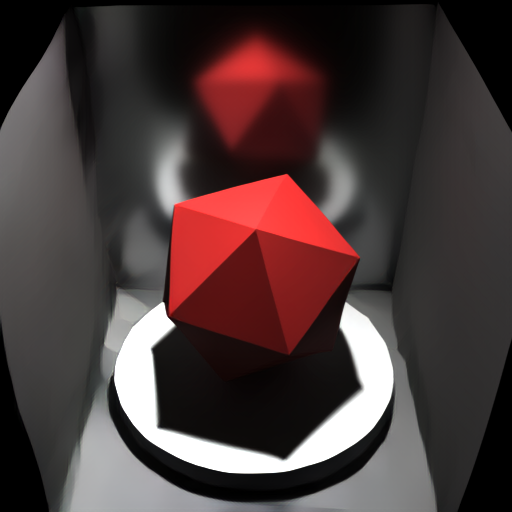}
        \caption{RF, 21.8K}
    \end{subfigure}\hfill
    \begin{subfigure}{0.16666\textwidth}
        \centering
        \includegraphics[width=\linewidth,height=\linewidth]{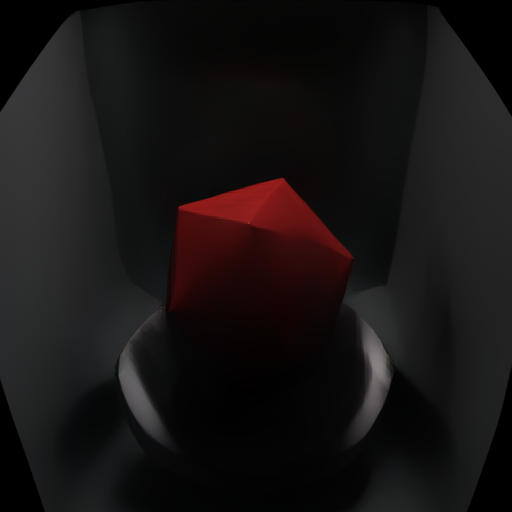}
        \caption{RF, 83.2K}
    \end{subfigure}\hfill
    \begin{subfigure}{0.16666\textwidth}
        \centering
        \includegraphics[width=\linewidth,height=\linewidth]{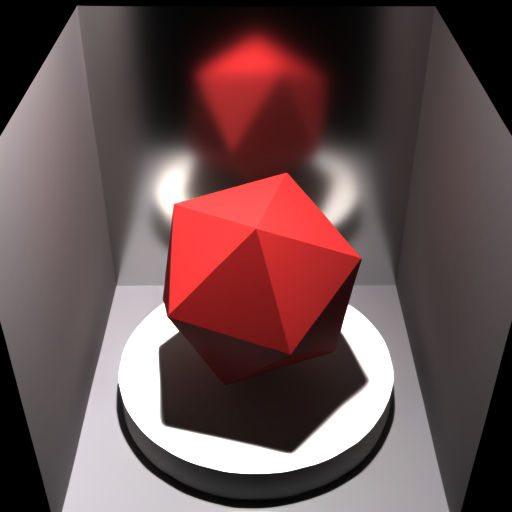}
        \caption{Ours, 6.4K}
    \end{subfigure}\hfill
    \begin{subfigure}{0.16666\textwidth}
        \centering
        \includegraphics[width=\linewidth,height=\linewidth]{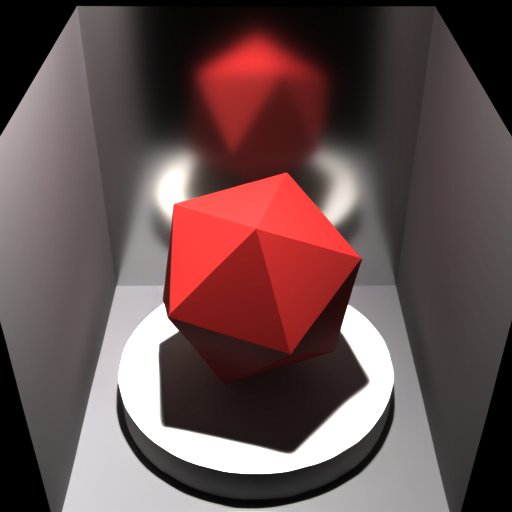}
        \caption{Ours, 21.8K}
    \end{subfigure}\hfill
    \begin{subfigure}{0.16666\textwidth}
        \centering
        \includegraphics[width=\linewidth,height=\linewidth]{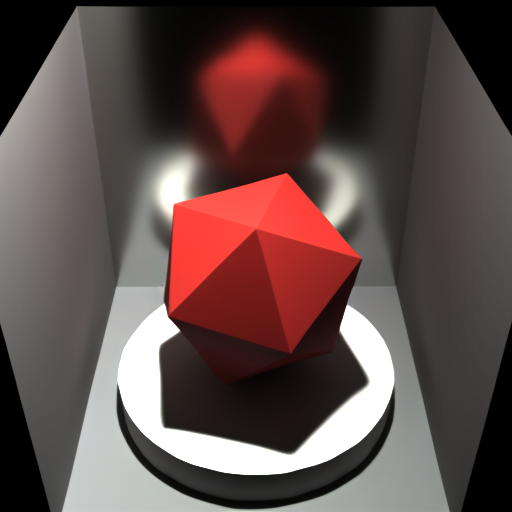}
        \caption{Ours, 83.2K}
    \end{subfigure}
    \vspace{-0.5cm}
    \caption{Qualitative comparison of render quality of RenderFormer
      vs. \RF2 for an increasing number of primitives.  At $83.2$K
      primitives, RenderFormer loses attentional focus, resulting in a
      darkened image.}
    \label{fig:seqlen_scaling}
\end{figure}


\begin{figure}[t!]
    \centering
    \begin{subfigure}{0.25\textwidth}
        \centering
        \includegraphics[
            width=\linewidth,
            height=\linewidth
        ]{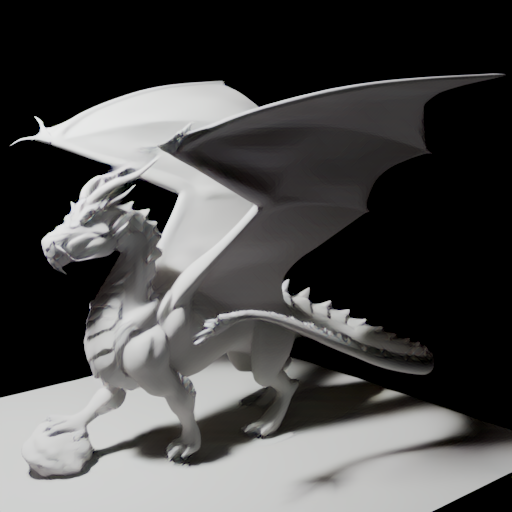}
    \end{subfigure}\hfill
    \begin{subfigure}{0.25\textwidth}
        \centering
        \includegraphics[
            width=\linewidth,
            height=\linewidth
        ]{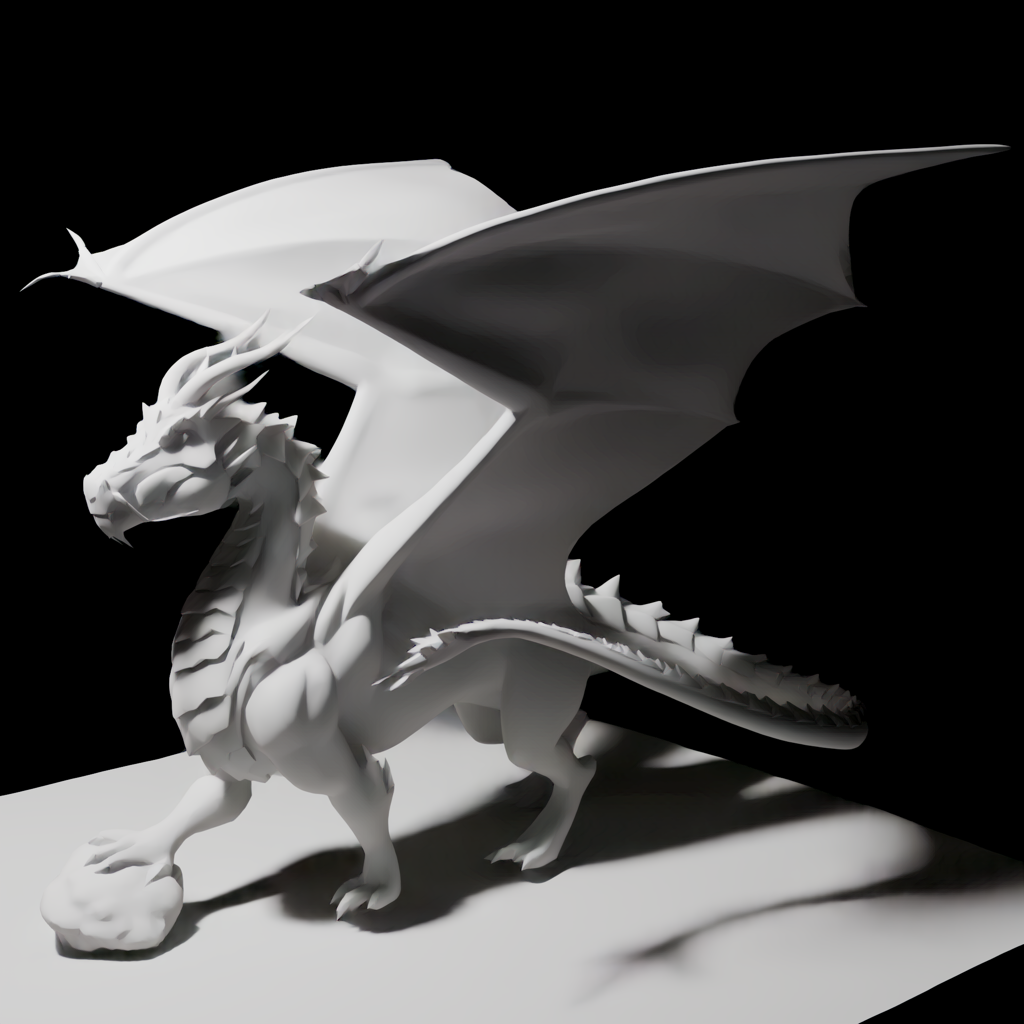}
    \end{subfigure}\hfill
    \begin{subfigure}{0.25\textwidth}
        \centering
        \includegraphics[
            width=\linewidth,
            height=\linewidth
        ]{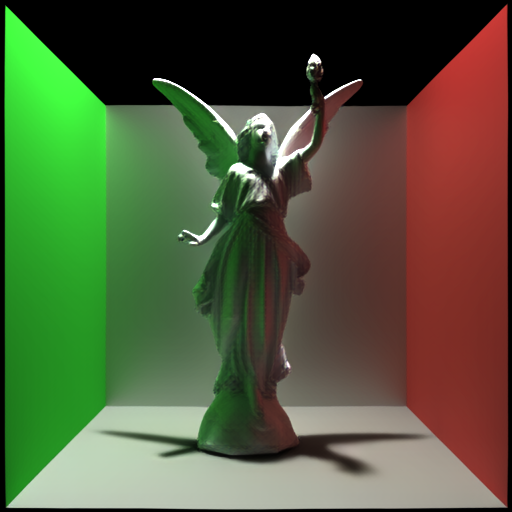}
    \end{subfigure}\hfill
    \begin{subfigure}{0.25\textwidth}
        \centering
        \includegraphics[
            width=\linewidth,
            height=\linewidth
        ]{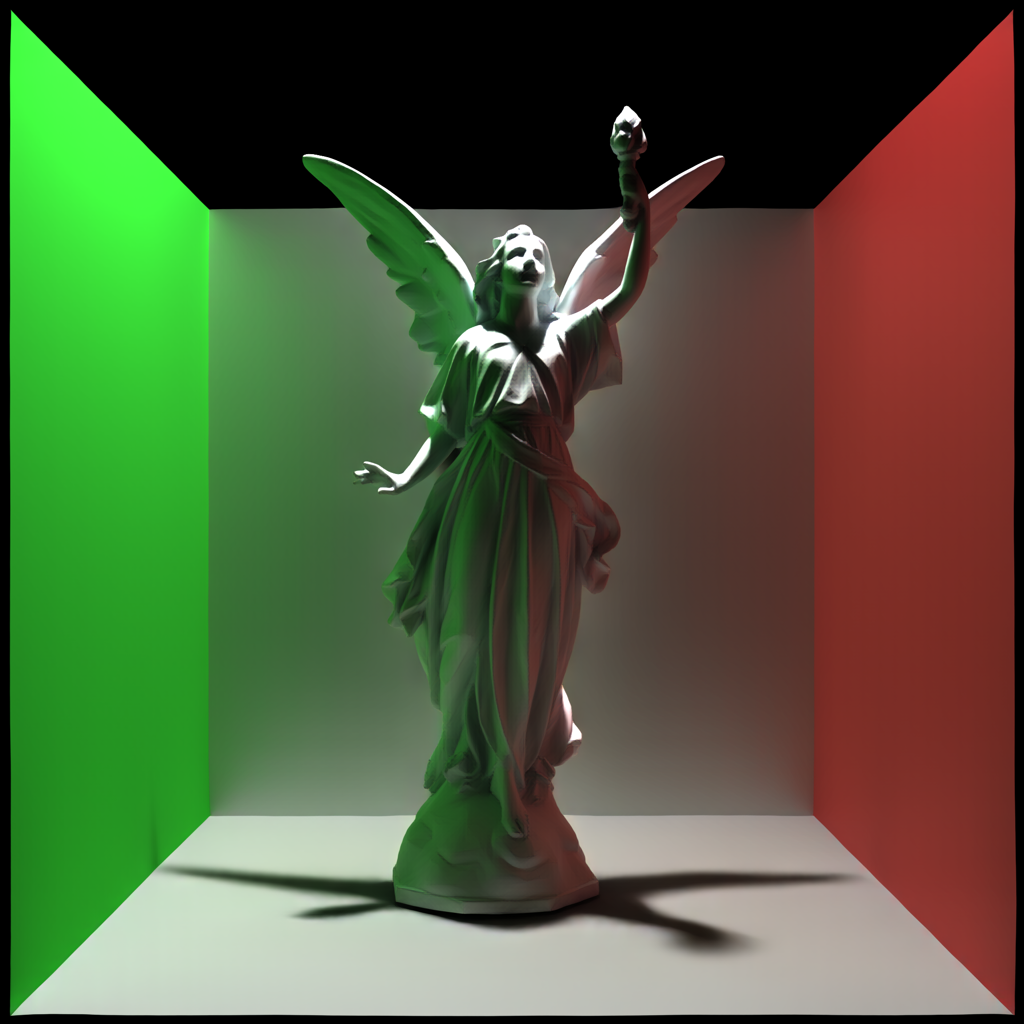}
    \end{subfigure}
    \vspace{-0.5cm}
    \caption{Rendering at higher resolutions also improves geometric
      details even for features larger than a pixel. \Eg, the claws on
      the dragon and Lucy's face and hands are better resolved at
      $2048$ (right) than at $512$ (left) resolution.}
    \label{fig:highres_scaling}
\end{figure}

\paragraph{Comparison to RenderFormer}
\RF2 is closely related to RenderFormer \cite{Zeng:2025:RFT}; both
use a similar two-stage transformer-based pipeline.  Compared to
RenderFormer, \RF2 achieves better accuracy when rendering scenes
with a large number of triangles thanks to its sparse attention
mechanism (\autoref{fig:lpips_tricount_small}).
\autoref{fig:seqlen_scaling} qualitatively compares render quality for
a scene with varying number of triangles; the darkening at $83.2$k
triangles when rendered with RenderFormer is a direct consequence of
loss of attention.  We used the publicly available version of
RenderFormer which is trained on $4k$ triangles; we found that
training RenderFormer for larger triangle meshes is unstable.
Moreover, \RF2 is also considerably more efficient as shown
in~\autoref{fig:second_metric_tricount} and~\autoref{fig:scaling}
thanks to the render-informed sparse attention in the view-independent
stage and the SWIN-attention in the view-dependent stage respectively
(all timings are measured on a single NVIDIA A100).  Empirically, we
found that rendering time is approximately equally distributed over
both stages (\ie, view-dependent vs. view-independent).  For
reference, we also include timings of Blender Cycles on the same
scenes using $4096$ adaptive samples per pixel (\ie, the same setting
as used for the training images).

\RF2 not only supports larger triangle meshes, it can also be more
easily fine-tuned for higher image
resolution. \autoref{fig:highres_scaling} compares two scenes rendered
at $512 \times 512$ and $2048 \times 2048$.  An interesting
observation is that despite both scenes containing the same number of
triangles, that at higher resolution \RF2 is able to more
faithfully render fine detailed geometry (\eg, the dragon's claws and
Lucy's face and hand).

\begin{table}[t]
  \caption{Sparse attention ablation with or without sliding windowed
    attention (SW) and attention sink (AS) with inclusion of light
    source tokens (L) and summary tokens (S), as well as varying
    number of summarization ratios. For each metric, the
    \colorbox{best}{best}, \colorbox{second}{second best}, and
    \colorbox{third}{third best} results are highlighted.}
    \centering
    \setlength{\tabcolsep}{6pt}
    \renewcommand{\arraystretch}{1.15}
    \begin{tabular}{lcccc}
    \specialrule{0.08em}{0pt}{0pt}
    \textbf{Model Variant} & \textbf{PSNR} $\uparrow$ & \textbf{SSIM} $\uparrow$ & \textbf{LPIPS} $\downarrow$ & \textbf{HDR-FLIP} $\downarrow$ \\
    \specialrule{0.05em}{0pt}{0pt}

    \textbf{\RF2}
        & \cellcolor{best}28.25
        & \cellcolor{best}0.8982
        & \cellcolor{best}0.0997
        & \cellcolor{best}0.4200 \\
    \specialrule{0.05em}{0pt}{0pt}

    w/o AS        & 27.36 & 0.8841 & 0.1247 & 0.4359 \\
    w/o SW        & 26.92 & 0.8715 & 0.1289 & 0.4517 \\
    \specialrule{0.05em}{0pt}{0pt}

    AS w/o L      & 27.40 & 0.8813 & 0.1081 & \cellcolor{third}0.4250 \\
    AS w/o S      & 27.09 & 0.8754 & 0.1210 & 0.4485 \\
    AS w/o L, S   & 26.13 & 0.8474 & 0.1514 & 0.4959 \\
    \specialrule{0.05em}{0pt}{0pt}

    More S (\#seq / 32)
        & \cellcolor{third}27.78
        & \cellcolor{third}0.8923
        & \cellcolor{third}0.1072
        & \cellcolor{second}0.4231 \\
    Less S (\#seq / 128)
        & 26.48 & 0.8690 & 0.1346 & 0.4655 \\
    Less S (\#seq / 256)
        & 26.93 & 0.8663 & 0.1316 & 0.4580 \\
    \specialrule{0.05em}{0pt}{0pt}

    Full Attention
        & \cellcolor{second}27.91
        & \cellcolor{second}0.8953
        & \cellcolor{second}0.1027
        & 0.4287 \\
    \specialrule{0.08em}{0pt}{0pt}
    \end{tabular}
    \label{tab:attention_ablation}
\end{table}


\begin{figure}[t]
    \captionsetup{subrefformat=empty}
    \captionsetup[subfigure]{labelformat=empty}
    \centering
    \begin{subfigure}{0.25\textwidth}
        \centering
        \includegraphics[
            width=\linewidth,
            height=\linewidth
        ]{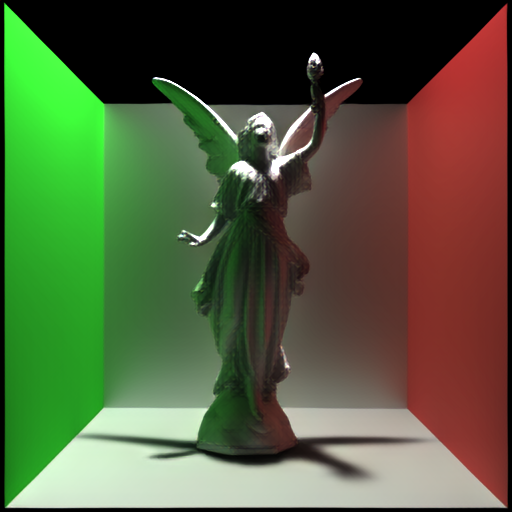}
        \caption{All components}
    \end{subfigure}\hfill
    \begin{subfigure}{0.25\textwidth}
        \centering
        \includegraphics[
            width=\linewidth,
            height=\linewidth
        ]{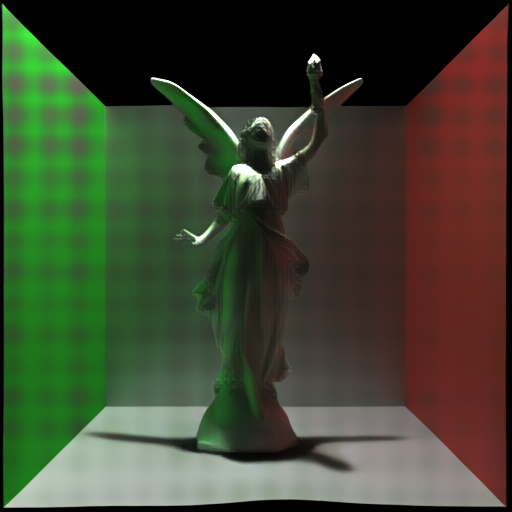}
        \caption{w/o AS}
    \end{subfigure}\hfill
    \begin{subfigure}{0.25\textwidth}
        \centering
        \includegraphics[
            width=\linewidth,
            height=\linewidth
        ]{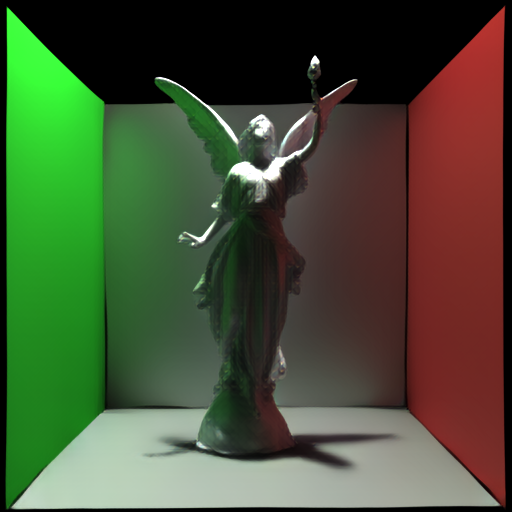}
        \caption{w/o SW}
    \end{subfigure}\hfill
    \begin{subfigure}{0.25\textwidth}
        \centering
        \includegraphics[
            width=\linewidth,
            height=\linewidth
        ]{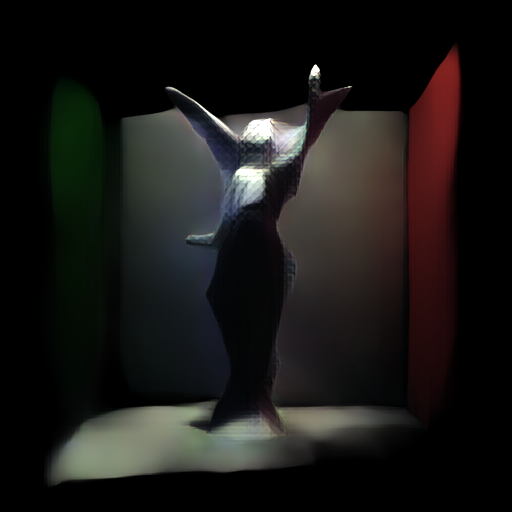}
        \caption{AS w/o L, S}
    \end{subfigure}
    \vspace{-0.5cm}
    \caption{A qualitative comparison of the ablation variants
      indicates that all components are essential to achieve the
      highest render quality.}
    \label{fig:ablation}
\end{figure}

\paragraph{Ablation}
We ablate the different components of \RF2's sparse attention
mechanism. \autoref{tab:attention_ablation} shows average PSNR,
SSIM~\cite{Wang:2004:IQA}, LPIPS~\cite{Zhang:2018:TUE}, and
HDR-FLIP~\cite{Andersson:2020:FDE} over randomly generated scenes
(using the same distribution as the training data, but with a
different set of environment maps, SVBRDFs, and shapes than used for
training) for different ablation variants: without sliding windowed
attention, without attention sink, and leaving out lighting and/or
summarization token from the attention sink -- each ablation variant
is trained up to stage 4 ($16$k primitives). Only when all components
are included, we achieve the highest accuracy in rendering.
Surprisingly, our sparse attention outperforms full-attention, which
we attribute to the full-attention model having to distribute its
attention over too many tokens (\ie, loss of focus).
\autoref{fig:ablation} further qualitatively demonstrates the
difference in quality between the different ablation variants.
We refer to the supplemental material for additional results and
comparisons.

\paragraph{Limitations}
While \RF2 addresses many shortcomings of RenderFormer, it is not
without limitations.  First, \RF2 is still limited to maximum $8$
light sources per scene, a constraint inherited from its training
data.  However, we argue that more complex lighting conditions are
more efficiently modeled with the addition of an environment map.
Furthermore, similar to RenderFormer, our model is trained on single
frames, and it does not explicitely enforce temporal coherence.  While
\RF2 supports textures/SVBRDFs, the resolution is fixed per triangle
primitive ($32 \times 32$). Consequently, significant texture-quality
degradation can occur for large triangles.  However, this can easily
be resolved by subdividing triangles to impose a maximum triangle
size. By supporting heterogeneous primitives and using a material
appearance parameterization independent of a hard-coded BRDF model,
\RF2 can easily be extended to support new primitives.  However,
training is most effective when new primitives are added in the first
training stage. Consequently, adding a new primitive typically
requires significant retraining.  Improving extensibility without
requiring a full retraining is an interesting avenue for future
research.


\section{Conclusion}
In this paper we presented \RF2, a transformer-based neural
rendering model that takes as input a sequence of primitives (\ie,
textured triangles, volumetric elements, light sources, environment
maps, and camera) and outputs a rendered image of the scene with full
global illumination.  \RF2 employs a rendering-aware
sparse-attention to resolve intra-primitive light transport which
enables \RF2 to scale better to larger scenes, both for training as
well as inference.  Furthermore, we employ a flexible latent material
appearance space to specify surface reflectance. \RF2 is trained
end-to-end, thereby avoiding the need for specialized code to handle
displacement mapping or volumetric scattering. 

\section*{Acknowledgment}
Chong Zeng was supported by the Stanford Graduate Fellowship. This work was partially supported by the Brown Institute for Media Innovation at Stanford University.

\bibliographystyle{splncs04}
\bibliography{src/reference}
\end{document}


\title{\RF2: Neural Rendering with Heterogeneous Scene Primitives}

\titlerunning{\RF2}

\author{Chong Zeng\inst{1}\orcidlink{0009-0004-6373-6848} \and
  Yue Dong\inst{2}\orcidlink{0000-0003-0362-337X} \and
  Pieter Peers\inst{3}\orcidlink{0000-0001-7621-9808} \and
  Lvmin Zhang\inst{1}\orcidlink{0000-0003-3503-5791} \and
  Maneesh Agrawala\inst{1}\orcidlink{0000-0002-8996-7327}
  }

\authorrunning{C. Zeng et al.}

\institute{Stanford University, USA \and
  Microsoft Research, China \and
  College of William \& Mary, USA}

\maketitle

\appendix

\newcommand{\centmark}[1]{\raisebox{-0.75\height}{#1}} 

\section{Additional Implementation Details}

~\autoref{tab:supp_architecture} summarizes the key hyper-parameters
of \RF2's two-stage transformer pipeline. Both stages share the same
model dimension (768), number of attention heads (6), and Feed-Forward
Network (FFN) dimension (3072). The sparse attention in the
view-independent stage, consisting of a local sliding window (total
size 512) combined with global attention sink tokens, is implemented
using PyTorch's \texttt{flex\_attention}
API~\cite{dong2025flexattention}. We also employ
QK-Norm~\cite{henry2020querykeynormalizationtransformers} to stabilize
the attention mechanism. The complete \RF2 model comprises
approximately 207M parameters in total.

\begin{table}
  \caption{Architectural details of \RF2's two-stage transformer pipeline.}
  \centering
  \setlength{\tabcolsep}{10pt}
  \renewcommand{\arraystretch}{1.15}
  \begin{tabular}{c|cc}
    \specialrule{0.08em}{0pt}{0pt}
    & \textbf{View-indep.} & \textbf{View-dep.} \\
    \specialrule{0.05em}{0pt}{0pt}
    Layers & 12 & 6 \\
    Model Dimension & 768 & 768 \\
    Attention Heads & 6 & 6 \\
    Attention Type & Sparse Self-Attn. & Cross-Attn. + Swin Self-Attn. \\
    FFN Dimension & 3072 & 3072 \\
    FFN Activation & \multicolumn{2}{c}{SwiGLU} \\
    Normalization & \multicolumn{2}{c}{RMSNorm} \\
    \specialrule{0.08em}{0pt}{0pt}
  \end{tabular}
  \label{tab:supp_architecture}
\end{table}

\begin{figure}[t]
\centering
\setlength{\tabcolsep}{0pt}
\renewcommand{\arraystretch}{0}
\begin{tabular}{c c c c c}
\textbf{\RF2} & \textbf{Reference} & \textbf{Diff ($\times 5$)} & \textbf{FLIP} & \textbf{Metrics} \\
\centmark{\includegraphics[width=0.2\linewidth]{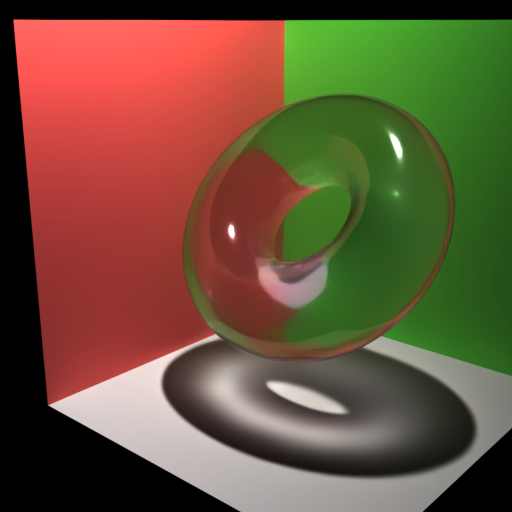}} &
\centmark{\includegraphics[width=0.2\linewidth]{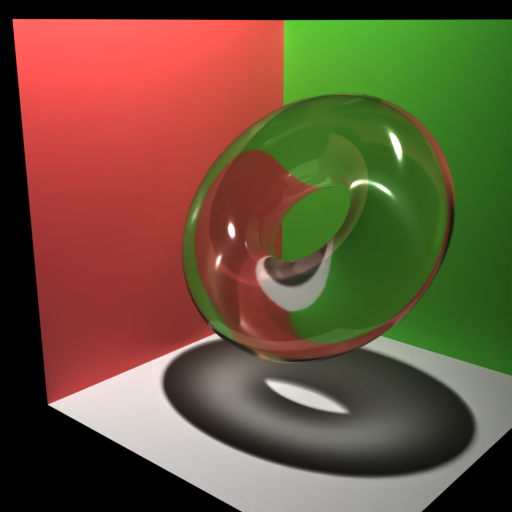}} &
\centmark{\includegraphics[width=0.2\linewidth]{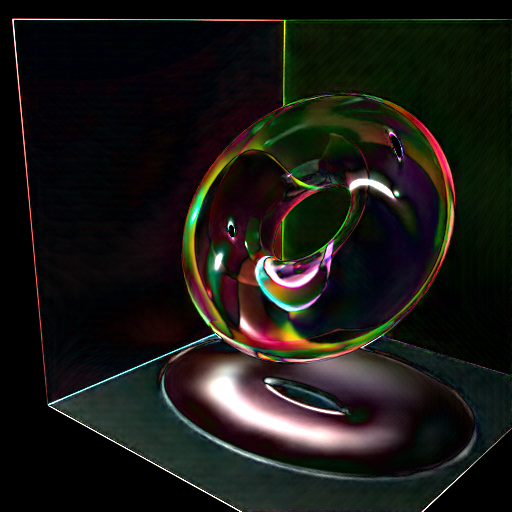}} &
\centmark{\includegraphics[width=0.2\linewidth]{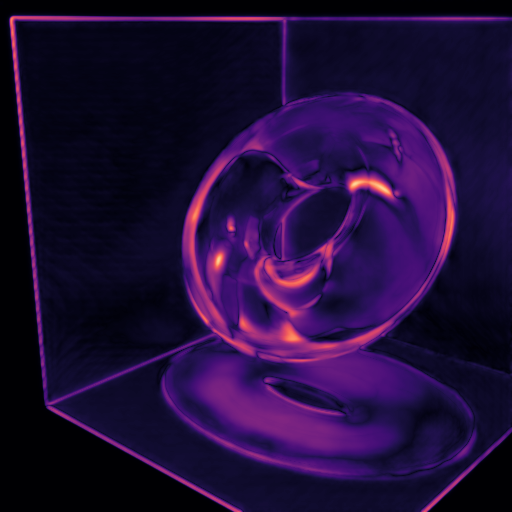}} &
\centmark{\parbox{0.2\linewidth}{\centering\small \#Tokens: 4481\newline\small PSNR: 26.50\newline SSIM: 0.9481\newline LPIPS: 0.0485\newline FLIP: 0.1189}} \\
\centmark{\includegraphics[width=0.2\linewidth]{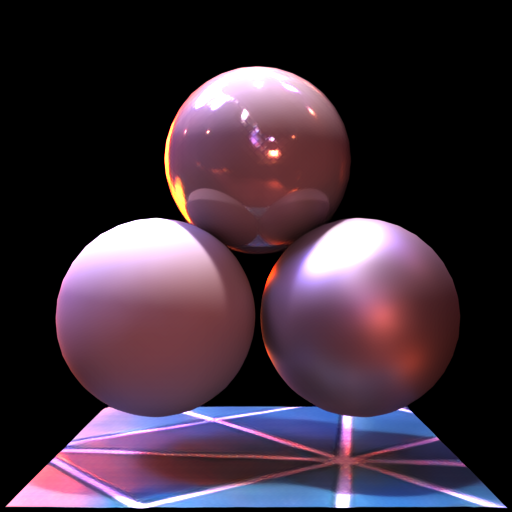}} &
\centmark{\includegraphics[width=0.2\linewidth]{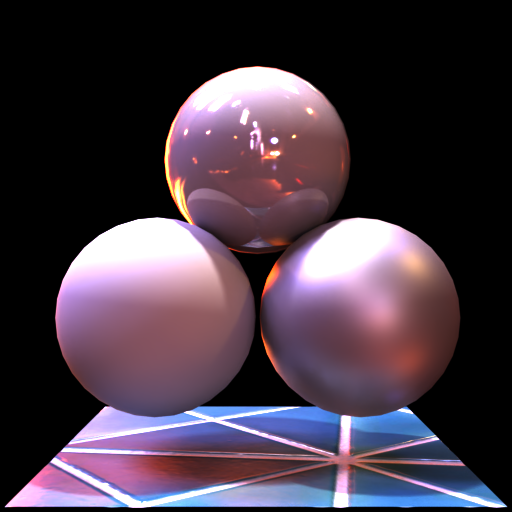}} &
\centmark{\includegraphics[width=0.2\linewidth]{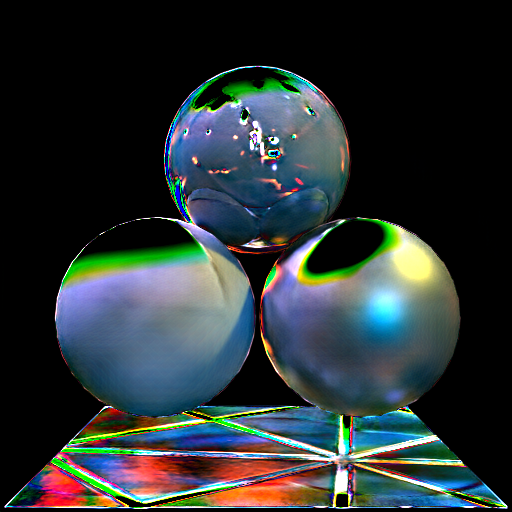}} &
\centmark{\includegraphics[width=0.2\linewidth]{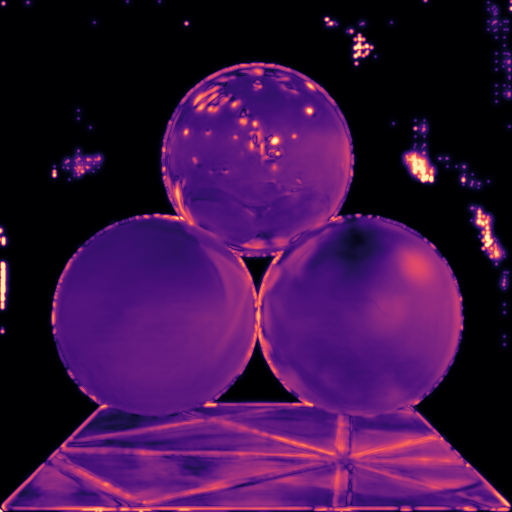}} &
\centmark{\parbox{0.2\linewidth}{\centering\small \#Tokens: 3009\newline\small PSNR: 22.24\newline SSIM: 0.9265\newline LPIPS: 0.0398\newline FLIP: 0.1773}} \\
\centmark{\includegraphics[width=0.2\linewidth]{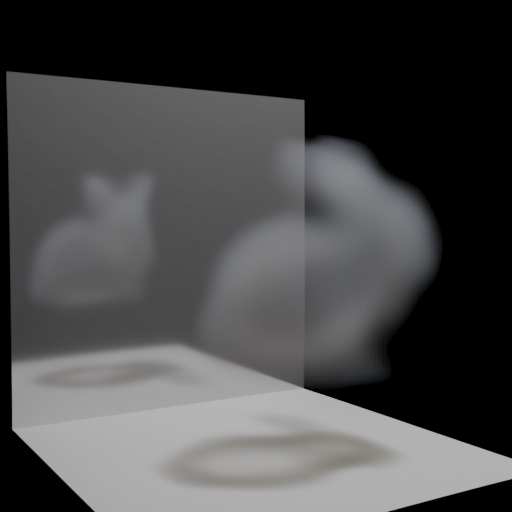}} &
\centmark{\includegraphics[width=0.2\linewidth]{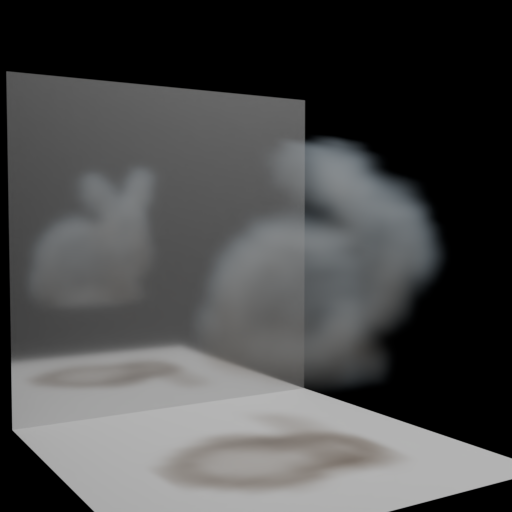}} &
\centmark{\includegraphics[width=0.2\linewidth]{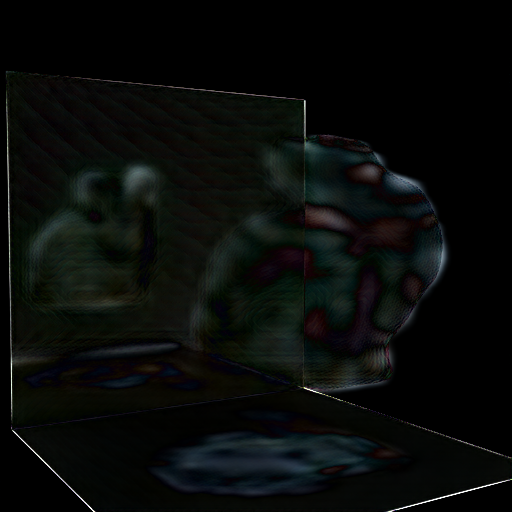}} &
\centmark{\includegraphics[width=0.2\linewidth]{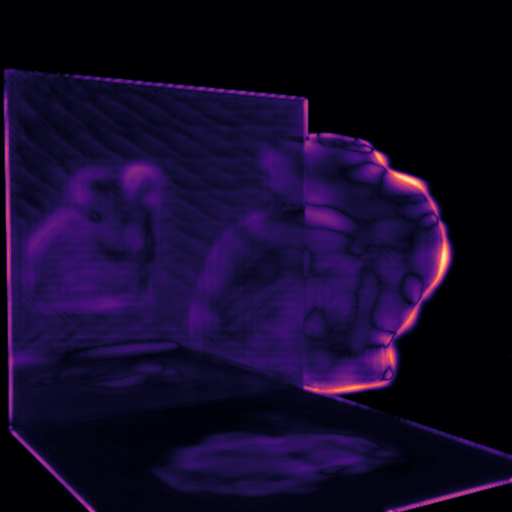}} &
\centmark{\parbox{0.2\linewidth}{\centering\small \#Tokens: 490\newline\small PSNR: 36.51\newline SSIM: 0.9861\newline LPIPS: 0.0373\newline FLIP: 0.0825}} \\
\centmark{\includegraphics[width=0.2\linewidth]{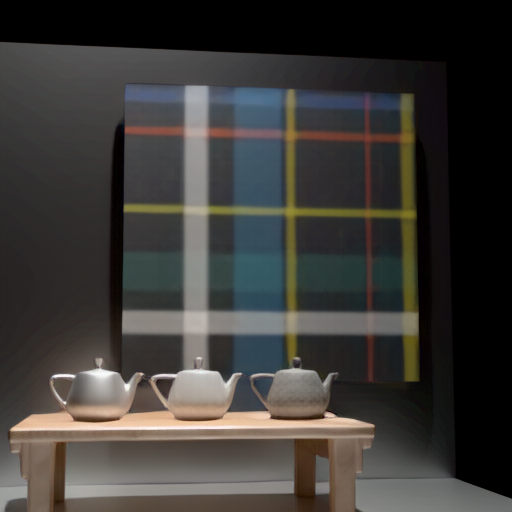}} &
\centmark{\includegraphics[width=0.2\linewidth]{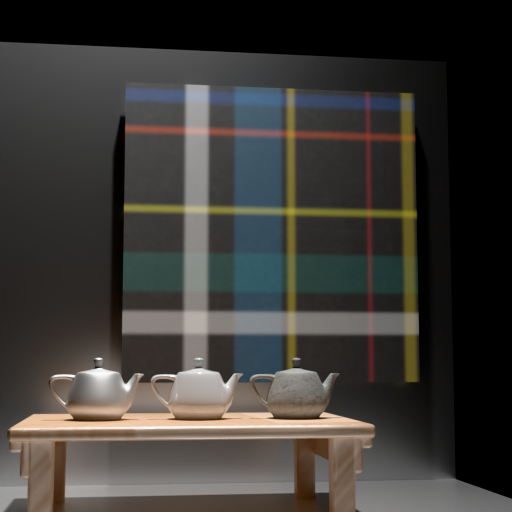}} &
\centmark{\includegraphics[width=0.2\linewidth]{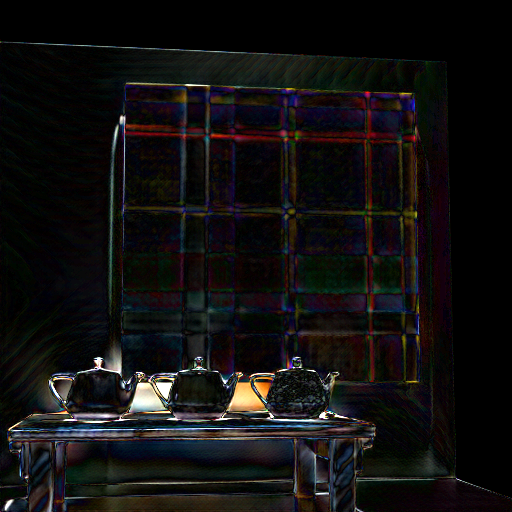}} &
\centmark{\includegraphics[width=0.2\linewidth]{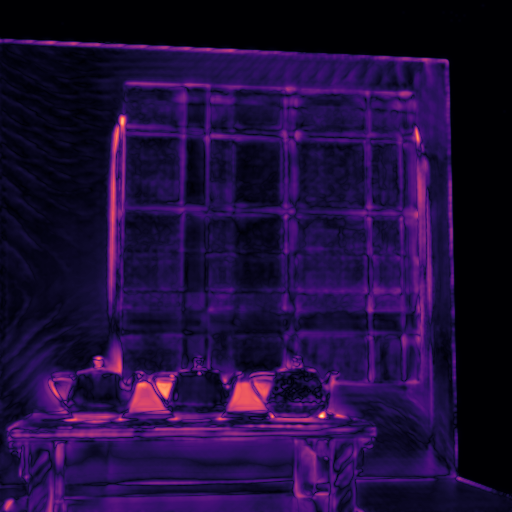}} &
\centmark{\parbox{0.2\linewidth}{\centering\small \#Tokens: 17653\newline\small PSNR: 30.99\newline SSIM: 0.9653\newline LPIPS: 0.0320\newline FLIP: 0.1168}} \\
\end{tabular}
\caption{
Scenes from the main submission rendered with \RF2 and compared to path-traced reference images.}
\label{fig:supp_omniren_results_demos}
\end{figure}

\begin{figure}[t]
\centering
\setlength{\tabcolsep}{0pt}
\renewcommand{\arraystretch}{0}
\begin{tabular}{c c c c c}
\textbf{\RF2} & \textbf{Reference} & \textbf{Diff ($\times 5$)} & \textbf{FLIP} & \textbf{Metrics} \\
\centmark{\includegraphics[width=0.2\linewidth]{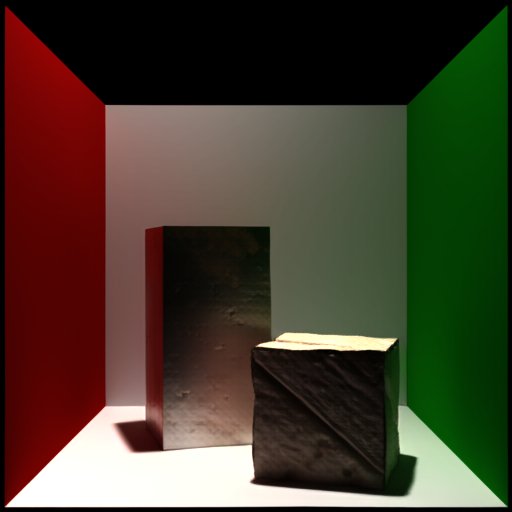}} &
\centmark{\includegraphics[width=0.2\linewidth]{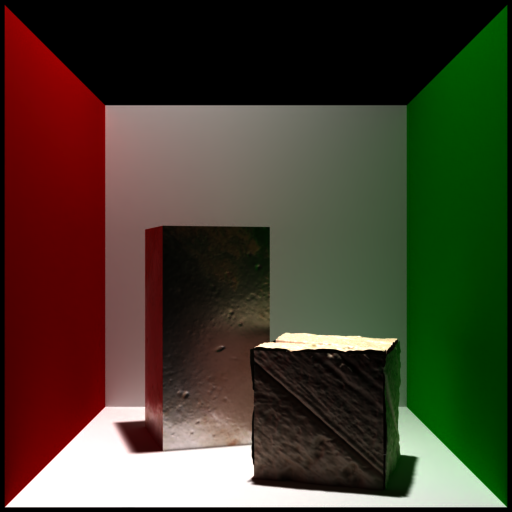}} &
\centmark{\includegraphics[width=0.2\linewidth]{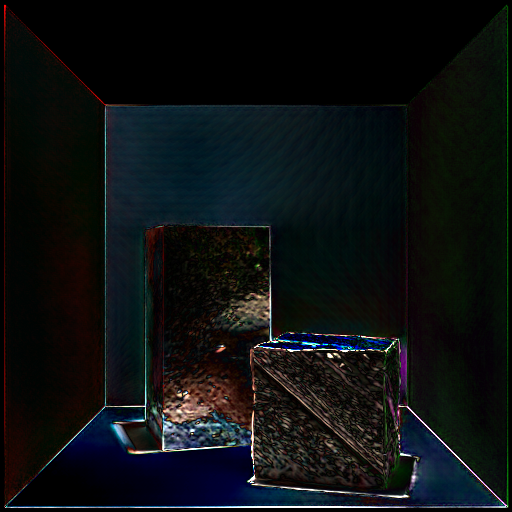}} &
\centmark{\includegraphics[width=0.2\linewidth]{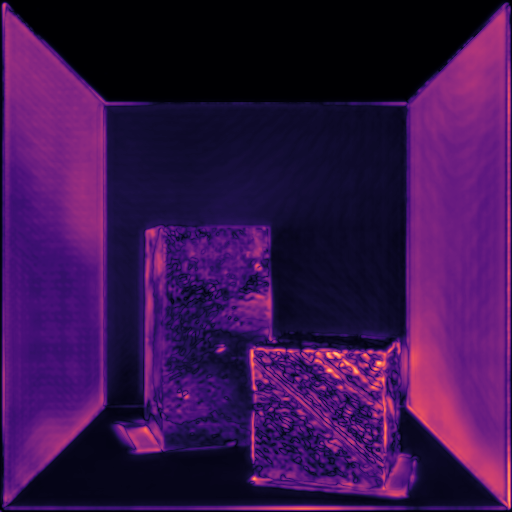}} &
\centmark{\parbox{0.2\linewidth}{\centering\small \#Tokens: 5633\newline\small PSNR: 30.97\newline SSIM: 0.9297\newline LPIPS: 0.0319\newline FLIP: 0.1845}} \\
\centmark{\includegraphics[width=0.2\linewidth]{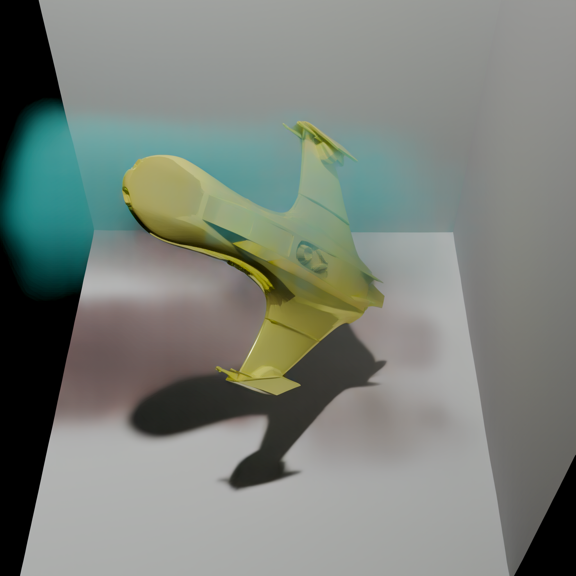}} &
\centmark{\includegraphics[width=0.2\linewidth]{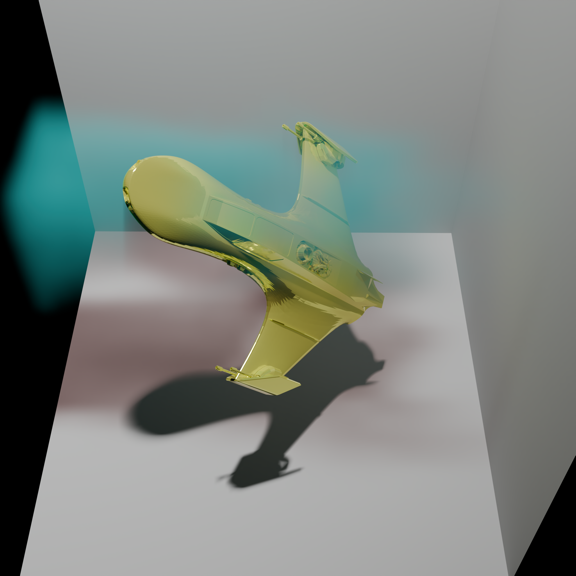}} &
\centmark{\includegraphics[width=0.2\linewidth]{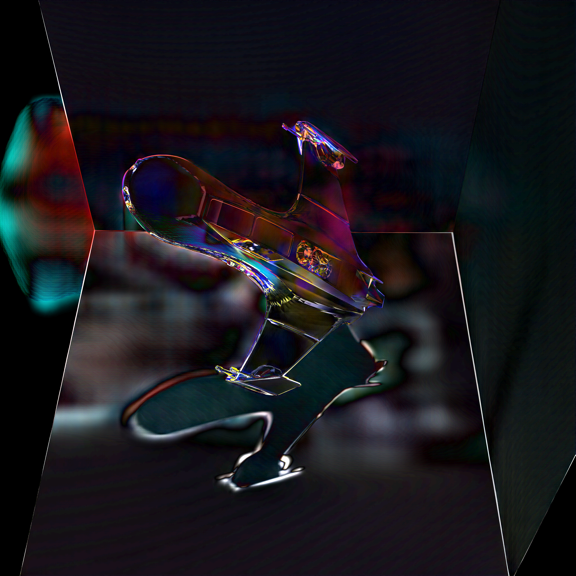}} &
\centmark{\includegraphics[width=0.2\linewidth]{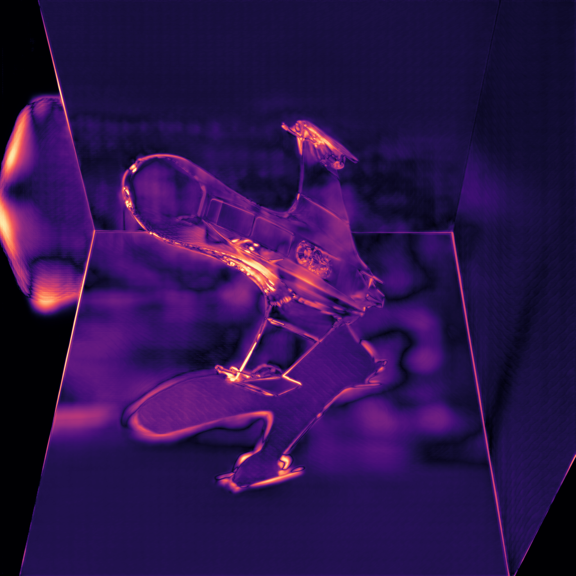}} &
\centmark{\parbox{0.2\linewidth}{\centering\small \#Tokens: 36353\newline\small PSNR: 27.05\newline SSIM: 0.9633\newline LPIPS: 0.0648\newline FLIP: 0.1605}} \\
\centmark{\includegraphics[width=0.2\linewidth]{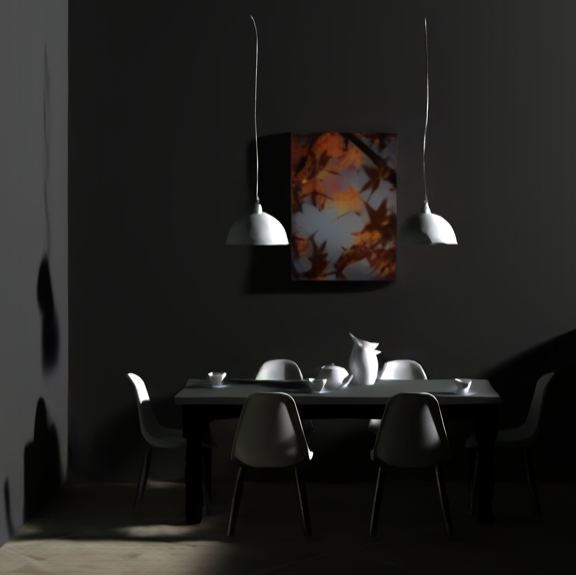}} &
\centmark{\includegraphics[width=0.2\linewidth]{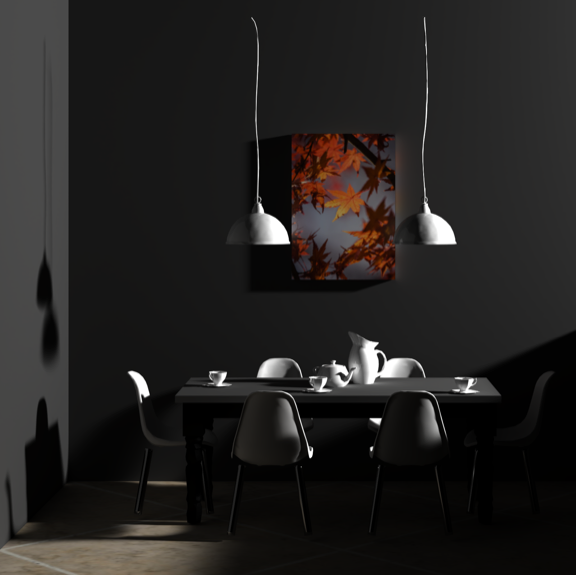}} &
\centmark{\includegraphics[width=0.2\linewidth]{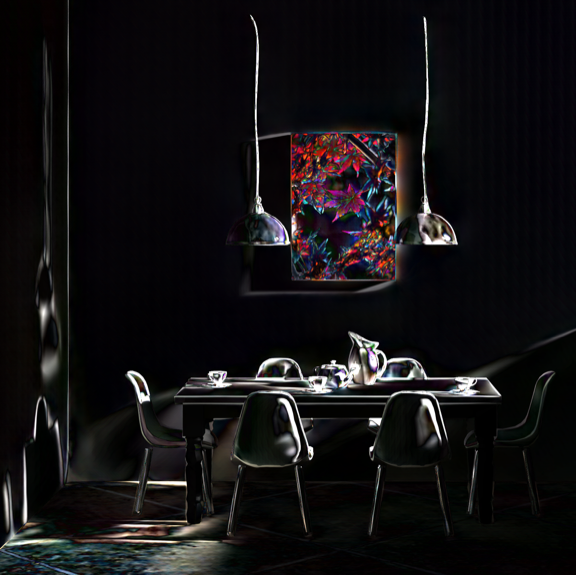}} &
\centmark{\includegraphics[width=0.2\linewidth]{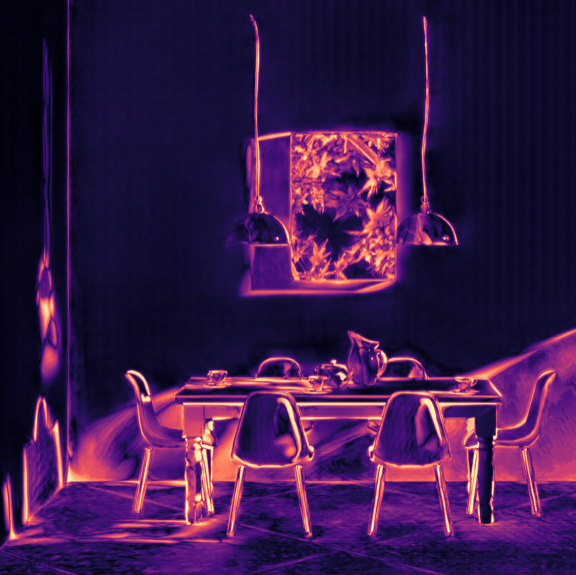}} &
\centmark{\parbox{0.2\linewidth}{\centering\small \#Tokens: 29276\newline\small PSNR: 24.34\newline SSIM: 0.9171\newline LPIPS: 0.0712\newline FLIP: 0.1992}} \\
\centmark{\includegraphics[width=0.2\linewidth]{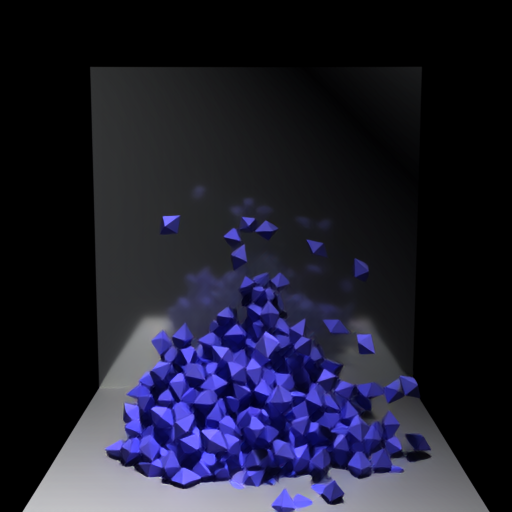}} &
\centmark{\includegraphics[width=0.2\linewidth]{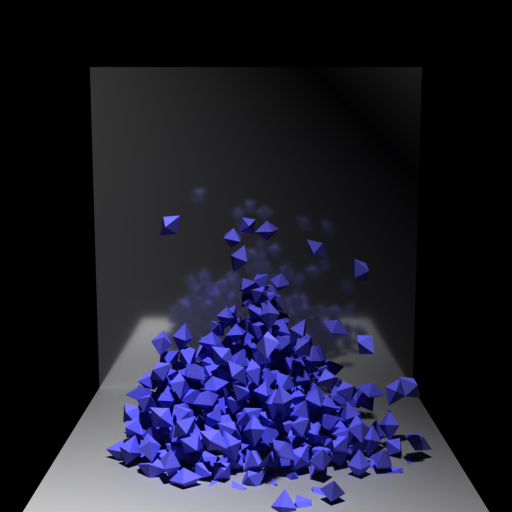}} &
\centmark{\includegraphics[width=0.2\linewidth]{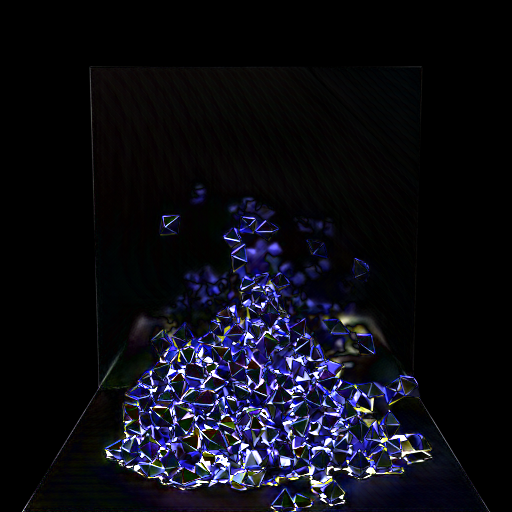}} &
\centmark{\includegraphics[width=0.2\linewidth]{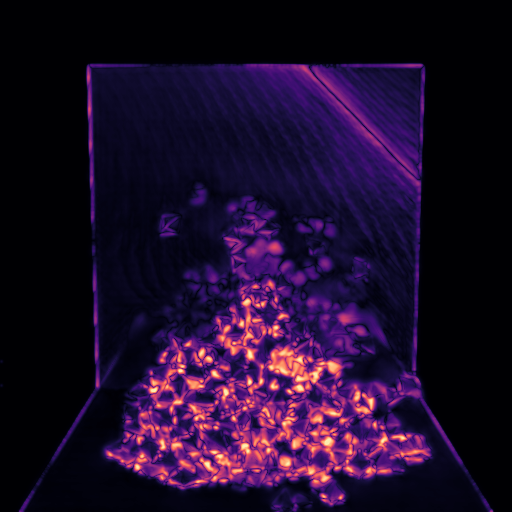}} &
\centmark{\parbox{0.2\linewidth}{\centering\small \#Tokens: 3993\newline\small PSNR: 26.41\newline SSIM: 0.9243\newline LPIPS: 0.0474\newline FLIP: 0.1031}} \\
\end{tabular}
\caption{
Scenes from the main submission rendered with \RF2 and compared to path-traced reference images.}
\label{fig:supp_omniren_results_demos2}
\end{figure}

\section{Qualitative and Quantitative Comparison}

\autoref{fig:supp_omniren_results_demos}
and~\ref{fig:supp_omniren_results_demos2} show qualitative comparisons
for all the scenes from the main submission with respect to reference
Blender Cycles path-traced renderings.  For each scene we also show a
difference image (scaled $5\times$ to better show the differences) and
a FLIP error image~\cite{Andersson:2020:FDE}.  In addition, we list
the total number of tokens per scene, and the PSNR, SSIM, LPIPS and
FLIP errors.
\begin{compactenum}
\item \textbf{Transparent Torus}:
  (\autoref{fig:supp_omniren_results_demos}, 1st row) the differences
  are mainly visible on high curvature areas of the refractive torus,
  as well as in the intensity of the caustic on the ground plane.
\item \textbf{Environment Lit Spheres}: (2nd row) the differences are
  mainly due to differences at high-frequency edges in the image at
  texture / environment map pixel edges. Furthermore, we can also
  observe a slight overall brightness difference.
\item \textbf{Smoky Bunny}: (3rd row) Due to differences in how
  anti-aliasing is handled (Blender Cycles uses adaptive filtering),
  larger differences are visible at high frequency edges in the
  rendered images.
\item \textbf{Three Teapot Scene}: (4th row) Similar to the previous
  scene, differences are mainly concentrated at high frequency edges
  in the image.  Another area of difference is the inter-reflection
  below the painting; \RF2 assumes the back of the painting is also
  textured and reflected onto the wall.
\item \textbf{Displacement Mapped Cornell Cube}:
  (\autoref{fig:supp_omniren_results_demos2}, 1st row). Again most
  differences are in high-frequency areas, due to (1) differences in
  filtering, and (2) \RF2 sometimes misses small details.
\item \textbf{Spaceship in Smoke}: (2nd row) The main differences are
  due to minor inaccuracies in reflected directions (\ie, shifted or
  missing highlights).
\item \textbf{Dinner Scene}: (3rd row) Again, the main differences are
  at high frequency edges in the rendering, as well as slightly more
  blurred shadows.
\item \textbf{Cube Pile Scene}: (4th row) Similar as in prior scenes;
  the main differences are at high frequency edges.
\end{compactenum}

\section{Comparisons with RenderFormer}
\RF2 shares some architectural similarities with
RenderFormer~\cite{Zeng:2025:RFT}.  To better assess the differences
and improvements, we perform an in-depth comparison.

\subsection{Architecture}
\begin{figure}[t]
    \centering
    \includegraphics[
        width=\linewidth,
    ]{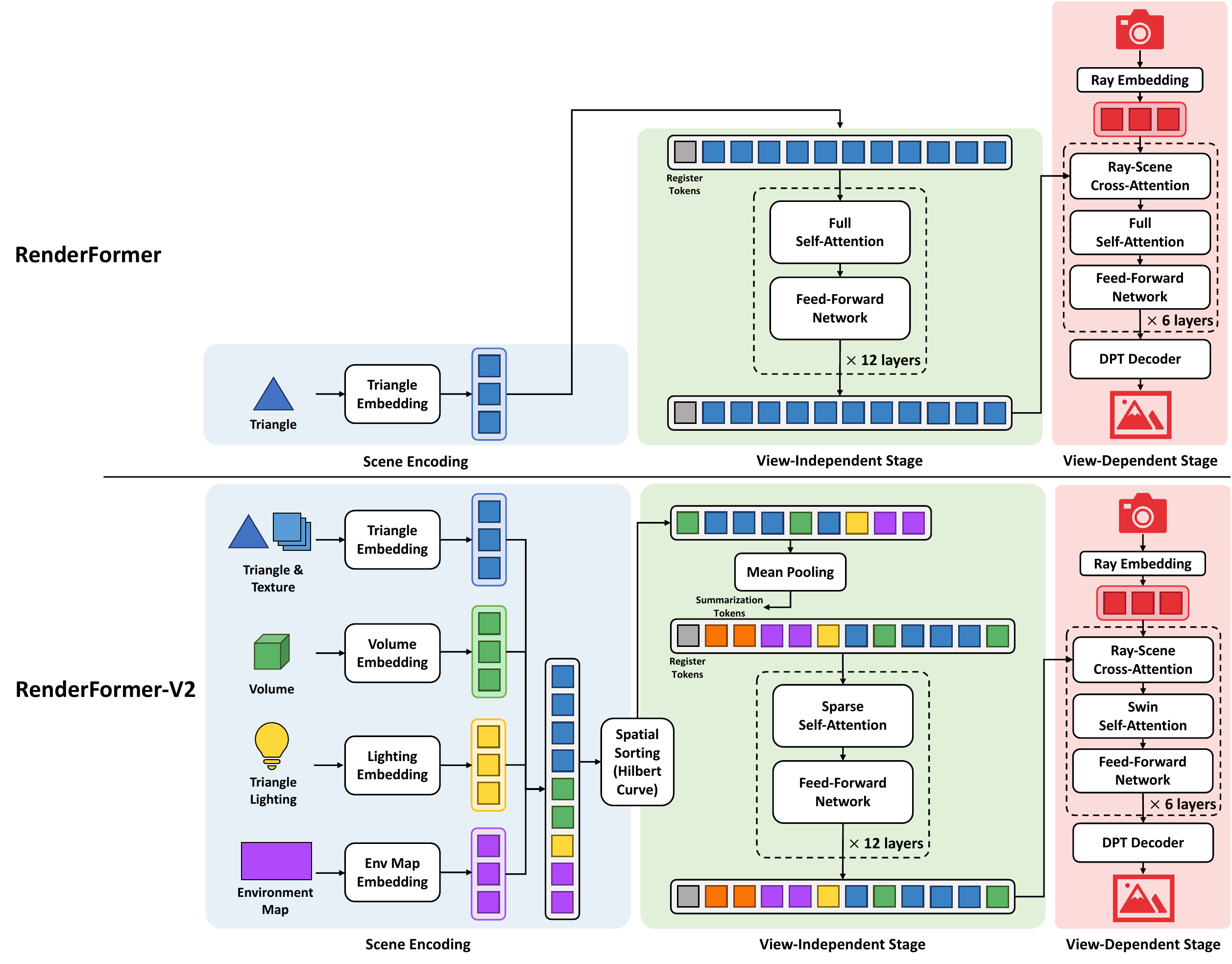}

    \caption{\RF2 and RenderFormer Model Architecture.} 
    \label{fig:supp_model_arch}
\end{figure}

\autoref{fig:supp_model_arch} contrasts the RenderFormer architecture
with \RF2's architecture.  The key differences are:
\begin{compactenum}
\item \textbf{Input Tokens}. RenderFormer encodes all primitives
  (including light sources) in a homogeneous triangle token. \RF2
  supports heterogeneous tokens, each with its own encoding procedure,
  supporting a wide range of primitives ranging from triangles, volume
  elements, and environment maps.  In addition, \RF2 adds support
  for spatially varying materials (with displacement mapping) and uses
  a BRDF-model-agnostic material specification.
\item \textbf{View-independent Attention}. Whereas RenderFormer
  utilizes a full (dense) self-attention, \RF2 uses a sliding
  windowed attention combined with attention sinks.  The sinks have
  rendering-aware semantics and include: global registers, light
  sources, and summarization tokens to model long-range light
  transport.
\item \textbf{View-dependent Attention}. RenderFormer uses full
  self-attention between all ray-bundle tokens.  This is costly and it
  is unlikely that pixels far away in the image will be meaningfully
  related to nearby pixels.  Therefore, \RF2 uses SWIN attention
  instead.
\end{compactenum}
By extending the supported token types and by adopting a novel sparse
attention architecture, \RF2 enables more accurate rendering of
more complex scenes with richer visual effects.


\subsection{Qualitative Comparison}
\begin{figure}[t]
\centering
\setlength{\tabcolsep}{0pt}
\renewcommand{\arraystretch}{0}
\begin{tabular}{c c c c c}
\textbf{Ours} & \textbf{Diff ($\times 5$)} & \textbf{RenderFormer} & \textbf{Diff ($\times 5$)} & \textbf{Reference} \\
\centmark{\includegraphics[width=0.2\linewidth]{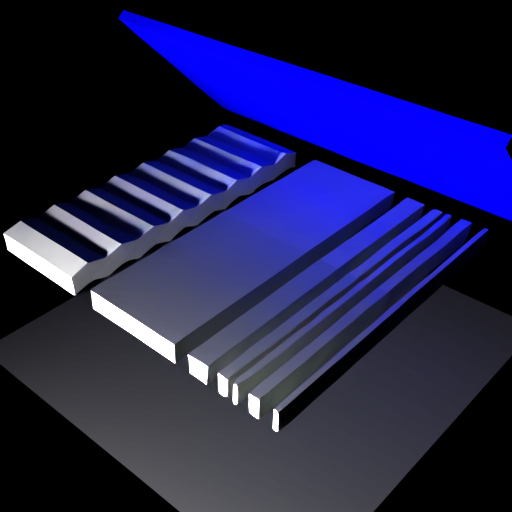}} &
\centmark{\includegraphics[width=0.2\linewidth]{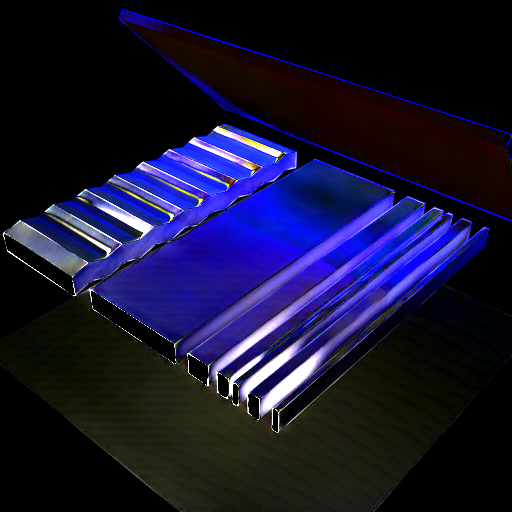}} &
\centmark{\includegraphics[width=0.2\linewidth]{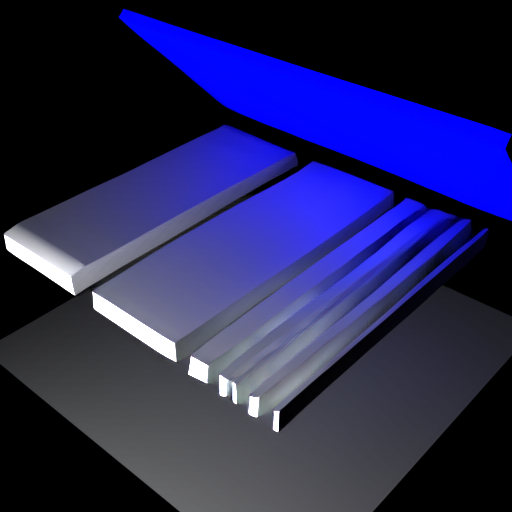}} &
\centmark{\includegraphics[width=0.2\linewidth]{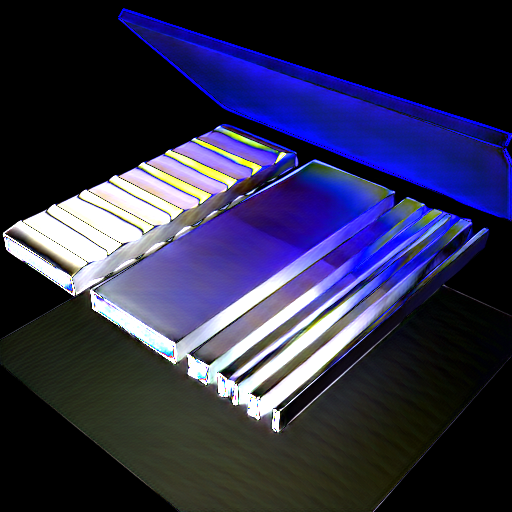}} &
\centmark{\includegraphics[width=0.2\linewidth]{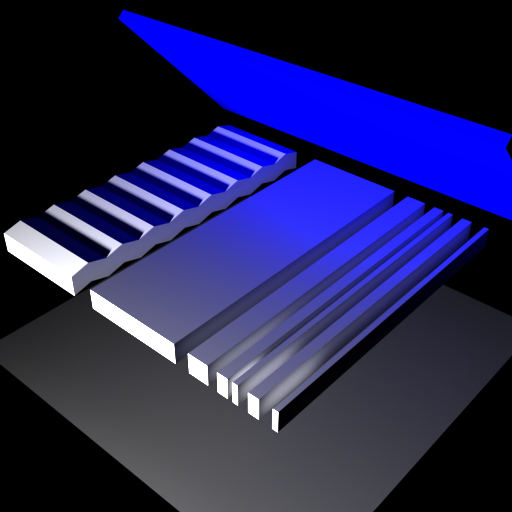}} \\
\centmark{\includegraphics[width=0.2\linewidth]{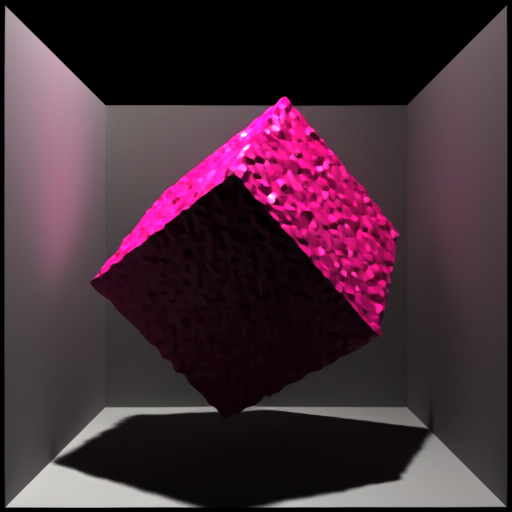}} &
\centmark{\includegraphics[width=0.2\linewidth]{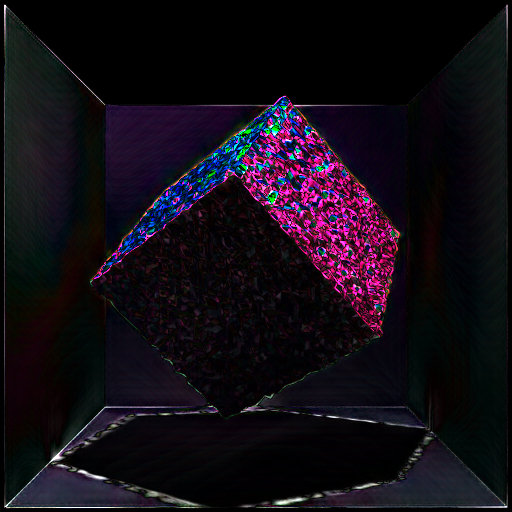}} &
\centmark{\includegraphics[width=0.2\linewidth]{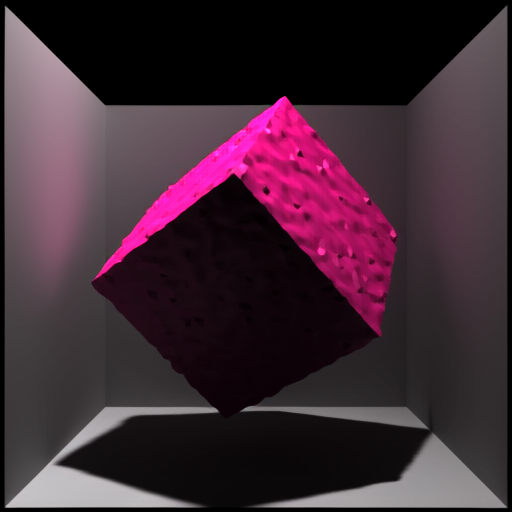}} &
\centmark{\includegraphics[width=0.2\linewidth]{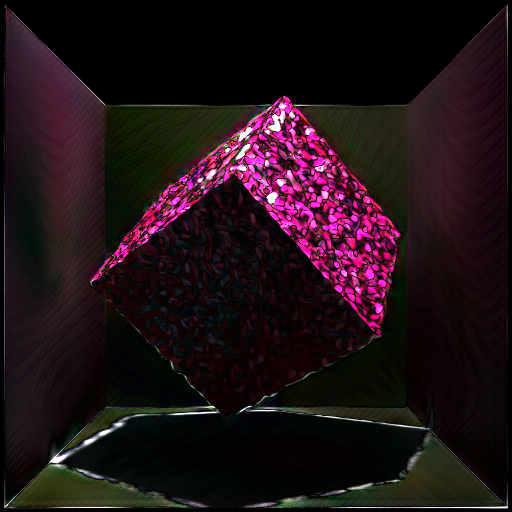}} &
\centmark{\includegraphics[width=0.2\linewidth]{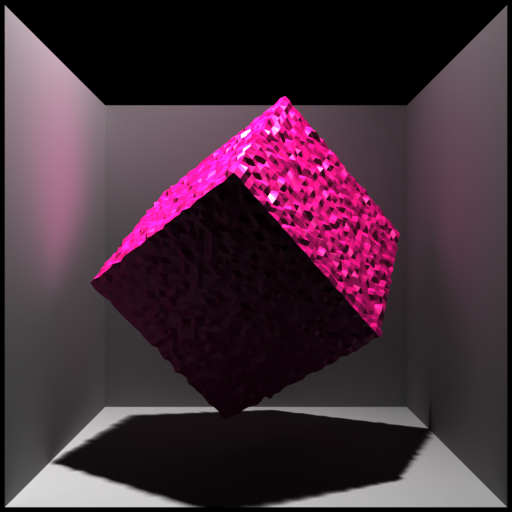}} \\
\centmark{\includegraphics[width=0.2\linewidth]{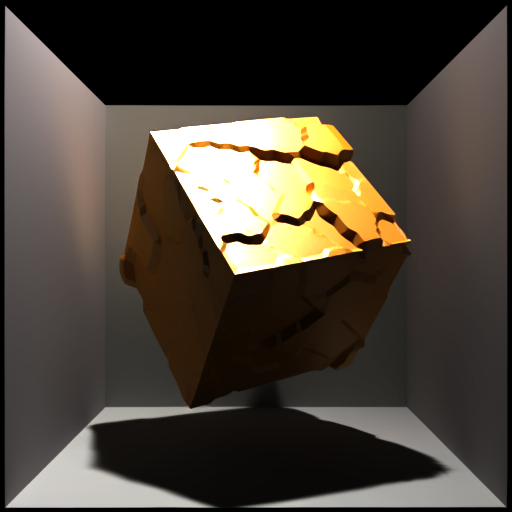}} &
\centmark{\includegraphics[width=0.2\linewidth]{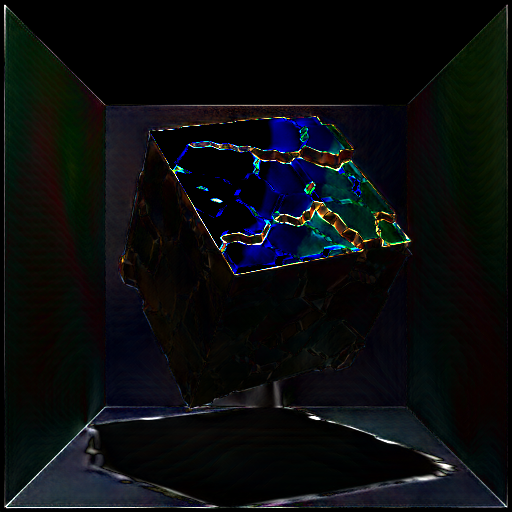}} &
\centmark{\includegraphics[width=0.2\linewidth]{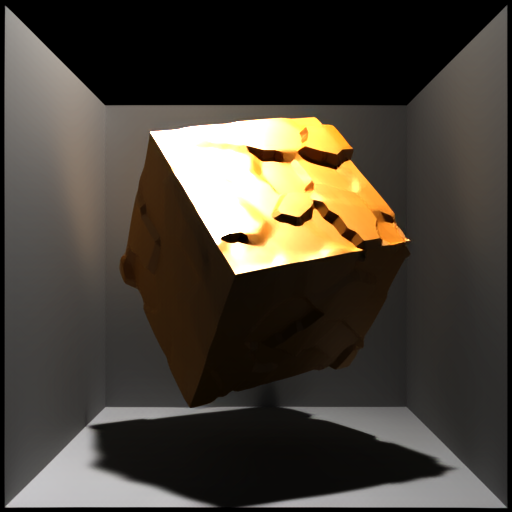}} &
\centmark{\includegraphics[width=0.2\linewidth]{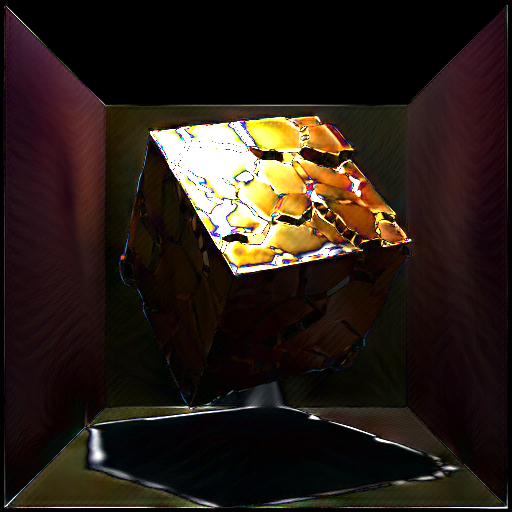}} &
\centmark{\includegraphics[width=0.2\linewidth]{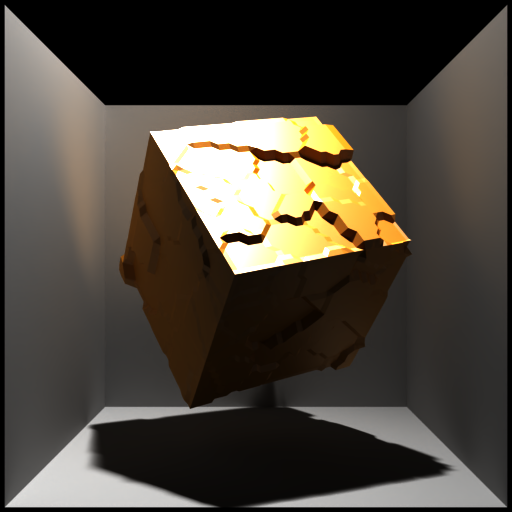}} \\
\end{tabular}
\vspace{-0.2cm}
\caption{
Visual comparison of \RF2 and RenderFormer against a path-traced reference.
Difference maps (scaled by $\times 5$) are shown to highlight rendering errors.
}
\label{fig:supp_rf1_comp}
\end{figure}

\autoref{fig:supp_rf1_comp} qualitatively compares three scenes
rendered with \RF2 vs. RenderFormer vs. a path-tracer reference
rendering (difference images are scaled $5\times$ to better highlight
discrepancies). In all cases, we observe that \RF2 more accurately
renders fine geometrical details such as the corrugated structures and
the narrow gaps between cuboids in the first row, as well as the
high-frequency geometric patterns in the second and third rows. The
difference images further show that \RF2 yields smaller rendering
errors overall.

\begin{table}[t]
  \caption{Comparison of high-resolution render accuracy between RenderFormer and \RF2 (with and without high-resolution fine-tuning). The metrics are computed on renderings at $2048 \times 2048$ resolution.}
  \centering
  \setlength{\tabcolsep}{6pt}
  \renewcommand{\arraystretch}{1.15}
  \begin{tabular}{lcccc}
    \specialrule{0.08em}{0pt}{0pt}
    \textbf{Model} & \textbf{PSNR} $\uparrow$ & \textbf{SSIM} $\uparrow$ & \textbf{LPIPS} $\downarrow$ & \textbf{FLIP} $\downarrow$ \\
    \specialrule{0.05em}{0pt}{0pt}

    RenderFormer & 26.2423 & 0.9473 & 0.0886 & 0.1332 \\
    \RF2 & 28.0810 & 0.9648 & 0.0503 & 0.1172 \\
    \RF2 (finetuned) & \textbf{30.7304} & \textbf{0.9742} & \textbf{0.0244} & \textbf{0.0952} \\

    \specialrule{0.08em}{0pt}{0pt}
  \end{tabular}
  \label{tab:rf1_rf2_scaling_comparison}
\end{table}

\begin{figure}
\centering
\setlength{\tabcolsep}{0pt}
\renewcommand{\arraystretch}{0}
\begin{tabular}{r @{\hspace{4pt}} c c c c}
& \scriptsize Reference & \scriptsize RenderFormer & \scriptsize \RF2 & \scriptsize RF2 (fine-tuned) \\[0.5ex]
\multirow{2}{*}[0.75em]{\rotatebox{90}{\small 512}} &
\includegraphics[width=0.22\linewidth]{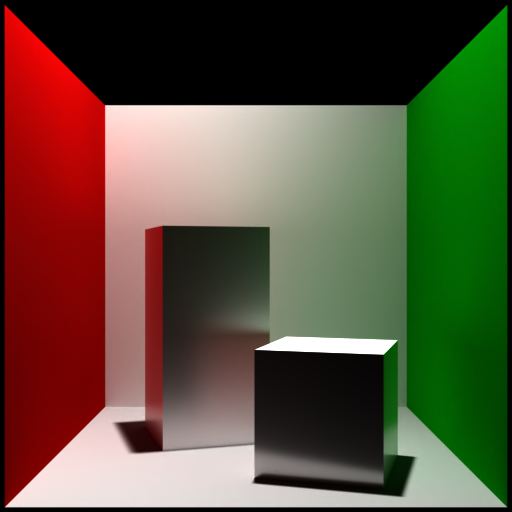}%
& \includegraphics[width=0.22\linewidth]{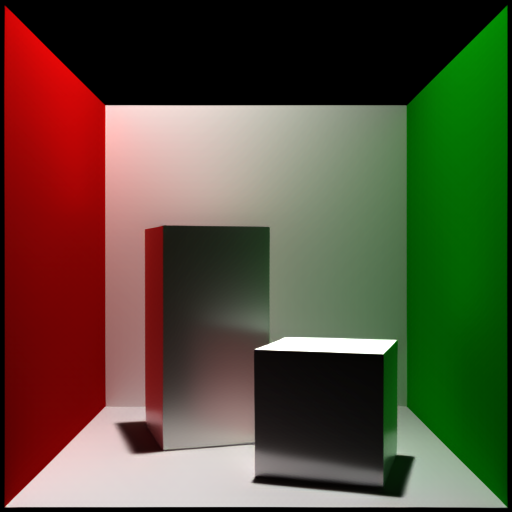}%
& \includegraphics[width=0.22\linewidth]{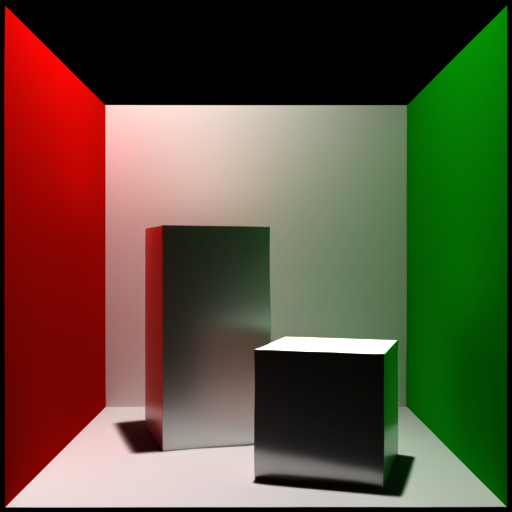}%
& \includegraphics[width=0.22\linewidth]{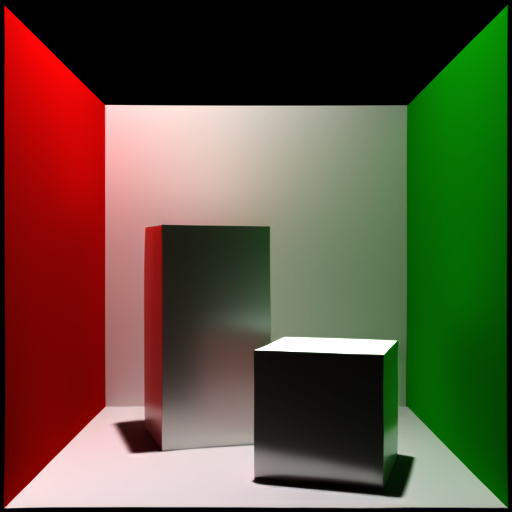} \\
& &
\includegraphics[width=0.22\linewidth]{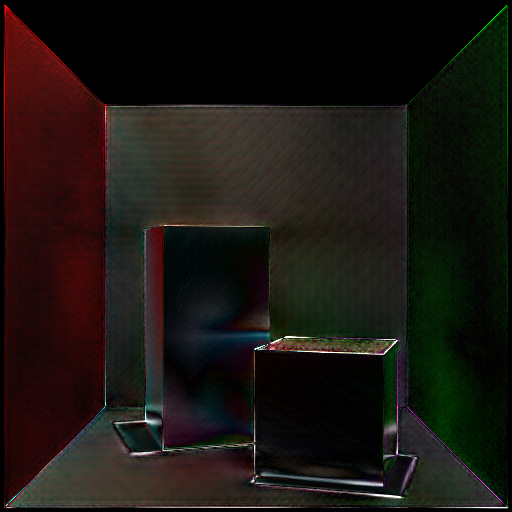}%
& \includegraphics[width=0.22\linewidth]{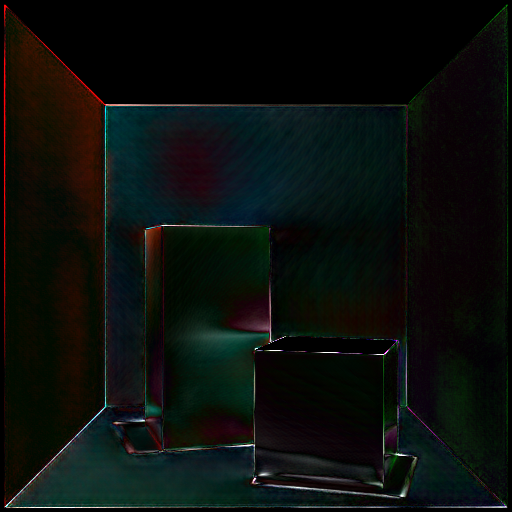}%
& \includegraphics[width=0.22\linewidth]{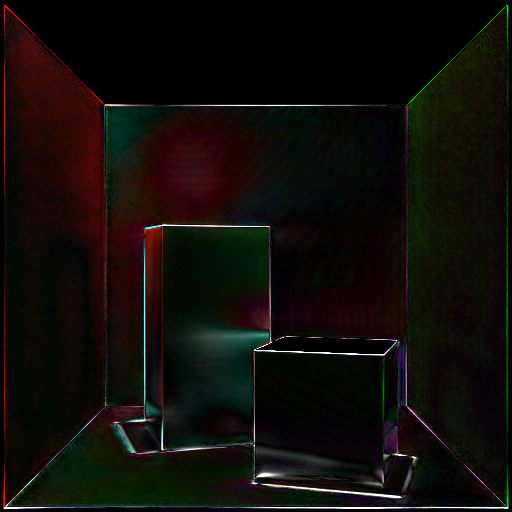} \\
\multirow{2}{*}[0.95em]{\rotatebox{90}{\small 1024}} &
\includegraphics[width=0.22\linewidth]{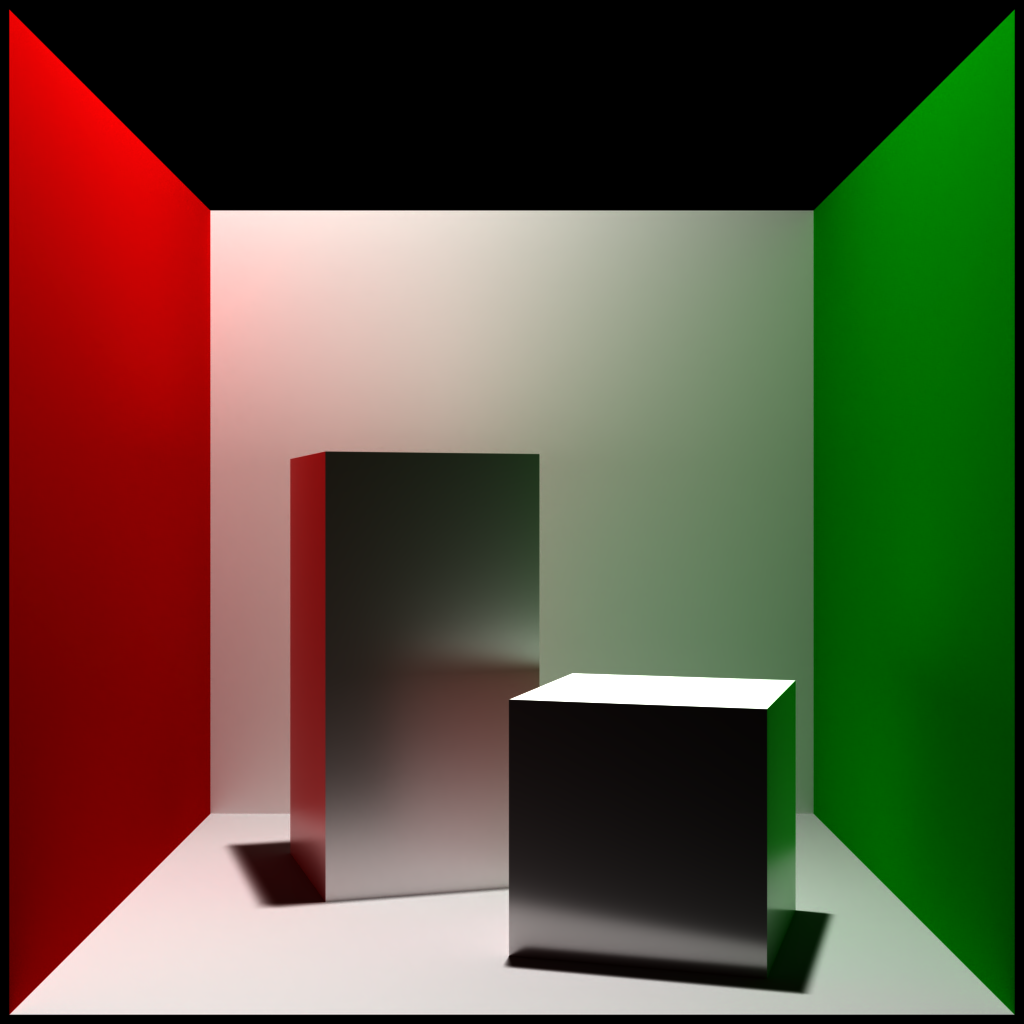}%
& \includegraphics[width=0.22\linewidth]{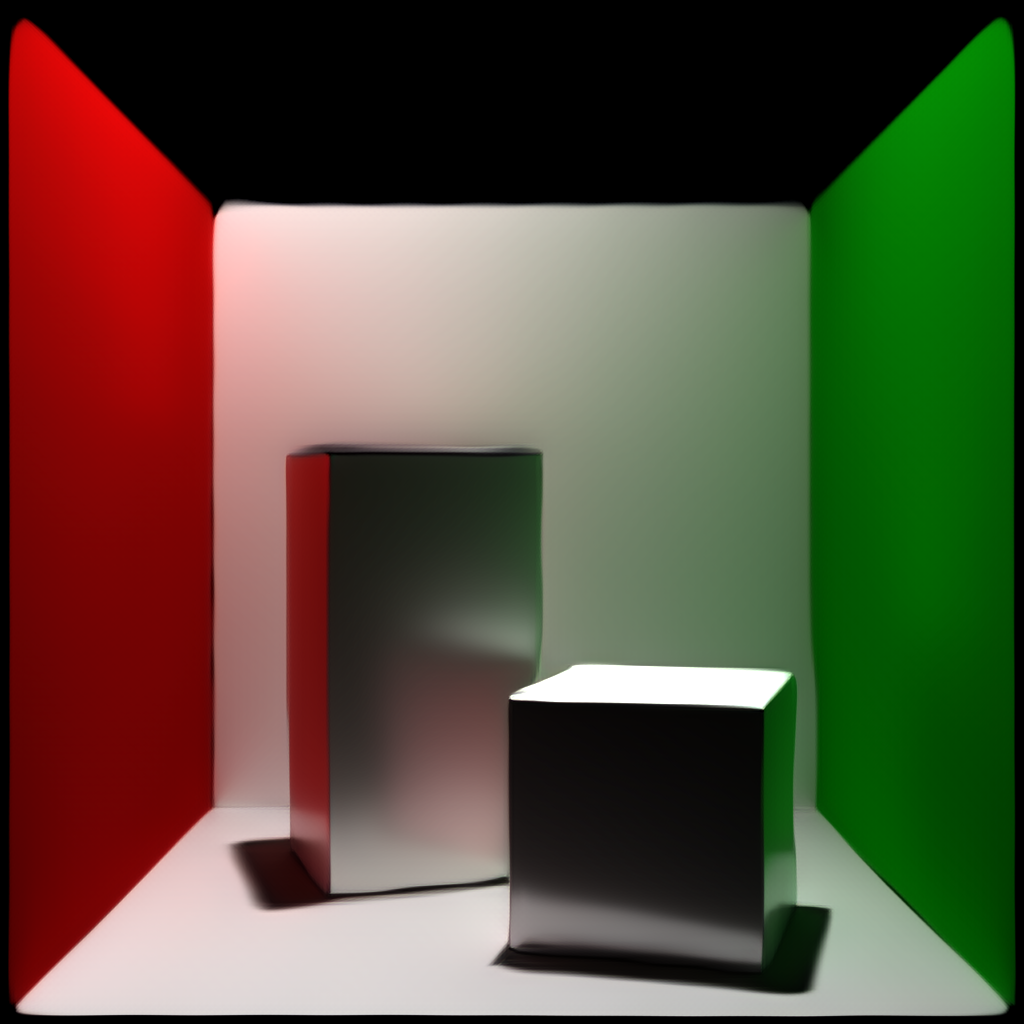}%
& \includegraphics[width=0.22\linewidth]{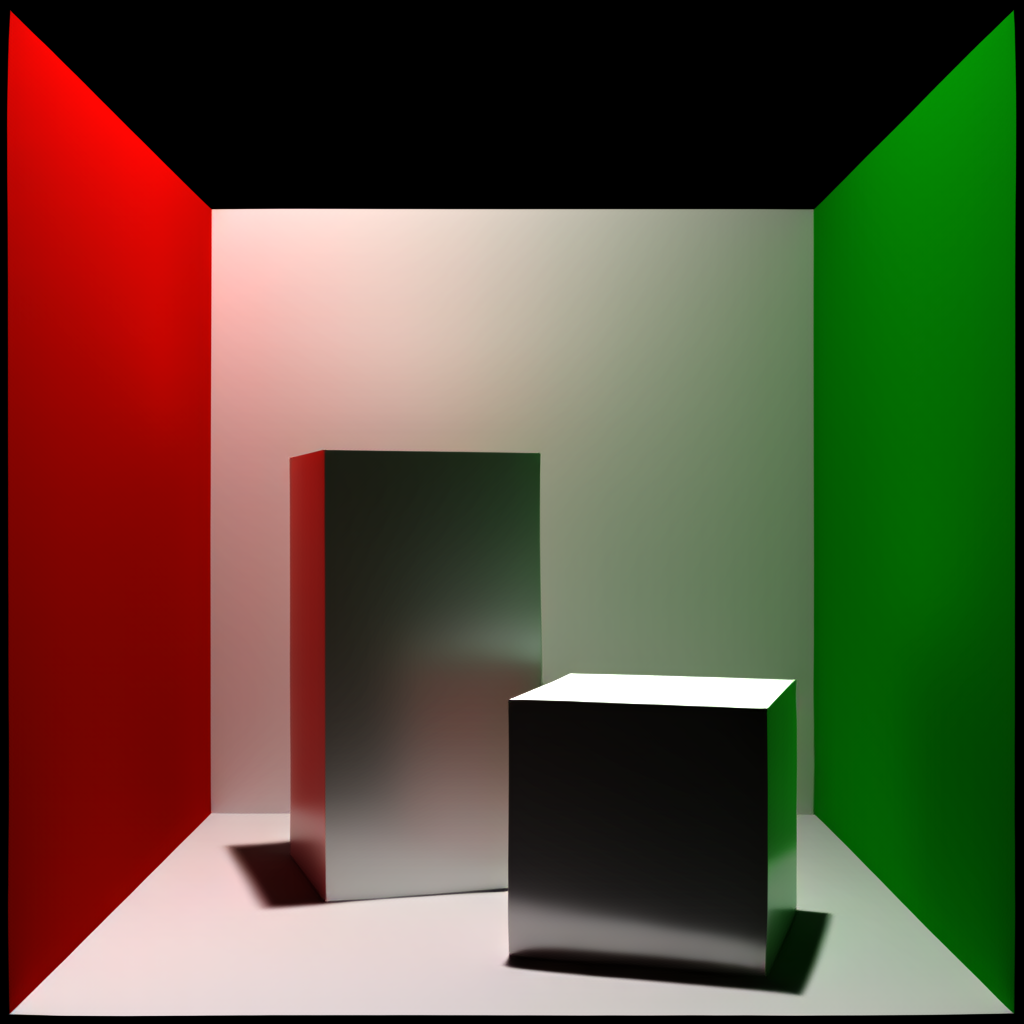}%
& \includegraphics[width=0.22\linewidth]{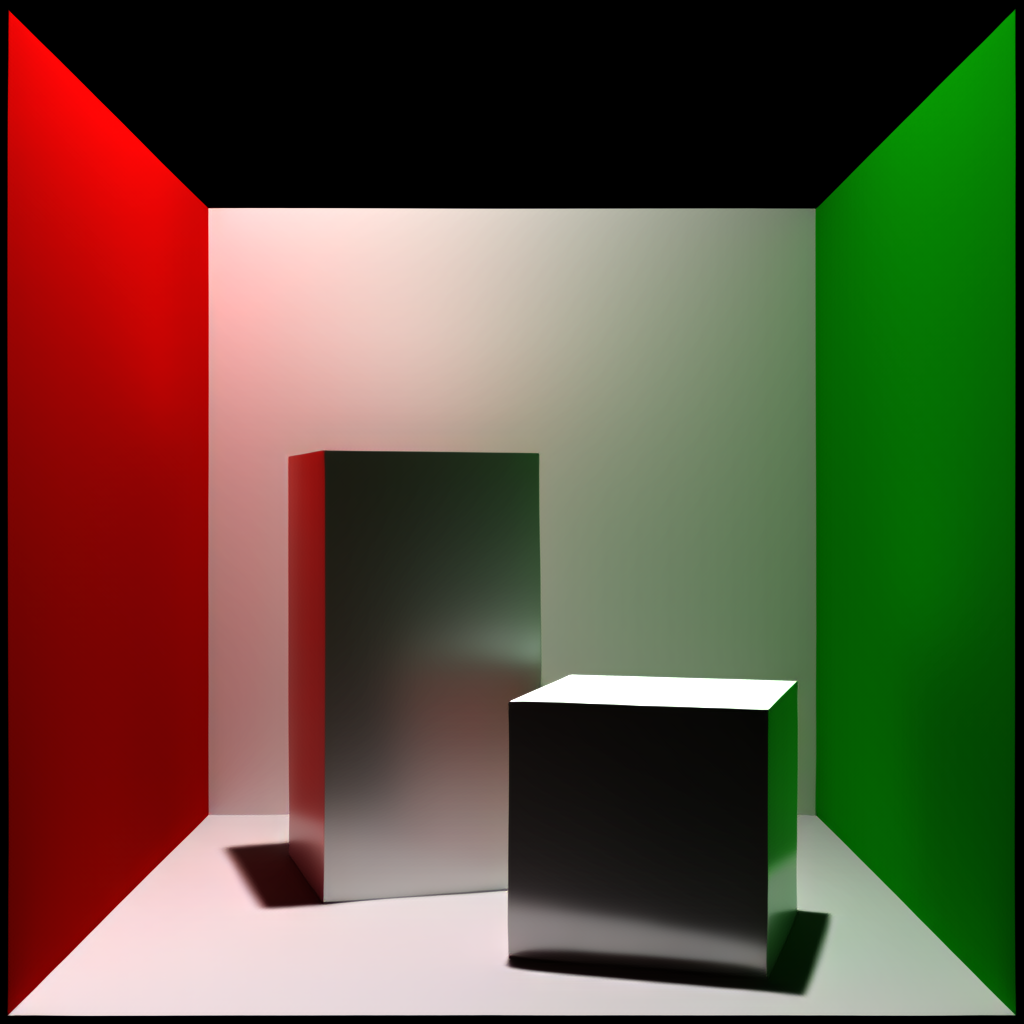} \\
& &
\includegraphics[width=0.22\linewidth]{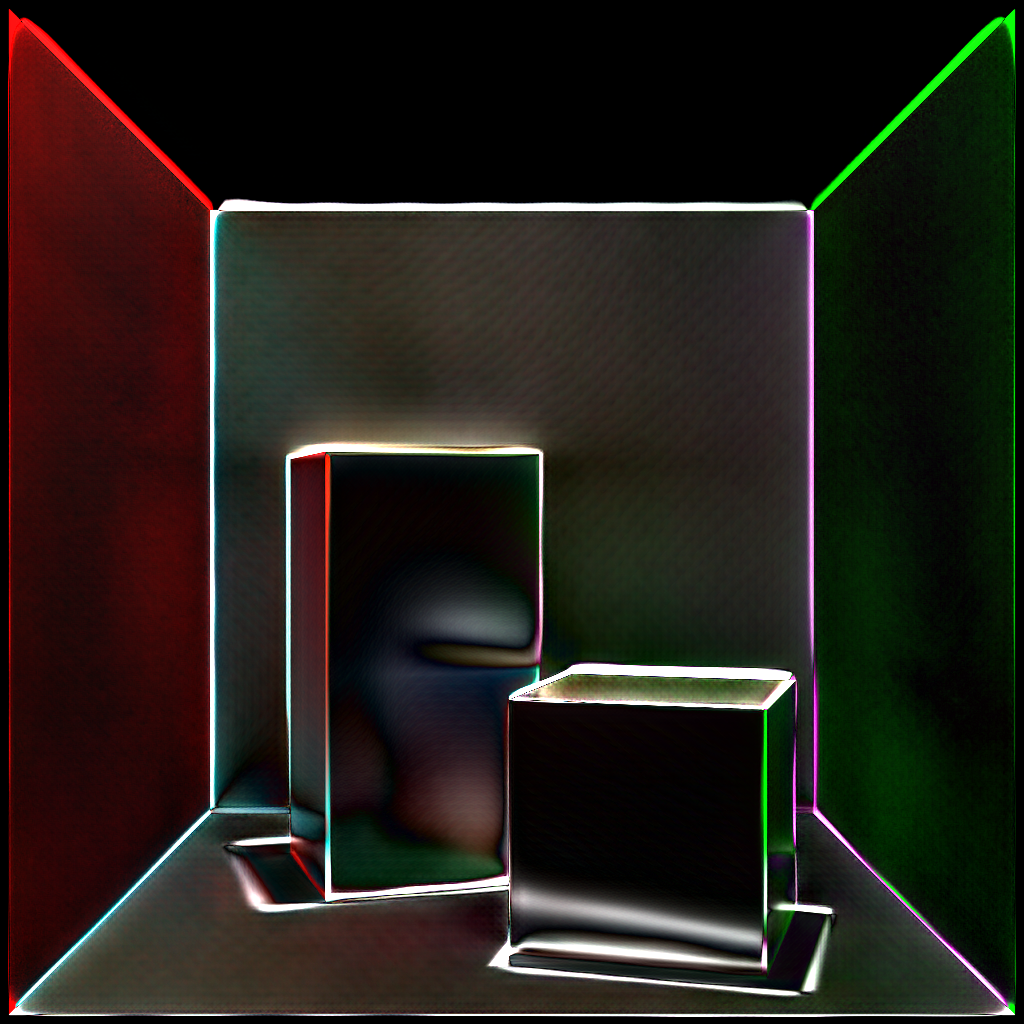}%
& \includegraphics[width=0.22\linewidth]{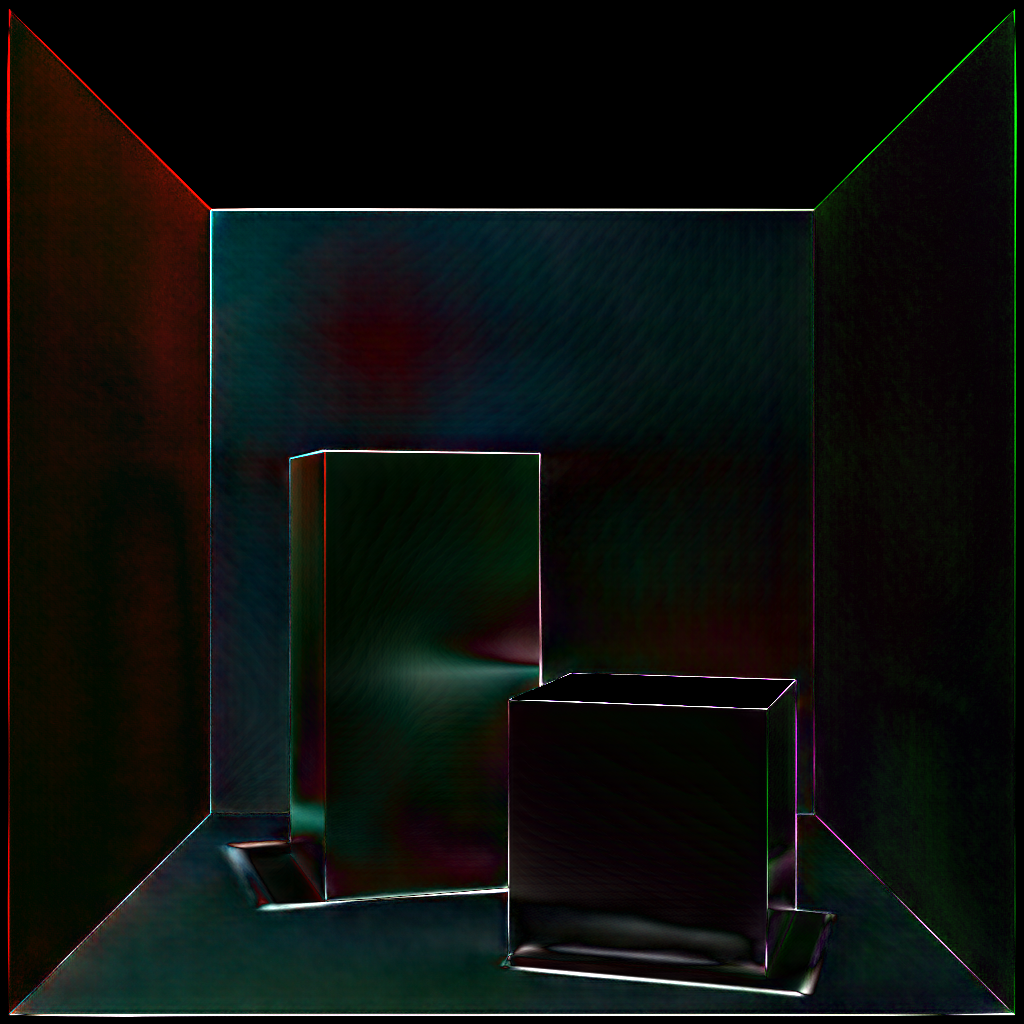}%
& \includegraphics[width=0.22\linewidth]{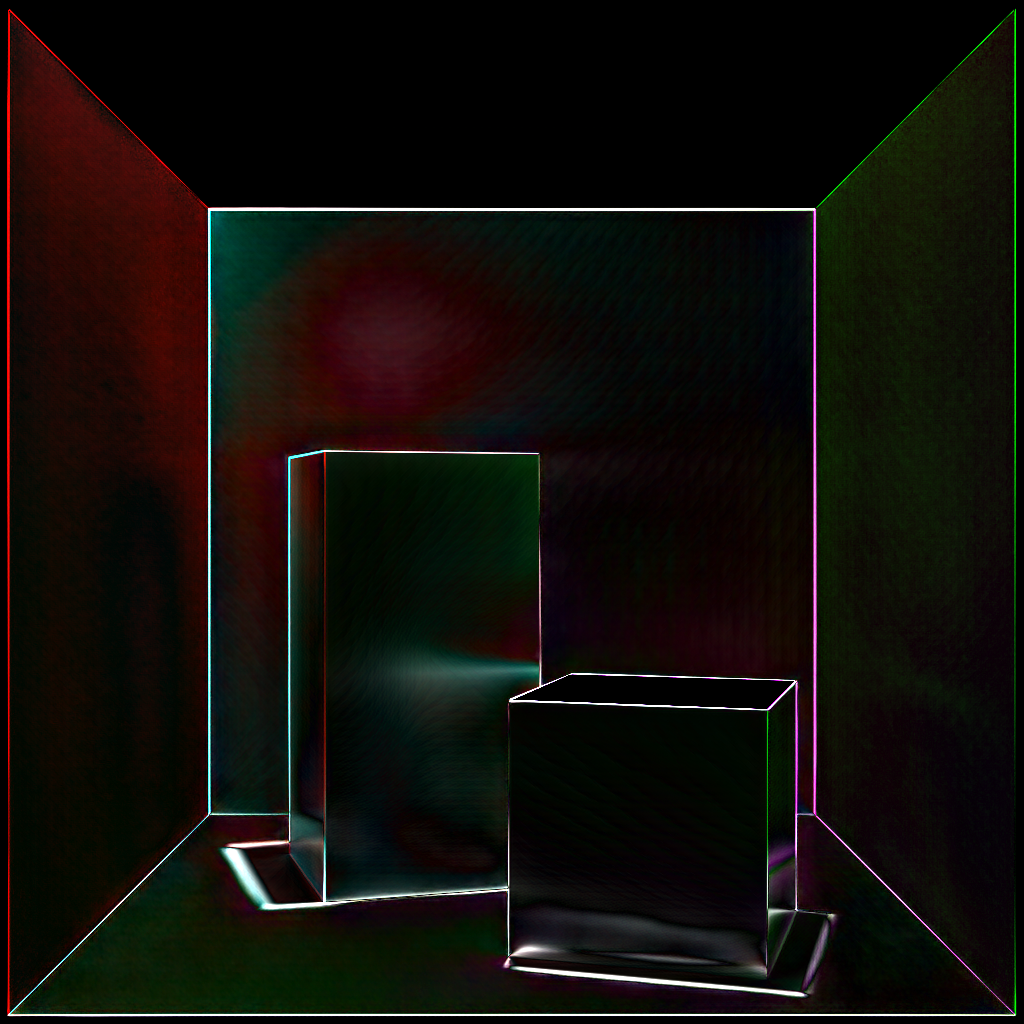} \\
\multirow{2}{*}[0.95em]{\rotatebox{90}{\small 2048}} &
\includegraphics[width=0.22\linewidth]{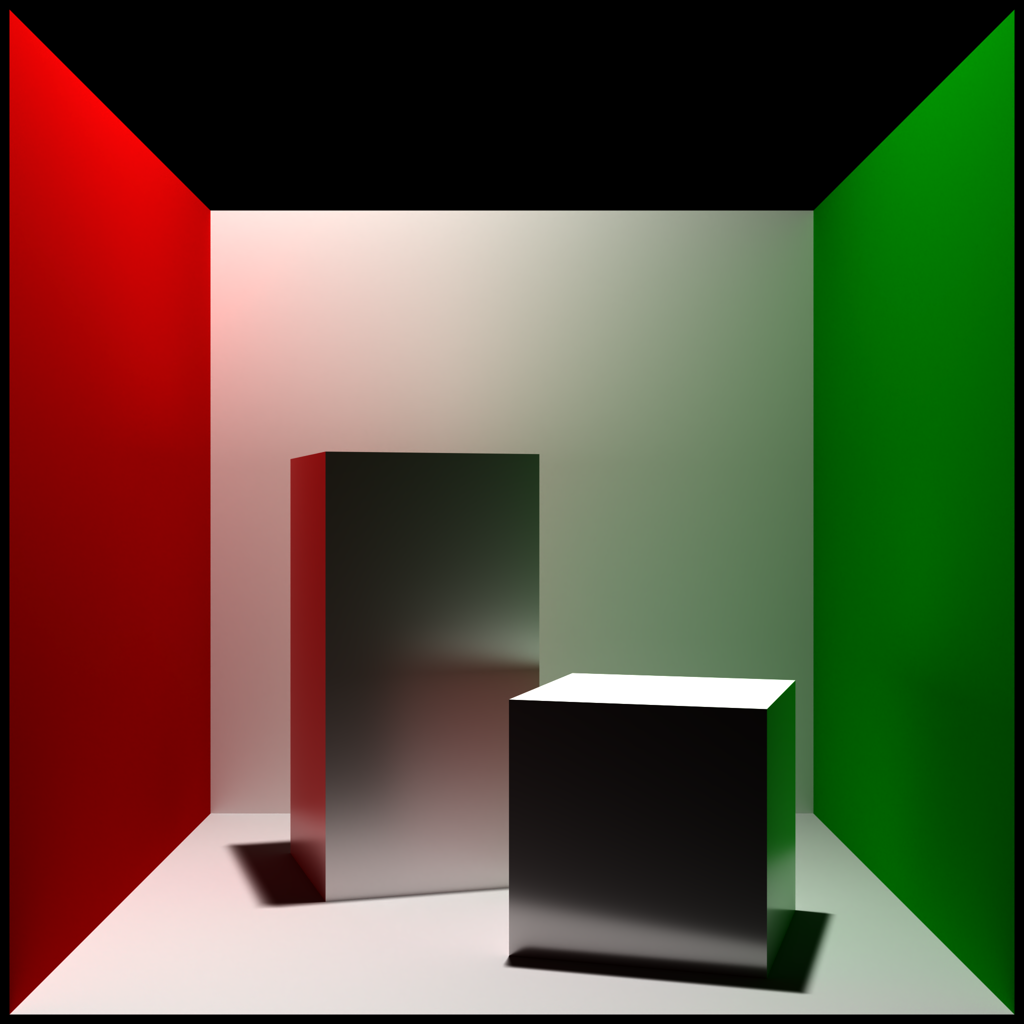}%
& \includegraphics[width=0.22\linewidth]{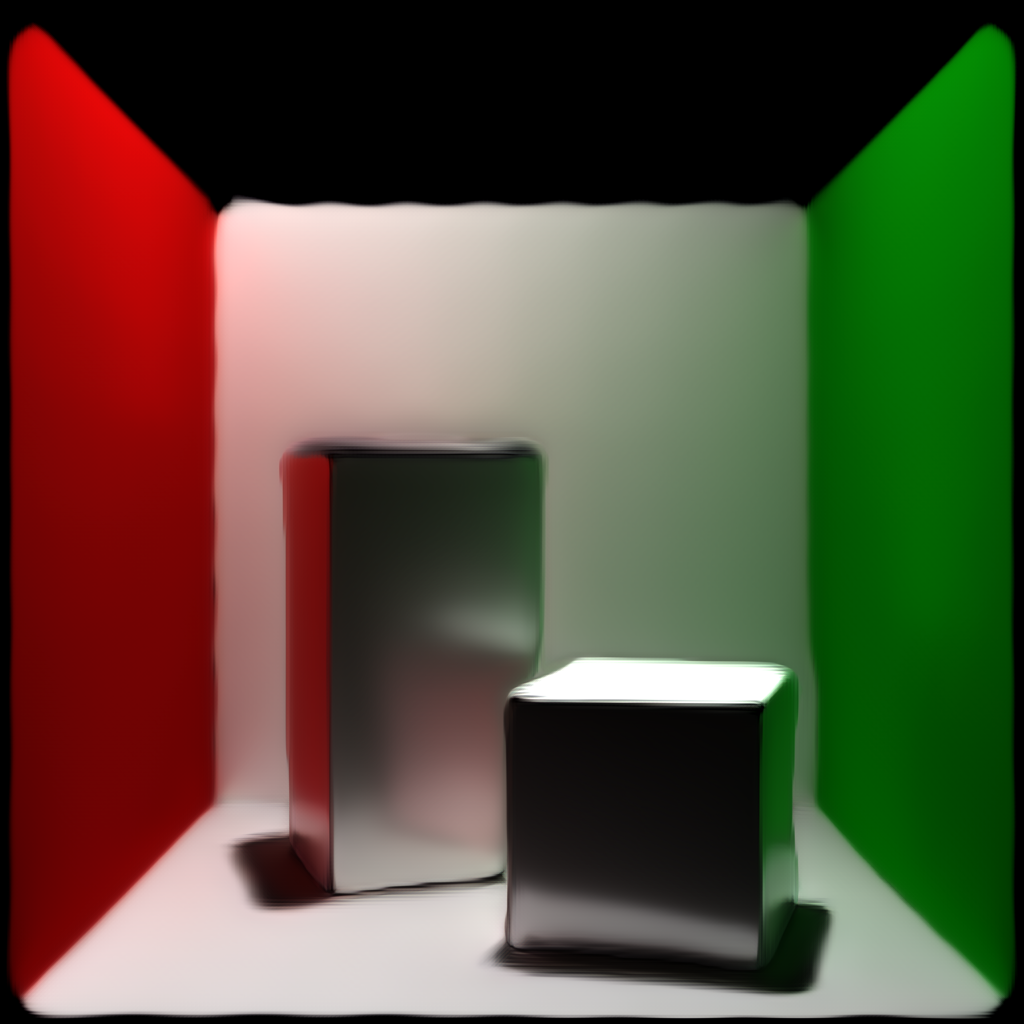}%
& \includegraphics[width=0.22\linewidth]{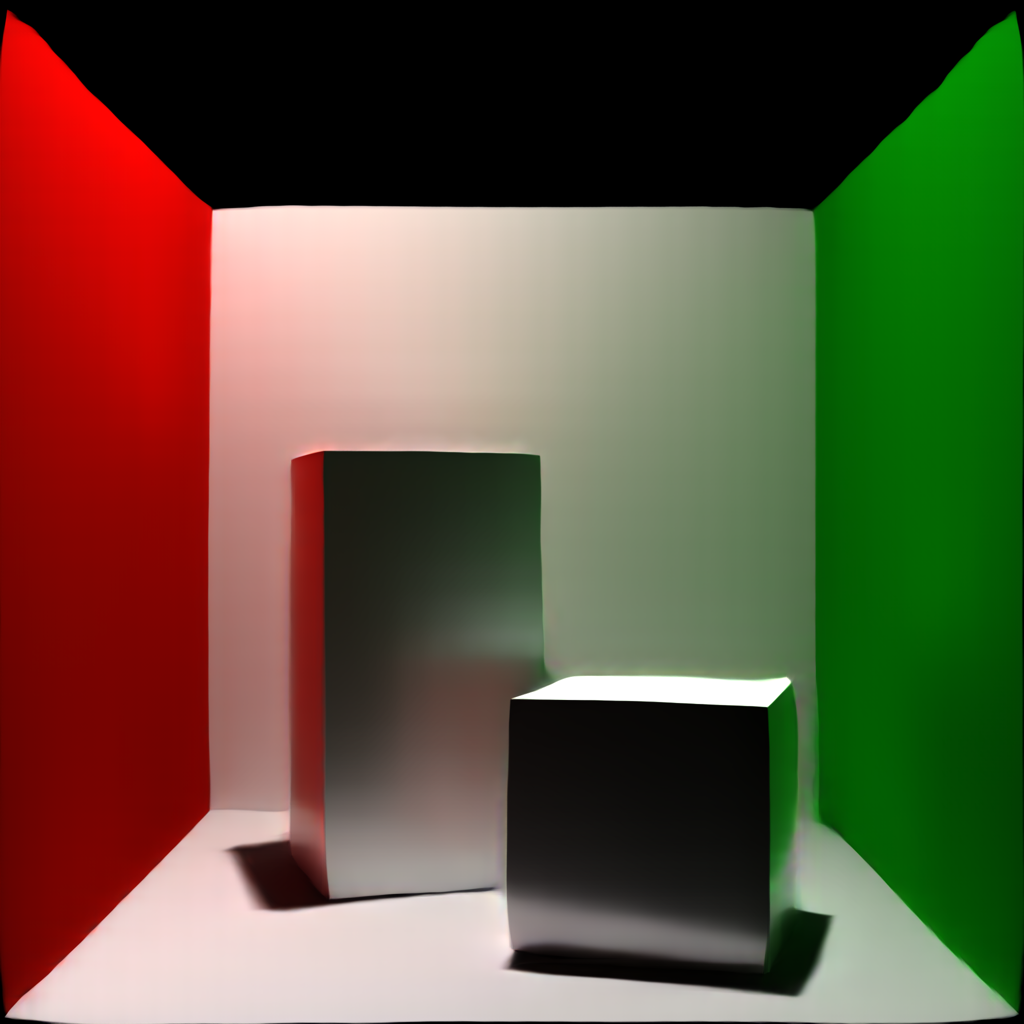}%
& \includegraphics[width=0.22\linewidth]{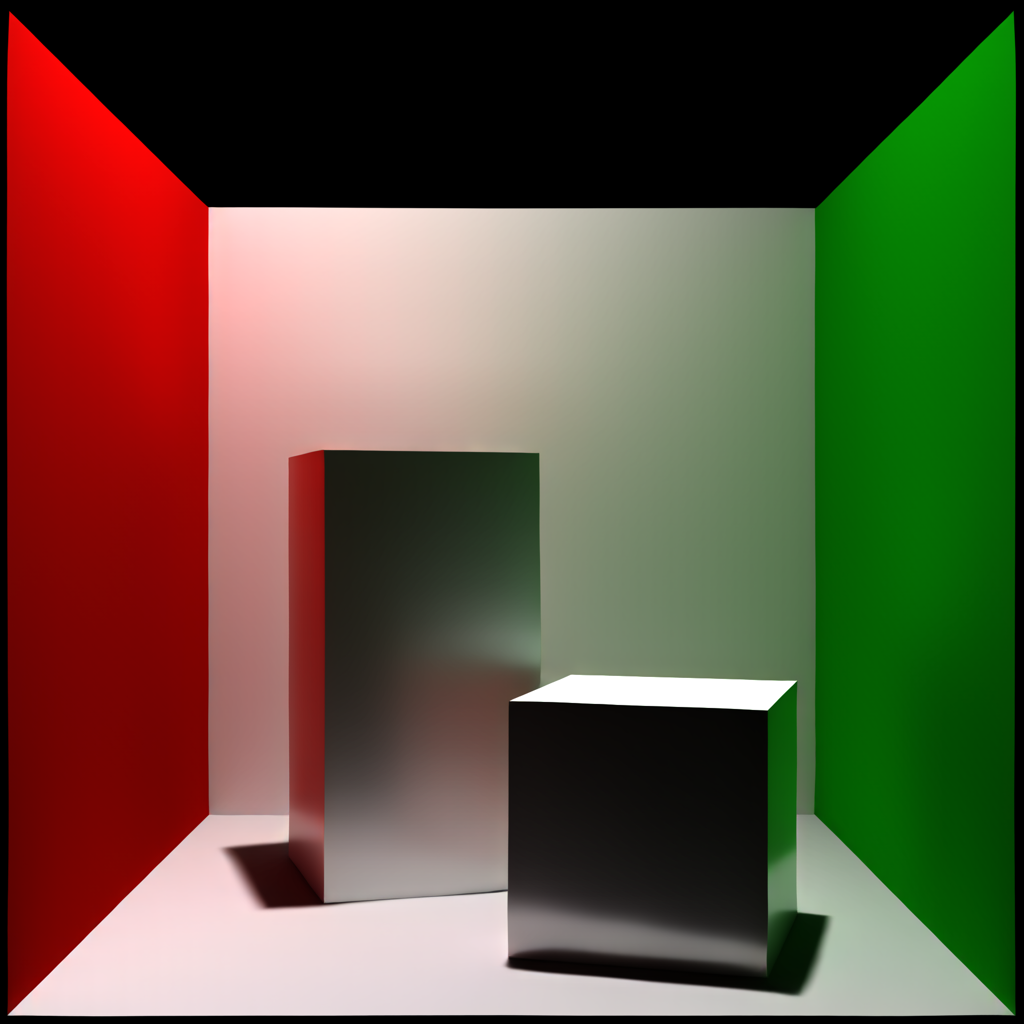} \\
& &
\includegraphics[width=0.22\linewidth]{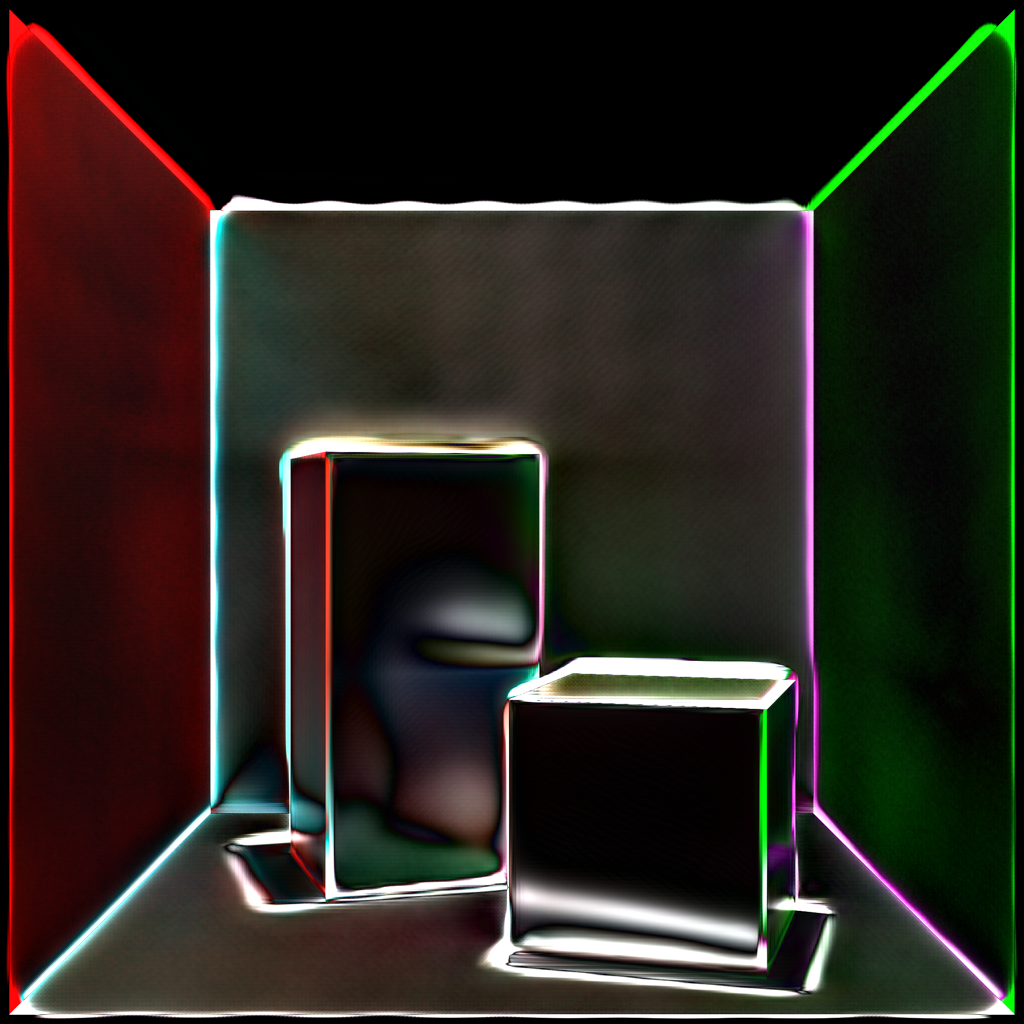}%
& \includegraphics[width=0.22\linewidth]{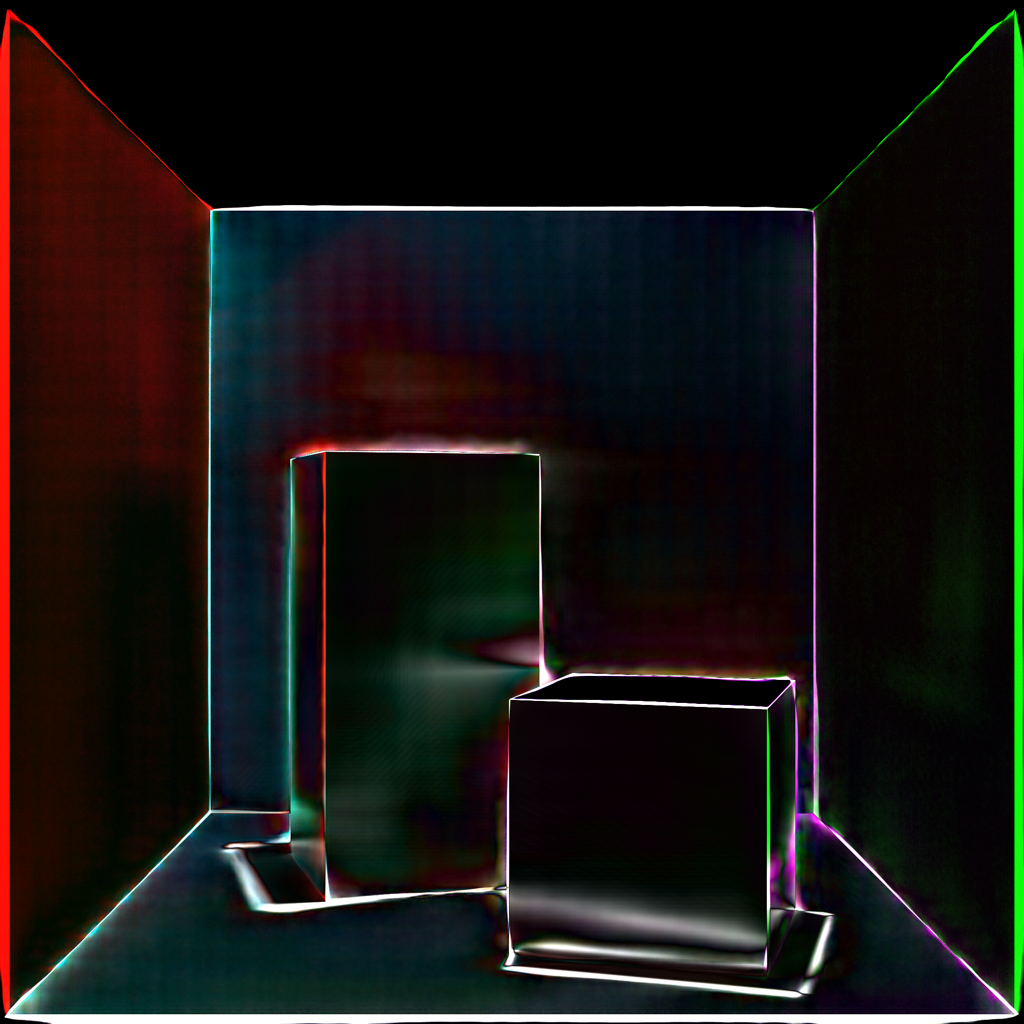}%
& \includegraphics[width=0.22\linewidth]{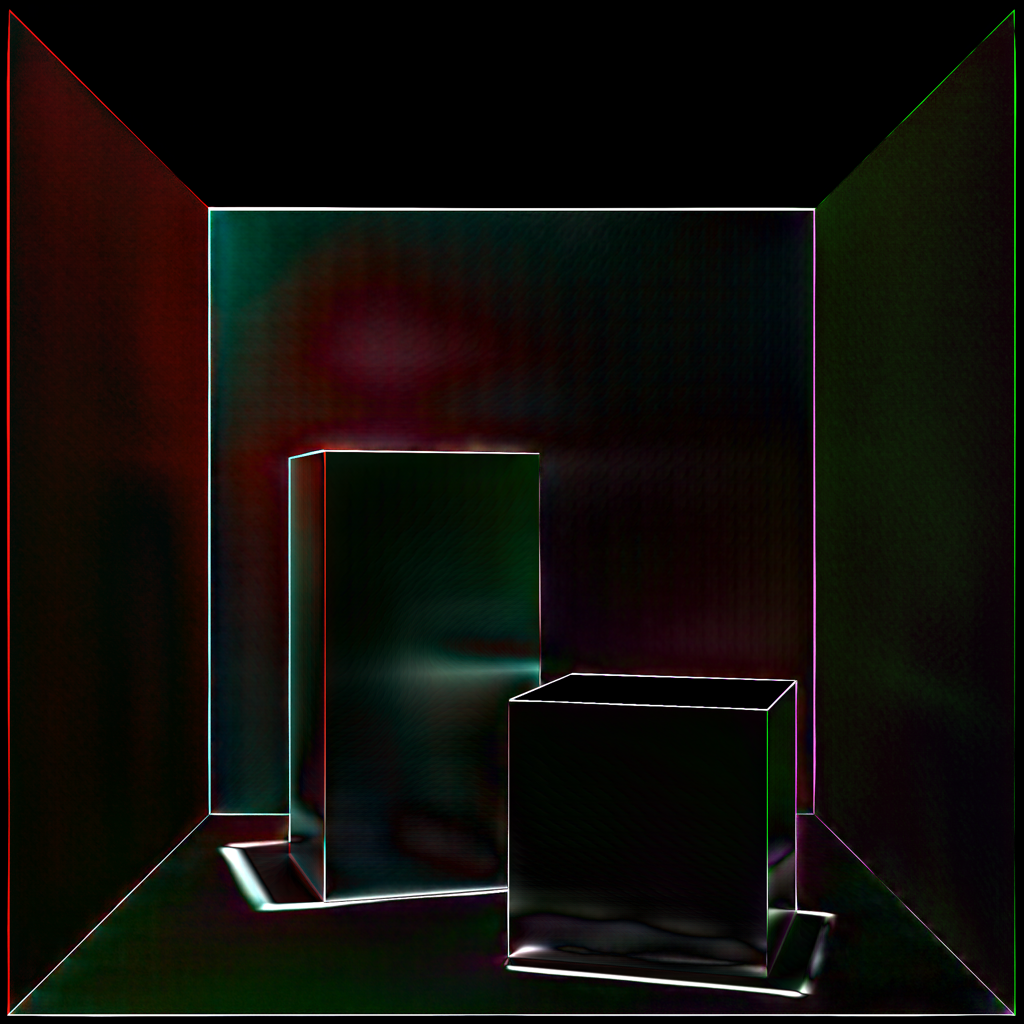} \\
\end{tabular}
\caption{Resolution scaling comparison of RenderFormer vs \RF2 (with and without high-resolution fine-tuning).}
\label{fig:supp_rf1_compare_res_scaling}
\end{figure}

\subsection{Resolution Scalability}
Thanks to the locality of the SWIN attention in the view-dependent
stage, \RF2 exhibits better scalability with respect to changes in
render-resolution.  We validate this quantitatively
(\autoref{tab:rf1_rf2_scaling_comparison}) and qualitatively
(\autoref{fig:supp_rf1_compare_res_scaling}) by comparing the accuracy
of RenderFormer vs. \RF2 vs. a resolution fine-tune of \RF2.
RenderFormer shows noticeably higher errors when rendering at higher
resolutions ($1024$ and $2048$). Moreover, fine-tuning \RF2 at
$2048$ resolution further improves \mbox{\RF2\!\!’s} ability to recover fine
details under high-resolution settings.

\begin{table}[t]
  \caption{Comparison between \RF2 (ours) and RenderFormer across different triangle counts. For each triangle count and each metric, the better result is highlighted in bold.}
  \centering
  \setlength{\tabcolsep}{6pt}
  \renewcommand{\arraystretch}{1.15}
  \begin{tabular}{lccccc}
    \specialrule{0.08em}{0pt}{0pt}
    \textbf{Tri Count} & \textbf{Model} & \textbf{PSNR} $\uparrow$ & \textbf{SSIM} $\uparrow$ & \textbf{LPIPS} $\downarrow$ & \textbf{FLIP} $\downarrow$ \\
    \specialrule{0.05em}{0pt}{0pt}

    4K & \RF2 & 33.84 & 0.9776 & \textbf{0.0198} & 0.1025 \\
       & RenderFormer & \textbf{34.51} & \textbf{0.9823} & 0.0244 & \textbf{0.0838} \\
    \specialrule{0.05em}{0pt}{0pt}

    8K & \RF2 & \textbf{33.00} & 0.9682 & \textbf{0.0305} & 0.1230 \\
       & RenderFormer & 32.89 & \textbf{0.9708} & 0.0358 & \textbf{0.1079} \\
    \specialrule{0.05em}{0pt}{0pt}

    16K & \RF2 & \textbf{29.96} & \textbf{0.9457} & \textbf{0.0502} & 0.1932 \\
        & RenderFormer & 29.36 & 0.9418 & 0.0604 & \textbf{0.1555} \\
    \specialrule{0.05em}{0pt}{0pt}

    32K & \RF2 & \textbf{28.07} & \textbf{0.9205} & \textbf{0.0729} & \textbf{0.2387} \\
        & RenderFormer & 24.17 & 0.8451 & 0.1500 & 0.3080 \\
    \specialrule{0.05em}{0pt}{0pt}

    64K & \RF2 & \textbf{26.79} & \textbf{0.9020} & \textbf{0.0873} & \textbf{0.2609} \\
        & RenderFormer & 18.72 & 0.6651 & 0.3070 & 0.5313 \\
    \specialrule{0.05em}{0pt}{0pt}

    128K & \RF2 & \textbf{25.82} & \textbf{0.8847} & \textbf{0.0999} & \textbf{0.3220} \\
         & RenderFormer & 15.88 & 0.4604 & 0.4772 & 0.7466 \\
    \specialrule{0.08em}{0pt}{0pt}
  \end{tabular}
  \label{tab:omniren_vs_renderformer}
\end{table}

\subsection{Full Detailed Metrics}
Finally, for completeness, in~\autoref{tab:omniren_vs_renderformer} we
show the full rendering quality comparison with RenderFormer which was
summarized in the main submission in the error plot (Fig.~6).

\begin{figure}[t!]
\centering
\setlength{\tabcolsep}{0pt}
\renewcommand{\arraystretch}{0}
\begin{tabular}{c c c c}
\includegraphics[width=0.25\linewidth]{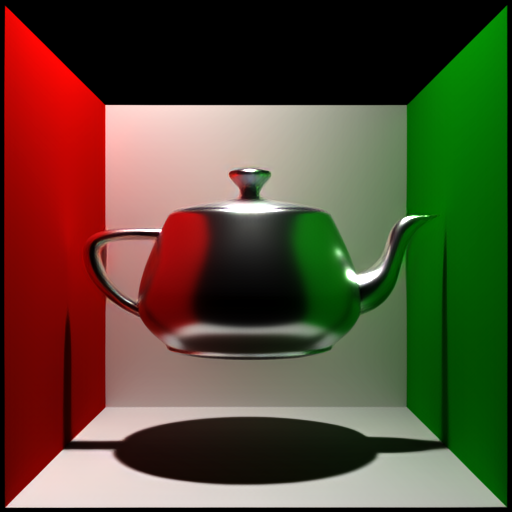} &
\includegraphics[width=0.25\linewidth]{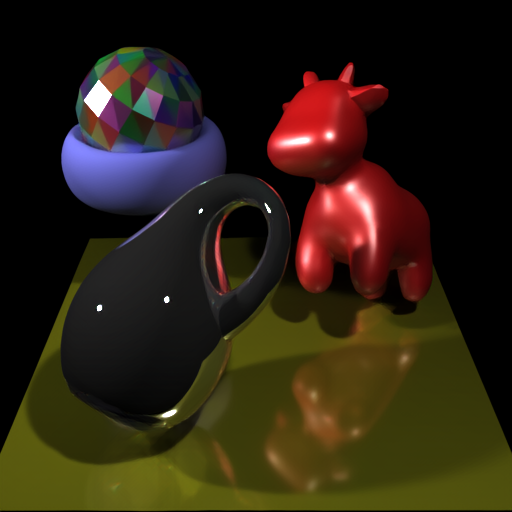} &
\includegraphics[width=0.25\linewidth]{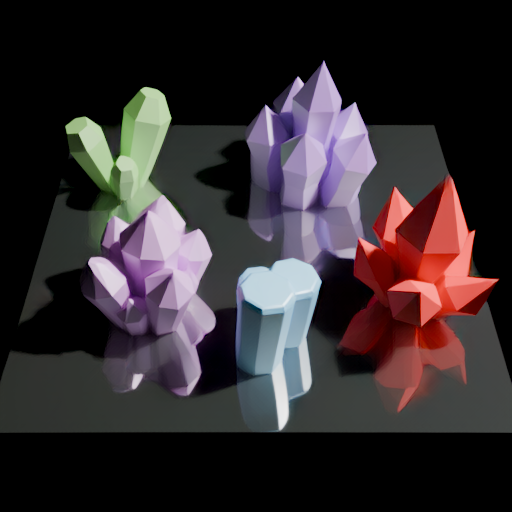} &
\includegraphics[width=0.25\linewidth]{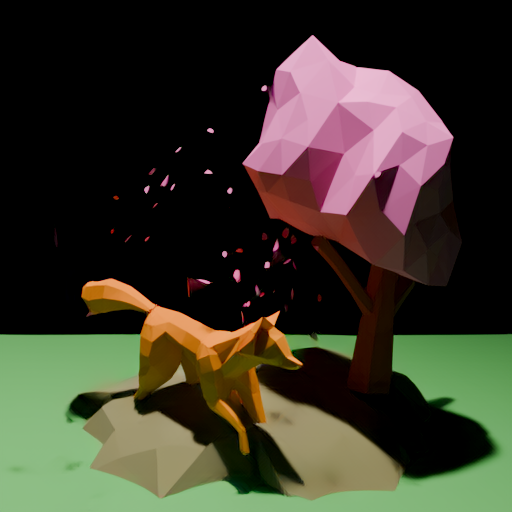} \\
\includegraphics[width=0.25\linewidth]{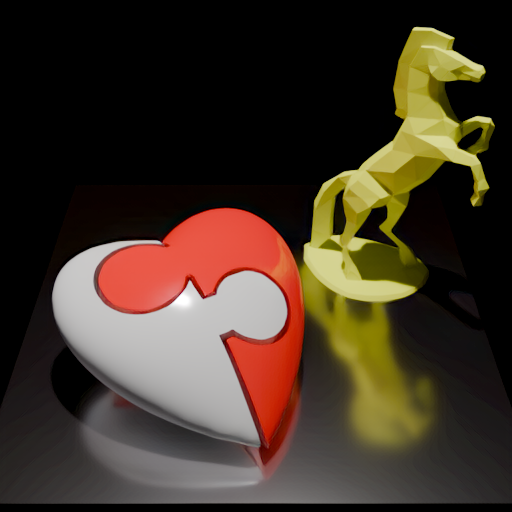} &
\includegraphics[width=0.25\linewidth]{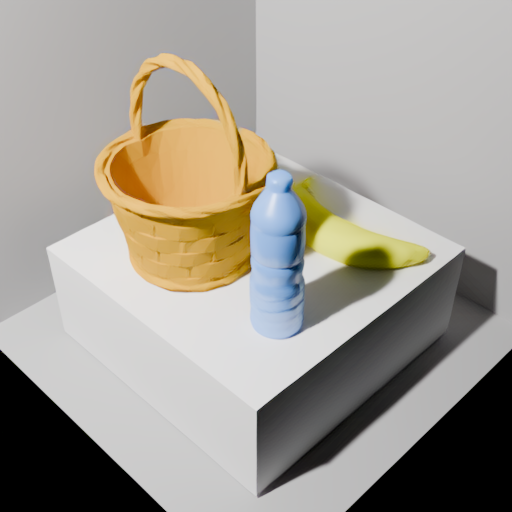} &
\includegraphics[width=0.25\linewidth]{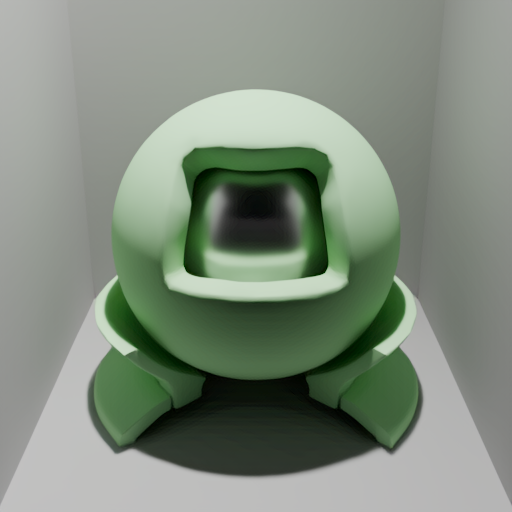} &
\includegraphics[width=0.25\linewidth]{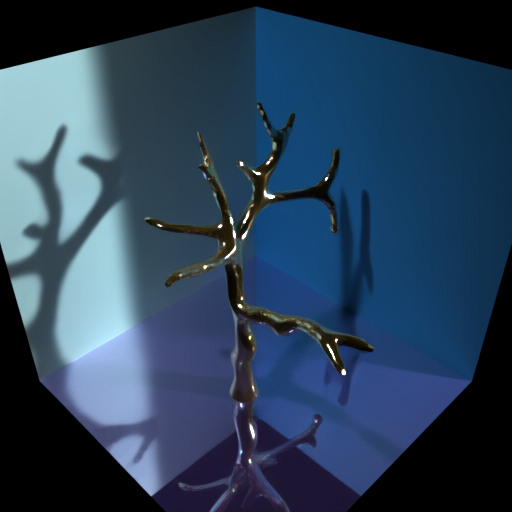} \\
\end{tabular}
\vspace{-0.2cm}
\caption{The $8$ test
  scenes from RenderFormer~\cite{Zeng:2025:RFT} rendered with \RF2.}
\label{fig:supp_rf1_compat_scenes}
\end{figure}

\begin{figure}[t]
\centering
\setlength{\tabcolsep}{0pt}
\renewcommand{\arraystretch}{0}
\begin{tabular}{c c c}
\includegraphics[width=0.33\linewidth]{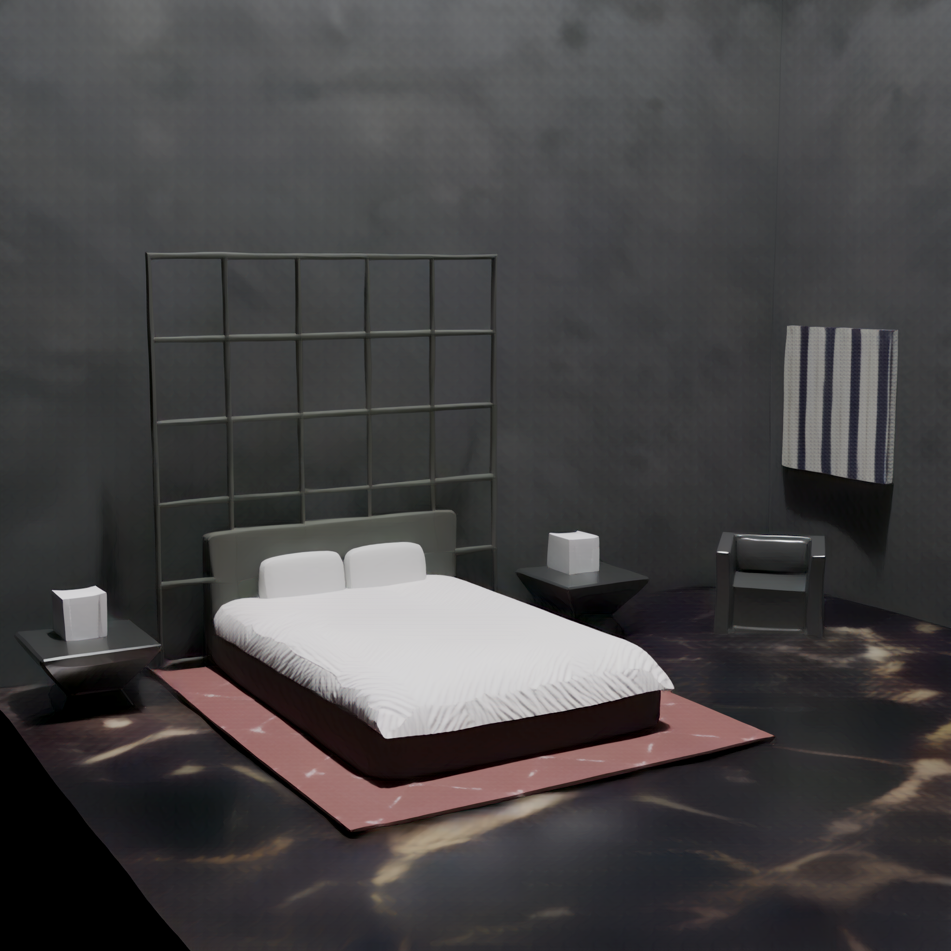} &
\includegraphics[width=0.33\linewidth]{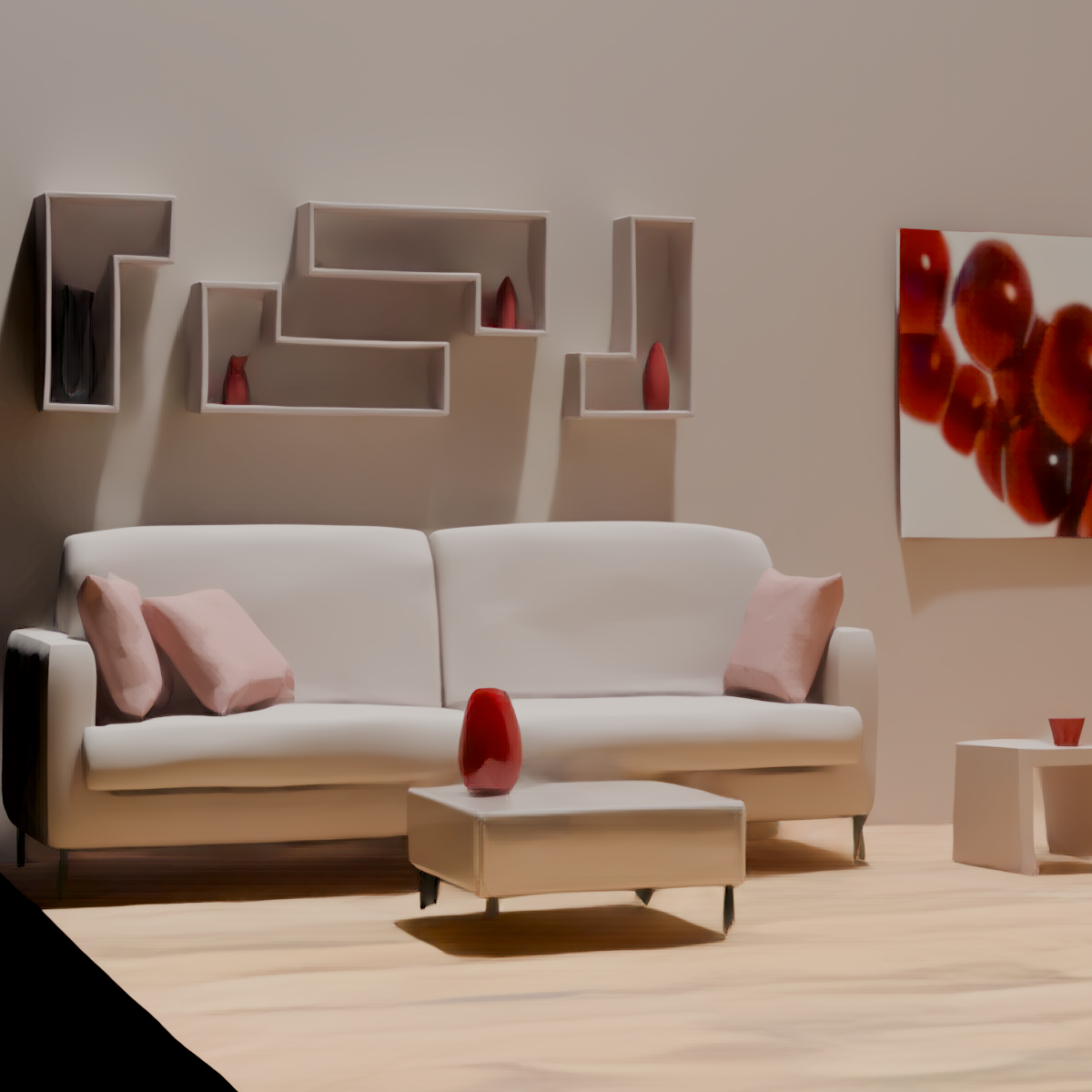} &
\includegraphics[width=0.33\linewidth]{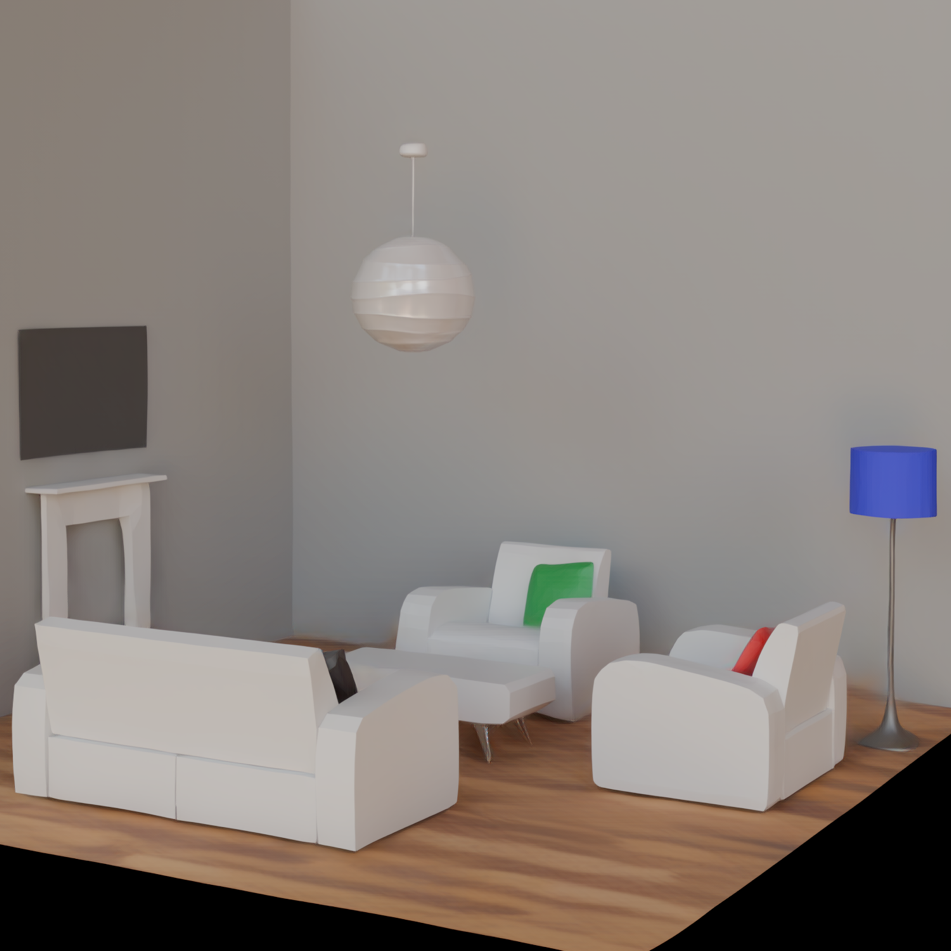} \\
\end{tabular}
\vspace{-0.2cm}
\caption{Additional room scenes rendered with \RF2.}
\label{fig:supp_room_rendering}
\end{figure}

\section{Additional  Results}
\autoref{fig:supp_rf1_compat_scenes} shows $8$ scenes used by
Zeng~\etal~\cite{Zeng:2025:RFT} to demonstrate the capabilities of
RenderFormer, which \RF2 can also handle without issues.

Finally, we show additional complex room scenes
in~\autoref{fig:supp_room_rendering} with fine geometrical details,
complex light transport, and textures.

We refer to the supplementary video for additional results
demonstrating \RF2's capabilities as well as its stability to
changes in scene and camera parameters.

\begin{figure}[t!]
\centering
\setlength{\tabcolsep}{0pt}
\renewcommand{\arraystretch}{0}
\begin{tabular}{c c c c c}
\textbf{Pred} & \textbf{Reference} & \textbf{Diff ($\times 5$)} & \textbf{FLIP} & \textbf{Metrics} \\
\centmark{\includegraphics[width=0.2\linewidth]{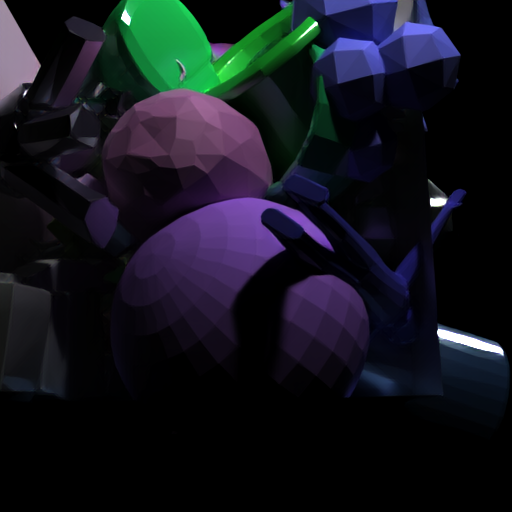}} &
\centmark{\includegraphics[width=0.2\linewidth]{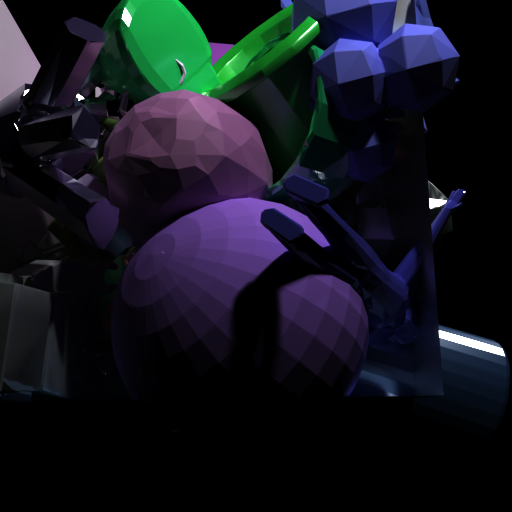}} &
\centmark{\includegraphics[width=0.2\linewidth]{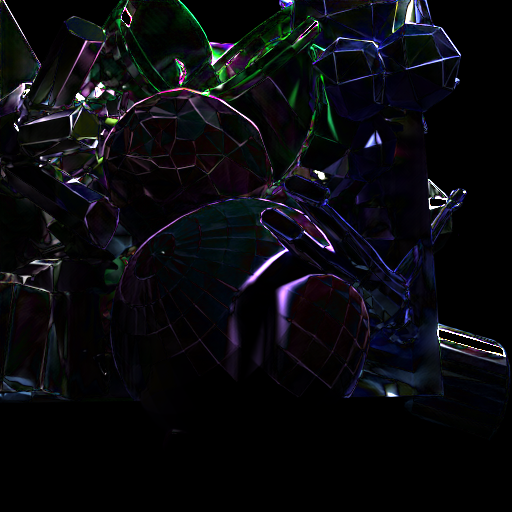}} &
\centmark{\includegraphics[width=0.2\linewidth]{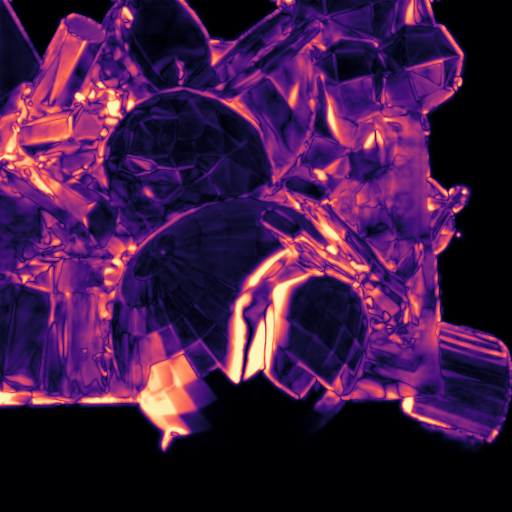}} &
\centmark{\parbox{0.2\linewidth}{\centering\small PSNR: 29.07\newline SSIM: 0.9285\newline LPIPS: 0.04980\newline FLIP: 0.2044}} \\
\centmark{\includegraphics[width=0.2\linewidth]{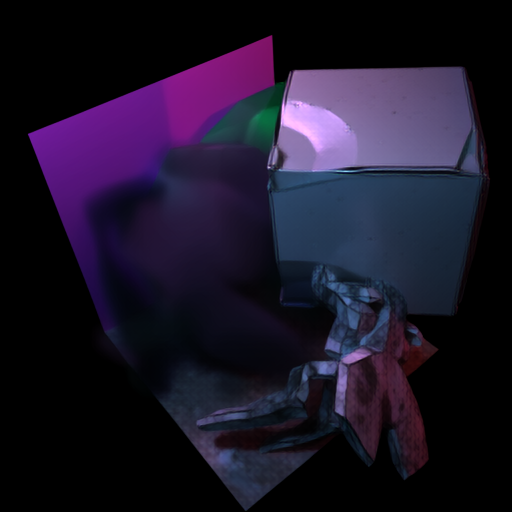}} &
\centmark{\includegraphics[width=0.2\linewidth]{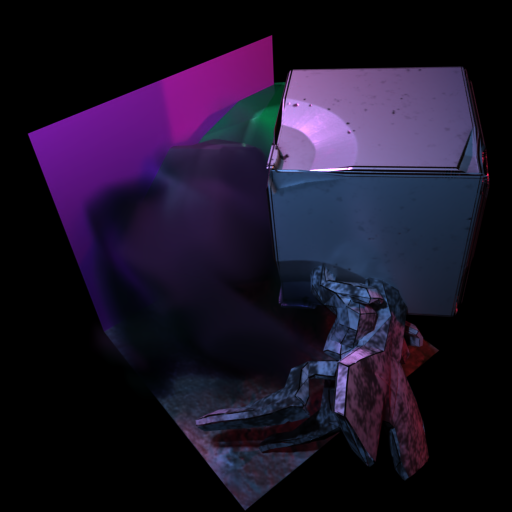}} &
\centmark{\includegraphics[width=0.2\linewidth]{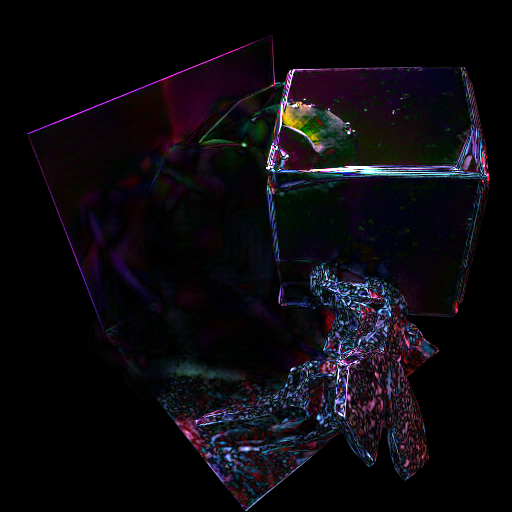}} &
\centmark{\includegraphics[width=0.2\linewidth]{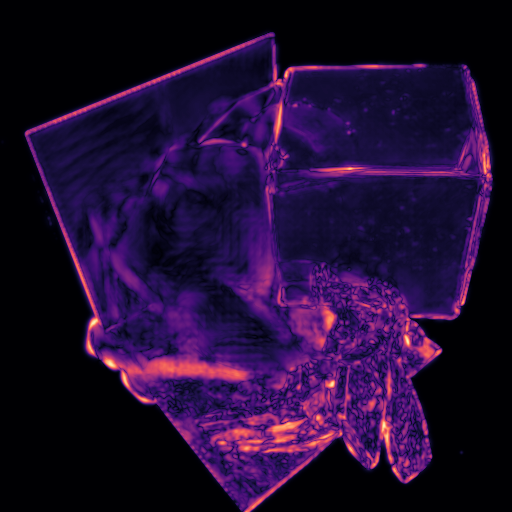}} &
\centmark{\parbox{0.2\linewidth}{\centering\small PSNR: 30.92\newline SSIM: 0.9353\newline LPIPS: 0.04512\newline FLIP: 0.1125}} \\
\centmark{\includegraphics[width=0.2\linewidth]{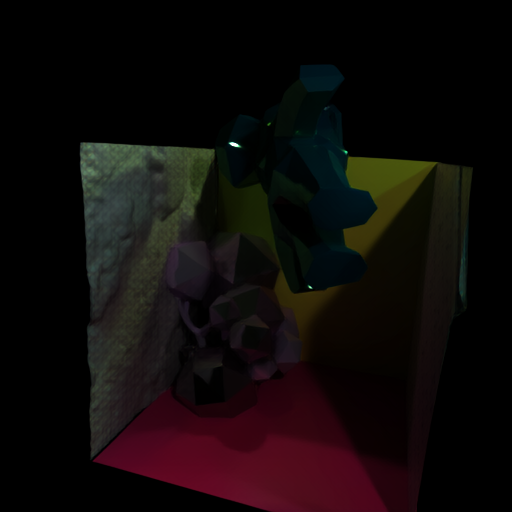}} &
\centmark{\includegraphics[width=0.2\linewidth]{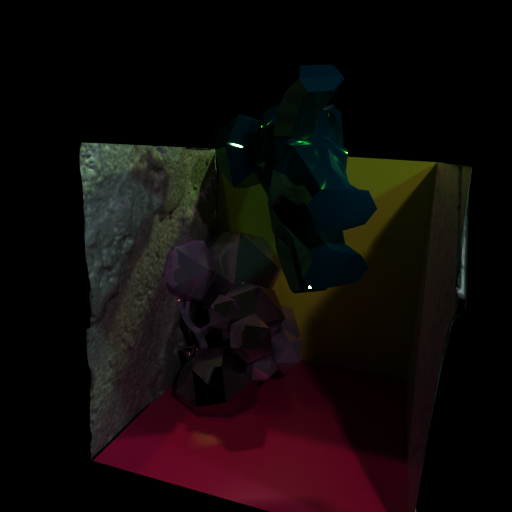}} &
\centmark{\includegraphics[width=0.2\linewidth]{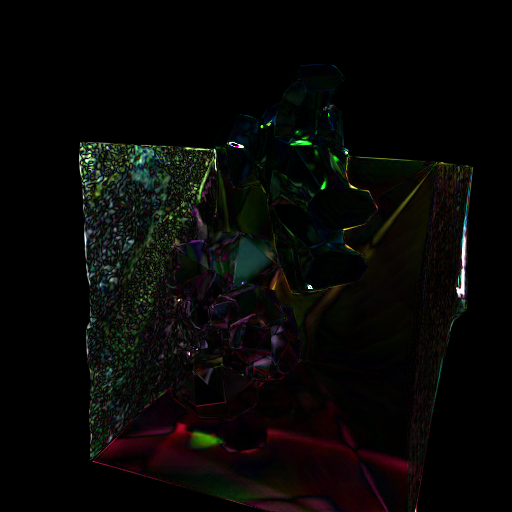}} &
\centmark{\includegraphics[width=0.2\linewidth]{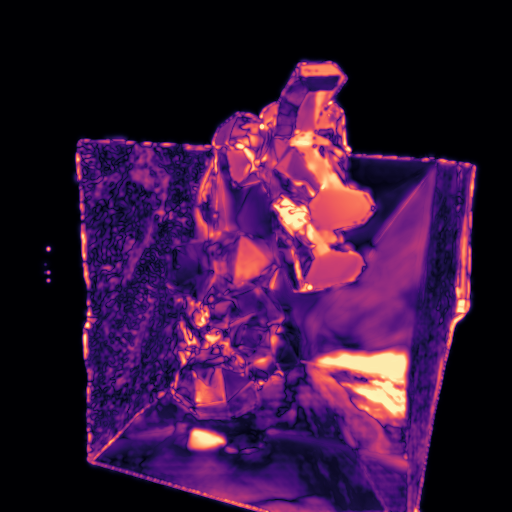}} &
\centmark{\parbox{0.2\linewidth}{\centering\small PSNR: 34.99\newline SSIM: 0.9506\newline LPIPS: 0.06065\newline FLIP: 0.1719}} \\
\centmark{\includegraphics[width=0.2\linewidth]{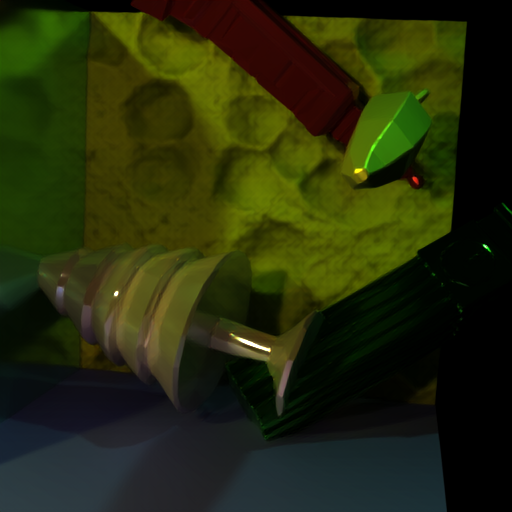}} &
\centmark{\includegraphics[width=0.2\linewidth]{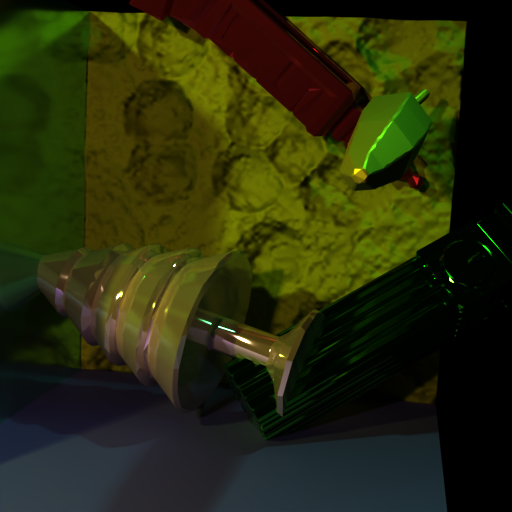}} &
\centmark{\includegraphics[width=0.2\linewidth]{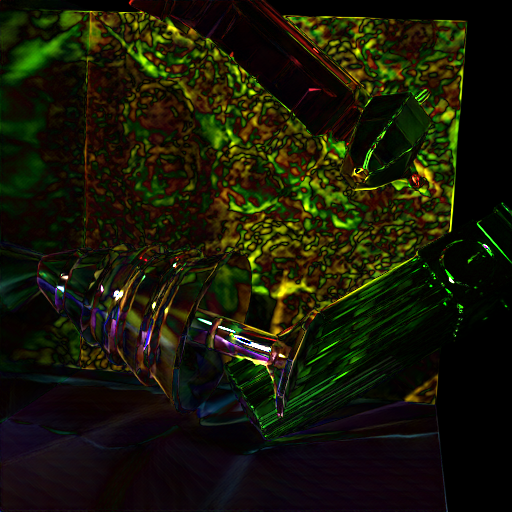}} &
\centmark{\includegraphics[width=0.2\linewidth]{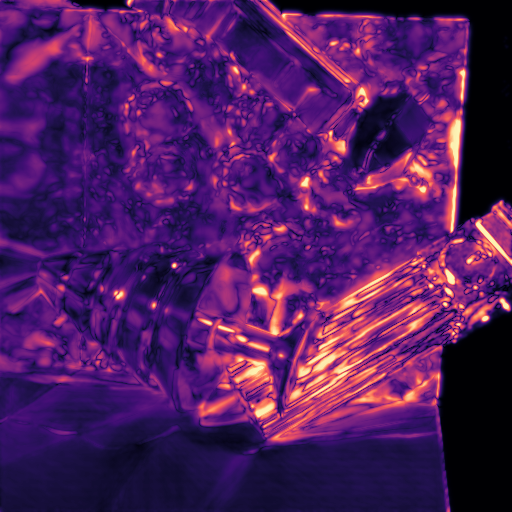}} &
\centmark{\parbox{0.2\linewidth}{\centering\small PSNR: 28.92\newline SSIM: 0.8827\newline LPIPS: 0.1240\newline FLIP: 0.2515}} \\
\end{tabular}
\vspace{-0.2cm}
\caption{ A variety of procedurally generated ablation scenes rendered
  with \RF2 and compared to path-traced reference images.  We also
  report PSNR, SSIM, LPIPS, and FLIP errors.  }
\label{fig:supp_omniren_results}
\end{figure}

\section{Ablation Test Scenes}
\autoref{fig:supp_omniren_results} shows example scenes from our test
set used for the ablation studies. These procedurally generated scenes
contain diverse textures, a variety of test objects, large scene token
counts, as well as volumetric effects, area lighting, and environment
lighting.

\section{Limitations}
\begin{figure}[t]
\centering
\setlength{\tabcolsep}{0pt}
\renewcommand{\arraystretch}{0}
\begin{tabular}{c c c}
\normalsize Reference & \normalsize No Subdiv. & \normalsize $2 \times$ \\[2pt]
\includegraphics[width=0.32\linewidth]{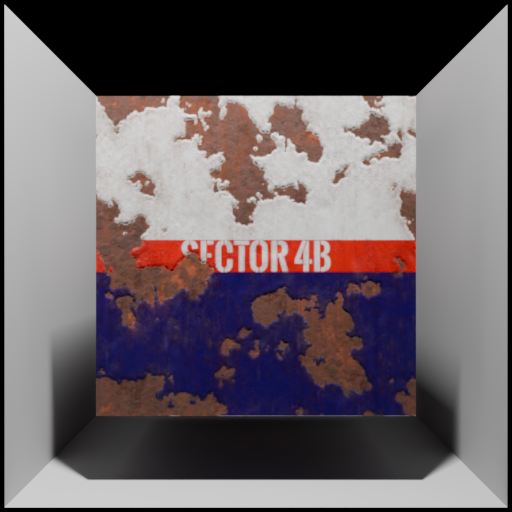} &
\includegraphics[width=0.32\linewidth]{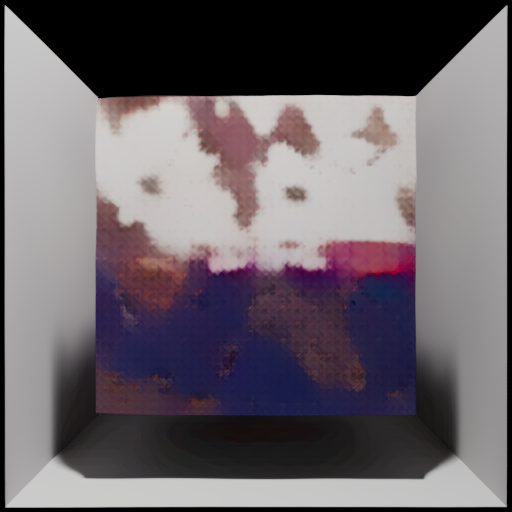} &
\includegraphics[width=0.32\linewidth]{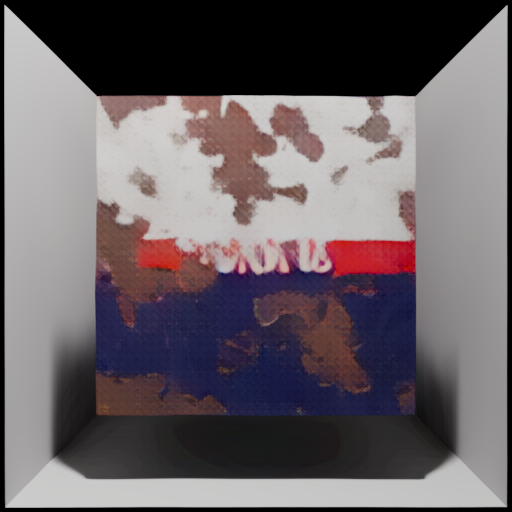} \\[2pt]
\normalsize $4 \times$ & \normalsize $8 \times$ & \normalsize $16 \times$ \\[2pt]
\includegraphics[width=0.32\linewidth]{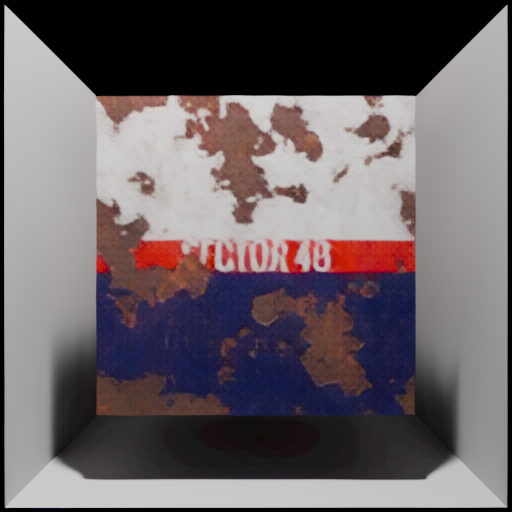} &
\includegraphics[width=0.32\linewidth]{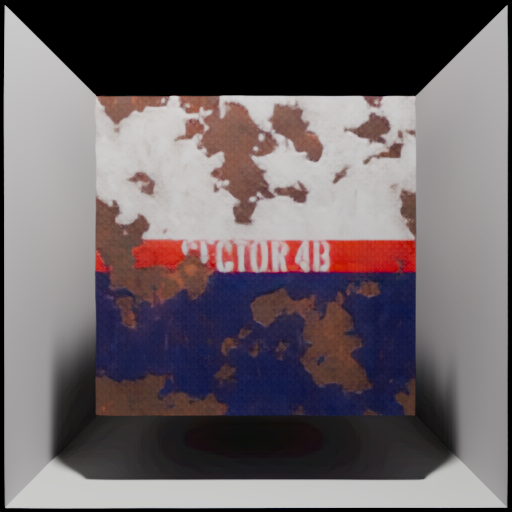} &
\includegraphics[width=0.32\linewidth]{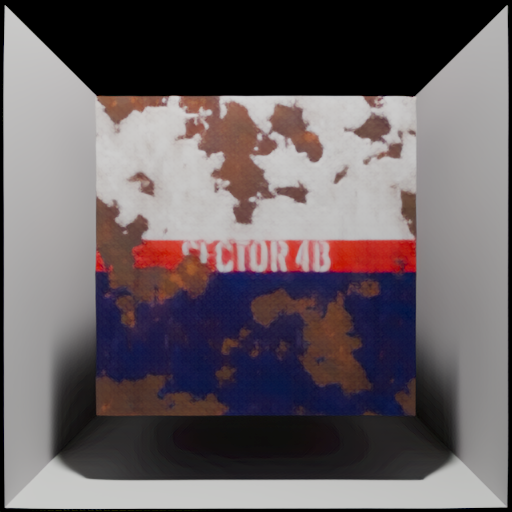} \\
\end{tabular}
\caption{ Limitations: large textured triangle can result in loss of
  texture sharpness.  Increasing triangle subdivision level improves
  texture quality, with finer details and fewer texture artifacts at
  higher subdivision levels.  }
\label{fig:supp_texture_limitations_subdivision}
\end{figure}

\autoref{fig:supp_texture_limitations_subdivision} illustrates the
limitations of \RF2 in rendering high-resolution textures on large
triangles. Since each triangle is associated with a fixed-resolution
texture embedding ($32 \times 32$), the effective texture resolution
becomes low when the triangle covers a large surface area. By
subdividing the mesh to increase the number of triangles and thus
reduce triangle size, the texture quality can be significantly
improved.

\bibliographystyle{splncs04}
\bibliography{src/reference}